\documentclass[12pt,a4paper,oneside]{report}

\usepackage[utf8]{inputenc} % UTF-8 encoding
\usepackage[T1]{fontenc}    % Type1 font encoding
\usepackage{lmodern}        % Latin Modern font
\usepackage{microtype}      % Better typography

\usepackage[a4paper, left=2.5cm, right=2.5cm, top=3cm, bottom=3cm, headheight=15pt]{geometry}

\usepackage[backend=biber,style=numeric-comp,natbib=true,sorting=none]{biblatex}
\usepackage{graphicx}
\usepackage[font=small]{caption}
\usepackage{subcaption}
\usepackage{booktabs}

\usepackage{amsmath,amssymb,amsthm}
\DeclareMathOperator{\Sign}{sgn}

\usepackage{algorithm,algpseudocode}

\usepackage{listings}
\usepackage[hidelinks]{hyperref}

\usepackage{setspace}
\usepackage{fancyhdr}
\newcommand{\ThesisTitle}{Learning by Steering the Neural Dynamics: A Statistical Mechanics Perspective}
\newcommand{\ThesisAuthor}{Mattia Scardecchia}
\newcommand{\ThesisSupervisor}{Prof. Riccardo Zecchina}
\newcommand{\ThesisDepartment}{Department of Computing Sciences}
\newcommand{\ThesisUniversity}{Bocconi University}
\newcommand{\ThesisDate}{June 12, 2025}
\newcommand{\ThesisLogo}{figures/bocconi-logo.png}

\begin{document}

% Title page
\begin{titlepage}
  \centering
  \includegraphics[width=4cm]{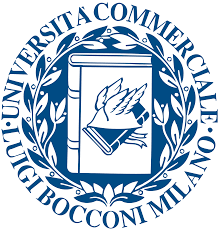}\par\vspace{1cm}
  {\LARGE \bfseries \ThesisTitle \par}
  \vspace{1cm}
  {\large by \par}
  {\Large \textbf{\ThesisAuthor} \par}
  \vfill
  {\Large Final Dissertation, MSc in Artificial Intelligence \par}
  {\Large \ThesisDepartment, \ThesisUniversity \par}
  {\Large Supervisor: \ThesisSupervisor \par}
  {\Large \ThesisDate \par}
  % \vspace{1cm}
\end{titlepage}

% Thanks
\newpage
\vspace*{5cm}              % vertical space before the text; tweak as desired

\begin{center}
  \textit{%
    Alla mia famiglia
  }
\end{center}
\thispagestyle{empty}

% % Declaration
% \chapter*{Declaration}
% I hereby declare that this thesis is my own work and has not been submitted elsewhere.

% Acknowledgements
\chapter*{Acknowledgements}
% \addcontentsline{toc}{chapter}{Acknowledgements}
I thank my supervisor, Prof. Riccardo Zecchina, for his invaluable guidance and support.
I am also grateful to Prof. Carlo Baldassi, Prof. Marc Mézard, and Davide Badalotti for their insightful
feedback and discussions.

% Abstract
\chapter*{Abstract}
% \addcontentsline{toc}{chapter}{Abstract}
Despite the striking successes of deep artificial neural networks trained with gradient-based optimization, these
methods differ fundamentally from their biological counterparts. This discrepancy raises important questions: how
does nature implement robust, sample-efficient learning at minimal energy costs? And how does it solve the credit
assignment problem, in light of the biological implausibility of backpropagation? In this thesis, we take a step 
towards bridging the gap between contemporary AI and computational neuroscience, by exploring how the neural dynamics
can facilitate fully local and distributed learning strategies that scale to simple machine learning benchmarks.

Using tools from the statistical mechanics of disordered systems, we identify conditions for the appearance of
robust dynamical attractors in a random asymmetric recurrent neural network model. First we derive a
closed-form expression for the number of fixed points, as a function of the self-coupling strength, valid
with high probability in the limit of large networks. Then, we find evidence of a phase transition affecting
the structure of the fixed points as the self-coupling strength crosses a critical value: below this threshold,
isolated fixed points coexist with exponentially many, extremely narrow clusters of fixed points that
satisfy the overlap-gap property; above it, subdominant but extremely dense and extensive clusters of fixed points
emerge. These fixed points are accessible to algorithms like focusing Belief Propagation (fBP), which explicitly target highly locally
entropic regions, and they become accessible to a simple asynchronous dynamical rule after a second size-dependent
threshold is crossed.

We then propose a simple, biologically plausible learning algorithm that can be used for supervised learning with any binary recurrent
neural network. The idea is to map input-output relationships to fixed points of the network dynamics,
by letting the network relax under the influence of the external stimuli, and then reinforcing the fixed point
according to a local synaptic plasticity rule inspired by perceptron learning. We show that our algorithm effectively
learns an entangled version of MNIST, and that it can leverage depth to increase its hetero-association
capacity. We also demonstrate that our approach can successfully handle several architectures, and that it can 
learn hierarchically organized features in deep networks. Finally, we show that the performance of our algorithm
is significantly affected by the choice of the self-coupling strength, hinting at the relevance for learning of the
accessible structures of fixed points identified in the first part of the thesis. Interestingly, introducing a few
strong ferromagnetic interactions in the network, inspired by known features of the cortex, plays a similar
stabilizing role as the self-coupling.

Future work will address the scalability of our approach to more challenging machine learning benchmarks, leveraging
the demonstrated flexibility to introduce more sophisticated architectures with inductive biases,
as well as take biological realism one step further by introducing spiking dynamics, populations of excitatory and inhibitory
neurons, sparsity, and insights from connectomics.

% Table of contents, figures, tables
\tableofcontents
\listoffigures
\listoftables

% Main chapters
\chapter{Introduction}
\label{ch:introduction}
% - success DL
% - theory not well developed (maybe mention some approaches like infinite width limit, etc.)
% - promising avenue is through stat mech
% - some successes of stat mech to understand the onset of hardness in constraint satisfaction problems
% - local entropy: stat mech formalism to study existence of dense regions of solutions
% - overlap gap property: the intuition behind local entropy is correct, dense clusters of solutions are the relevant ones for learning. onset of OGP related to the onset of hardness

% - would like to take this a step towards neuroscience
% - consider recurrent neural networks with asymmetric couplings --> caricature of biological networks
% - as a first step, learning could work by mapping memories to fixed points
% - need to understand the structure of fixed points in these networks. is there similar phenomenology as in shallow feed-forward networks?
% - if yes, is the emergence of the fixed points clusters relevant for learning in this setting?
% - bio-plausible algo that is a bridge btw ml and neuroscience; shows hierarchical features; can be studied through stat mech even with deep architecture.
\section{Motivation}

In the last decade, with the rise of deep learning \cite{krizhevskyImageNetClassificationDeep2012, lecunDeepLearning2015a}, we have witnessed tremendous advancements in the field of Artificial
Intelligence (AI). Fueled by massive datasets and huge amounts of compute \cite{bommasaniOpportunitiesRisksFoundation2021}, contemporary AI systems
are revolutionizing entire fields of science and technology \cite{silverMasteringGameGo2017, jumperHighlyAccurateProtein2021, priceProbabilisticWeatherForecasting2025, openaiGPT4TechnicalReport2024a}, and promise to have a transformative impact on our societies. 

However, despite their impressive achievements, the relationship between deep artificial neural networks (ANNs) and biological networks 
of neurons is still unclear. In fact, although they are loosely inspired by the brain \cite{rosenblattPerceptronProbabilisticModel1958}, deep learning systems
differ fundamentally from biological neural networks in several key aspects. Algorithmically, modern feed-forward deep
neural networks, like Transformers \cite{vaswaniAttentionAllYou2023}, rely on the backpropagation algorithm to solve the credit assignment problem \cite{rumelhartLearningRepresentationsBackpropagating1986}, and
they lack a time dimension. In contrast, biological neural networks exhibit rich temporal dynamics \cite{rabinovichNeuroscienceTransientDynamics2008, buonomanoStatedependentComputationsSpatiotemporal2009}, and their synaptic
plasticity is believed to be governed by local learning rules \cite{morrisHebbOrganizationBehavior1999, biSynapticModificationsCultured1998}. Furthermore, biological networks tend to be extremely
energy-efficient, sample-efficient, and robust to noise and adversarial perturbations, compared to their artificial counterparts \cite{yuEvaluatingGrayWhite2018, khajehnejadBiologicalNeuronsCompete2024, geirhosGeneralisationHumansDeep2018}.

In spite of these differences, there is ample margin for fruitful dialogue between Computational Neuroscience and Artificial
Intelligence. On one hand, there is already a history of biologically-inspired innovations in AI, for example with Convolutional
Neural Networks \cite{hubelReceptiveFieldsBinocular1962, fukushimaNeocognitronSelforganizingNeural1980, lecunGradientbasedLearningApplied1998}, TD-based Reinforcement Learning \cite{suttonLearningPredictMethods1988, schultzNeuralSubstratePrediction1997}, and attention mechanisms \cite{desimoneNeuralMechanismsSelective1995, bahdanauNeuralMachineTranslation2014}. Going forward,
a greater understanding of how the brain functions could inspire the next generation of energy-efficient, sample-efficient, and
robust AI systems, with innovations in both software \cite{hassabisNeuroscienceInspiredArtificialIntelligence2017a} and hardware \cite{merollaMillionSpikingneuronIntegrated2014, daviesLoihiNeuromorphicManycore2018, indiveriMemoryInformationProcessing2015}. On the other hand, deep learning systems
are starting to be used as \emph{in silico} computational models of neural circuits \cite{kriegeskorteDeepNeuralNetworks2015, yaminsUsingGoaldrivenDeep2016}, and for neural data analysis 
and brain decoding \cite{anumanchipalliSpeechSynthesisNeural2019, levyBraintoTextDecodingNoninvasive2025}.

This thesis aims to take a step towards bridging the gap between neuroscience models and deep learning systems, by exploring
how neural dynamics can facilitate fully local and distributed learning strategies in recurrent neural networks. Using tools
from the statistical mechanics of disordered systems \cite{mezardSpinGlassTheory1986}, we study the dynamical behavior of generic asymmetric neural systems,
identifying conditions for the emergence of robust dynamical attractors that, based on known results about the onset of hardness
in random constraint satisfaction problems \cite{monassonDeterminingComputationalComplexity1999, mezardAnalyticAlgorithmicSolution2002, baldassiSubdominantDenseClusters2015, baldassiUnreasonableEffectivenessLearning2016, gamarnikOverlapGapProperty2021}, are expected to be relevant for learning. Using insights from this theoretical
analysis, we propose a biologically plausible, local and distributed framework for supervised learning that can be used
with any recurrent neural network, including deep architectures. Our experiments demonstrate that our approach can
learn useful, hierarchical features without explicitly computing or approximating gradients, allowing it to solve
non-linearly-separable classification tasks.

\section{Structure of the Thesis}

The structure of the thesis is as follows:
\begin{itemize}
    \item In Chapter \ref{ch:background}, we provide an overview of some relevant background material. We first recall some ideas from statistical mechanics, the replica method, and the local entropy formalism, which will be useful for the theoretical analysis in Chapter \ref{ch:analytical}. Then, we review some known results about the dynamical behavior of simple models of neural networks, focusing on the existence of attractors and the onset of chaos. Finally, we review the literature on biologically plausible learning algorithms for neural networks, discussing the main trends and challenges in the field.
    \item In Chapter \ref{ch:analytical}, we use tools from the statistical mechanics of disordered systems to study the existence and structure of fixed points in random asymmetric recurrent neural networks. Using the replica method, we derive an analytical expression for the number of fixed points in the limit of large systems, valid with high probability. Then, we show how to use a modification of the 1RSB formalism to find evidence for the emergence, at a critical value of the self-interaction strength, of subdominant, dense regions of fixed points, and discuss their accessibility for simple dynamical rules.
    \item In Chapter \ref{ch:numerical}, using insights from Chapter \ref{ch:analytical}, we propose a novel learning algorithm for deep recurrent neural networks, that learns to map input-output relations onto fixed points of the network dynamics. Our algorithm is fully local and distributed, it does not explicitly compute nor approximate gradients, and it works with asymmetric couplings. We assess its performance in the classification and hetero-association tasks, using both synthetic datasets and a more difficult, "entangled" version of the MNIST dataset. We show that our approach can handle a variety of shallow and deep network architectures, and that it can learn a hierarchy of features to solve non-linearly-separable classification tasks.
\end{itemize}

\chapter{Background and Related Work}
\label{ch:background}
In this chapter, we review some background material that will be relevant for the rest of the thesis,
touching on three main topics.

First, we introduce the formalism of statistical mechanics and some ideas
from the theory of spin glasses. We illustrate their application to the analysis of constraint satisfaction
and optimization problems, and we discuss a recently developed large-deviation approach to study a class of rare
configurations that are accessible to stable algorithms and turn out to be relevant for learning.
These concepts will be crucial for the theoretical analysis presented in Chapter \ref{ch:analytical},
where we study the phase diagram of a rate-based model of recurrent neural network with asymmetric synapses
as a function of the strength of self-couplings.

Second, we review some known results on the theory of neural networks, which will provide important
background to properly situate the contributions of this thesis. We begin by introducing
the celebrated Hopfield model, a simple model of associative memory that can be studied using
the tools of statistical mechanics. We discuss some extensions of the model, and summarize some
results about the effects of asymmetry in the synaptic weights. Then, we review some 
works studying the onset of chaos in recurrent neural networks, focusing on models that are similar
to the one studied in this thesis. We conclude by discussing some existing approaches for learning
in recurrent networks that exhibit chaotic spontaneous activity.

Finally, we review the literature on biologically plausible learning in artificial neural networks (ANNs).
Since a lot of work in this area has been motivated by the search for alternatives to the backpropagation algorithm,
we begin by recalling its formulation and discussing why it is considered implausible as a model of learning in the cortex.
Then, we provide an overview of the most prominent existing alternatives, focusing on algorithms that are most closely
related to the one proposed in this thesis.

\section{Statistical Mechanics of Learning}

\subsection{The Statistical Mechanics Formalism} \label{sec:stat_phys}

Statistical mechanics is a powerful framework for understanding the collective behavior of large systems, and how it emerges from the
microscopic interactions between their components. It was originally developed by physicists to study thermodynamics, but it has since
found applications in many fields beyond physics, including, for example, chemistry \cite{stadlerGeneralizedTopologicalSpaces2002, wuAdaptationProteinFitness2016, mauriTransitionPathsPottslike2023}, 
biology \cite{macadangdangAcceleratedEvolutionDiversityGenerating2022, greenburyStructureGenotypephenotypeMaps2022, papkouRuggedEasilyNavigable2023}, 
and computer science \cite{mezardAnalyticAlgorithmicSolution2002, donohoMessagepassingAlgorithmsCompressed2009, addario-berryAlgorithmicHardnessThreshold2020, gamarnikDisorderedSystemsInsights2022}.
A thorough introduction to statistical mechanics and its applications to study computational problems is beyond the scope of this thesis;
we refer the interested reader to excellent resources \cite{martinStatisticalMechanicsMethods2001, bordelonReplicaMethodMachine} and textbooks \cite{mezardInformationPhysicsComputation2009, mezardSpinGlassTheory1986, engelStatisticalMechanicsLearning2001} on the subject.
Nevertheless, in this section we introduce some key concepts and vocabulary that will be useful to understand the theoretical contributions of this thesis.

\subsubsection{The Boltzmann Distribution}

Consider a system with \(N\) microscopic degrees of freedom, whose state can be described as a point \(x\) in an abstract configuration
space \(\mathcal X\). We want to provide a statistical description of the system, by defining a suitable probability measure over
\(\mathcal X\) and studying its properties. Historically, in the study of thermodynamic systems, this was necessary due
to the infeasibility of tracking the evolution of each microscopic degree of freedom in very large systems. The idea underlying this
simplification is that the details of the microscopic dynamics should not be essential to describe the macroscopic properties of the system.
In fact, starting from such a probabilistic description, one can often recover determinism at the macroscopic level, because
fluctuations tend to average out.

From here on, we are going to take a purely mathematical perspective on the subject. We need to define a probability distribution \(p(x)\)
over \(\mathcal X\), and in doing so we want to avoid assuming any more structure than we need to. A way to characterize
this requirement mathematically is by seeking a distribution that maximizes the Shannon entropy, compatibly with any constraints we
might have on the system. Assume that all constraints can be formulated as fixing the expectation values of a set of observables
\(O_i : \mathcal X \to \mathbb R\):
\begin{equation}
  \int_{\mathcal X} p(x) \, O_i(x) \, dx = \langle O_i \rangle
\end{equation}
Then, the distribution that maximizes the Shannon entropy is the Boltzmann distribution:
\begin{equation}
  p(x) = \frac{1}{Z} \exp \left(-\sum_i \beta_i \, O_i(x) \right) \quad\quad Z = \int_{\mathcal X} \exp \left(-\sum_i \beta_i \, O_i(x) \right) dx,
\end{equation}
where \(Z\) is a normalization constant, called the \emph{partition function}, and \(\beta_i\) are Lagrange multipliers that should be
chosen to satisfy the constraints on the first moments of the observables \(O_i\).

In many cases of interest, we consider a single observable, the \emph{energy} \(E(x)\), and we call the associated
Lagrange multiplier \(\beta\) the \emph{inverse temperature}. For example, in constraint satisfaction problems, the energy
function is often the number of unsatisfied constraints, while in optimization problems it can be the cost function. We consider two
limits: as \(\beta \to 0\), the Boltzmann distribution becomes uniform over \(\mathcal X\); as \(\beta \to \infty\), it concentrates
(uniformly) on the global minima of the energy function. Intermediate values of \(\beta\) focus the distribution on states at
different energy levels. Thus, studying the Boltzmann distribution allows to probe the structure of the energy landscape at
different heights.

\subsubsection{The Partition Function}

The partition function \(Z\), and the associated \emph{free energy} \(F = -\beta^{-1} \ln Z\), are central objects in statistical mechanics.
For example, an analytic expression for the partition function allows to compute the expectation of the observables \(O_i\) as:
\begin{equation}
  \langle O_i \rangle \;=\; \int_{\mathcal X} p(x)\,O_i(x)\,dx
  \;=\; -\frac{\partial}{\partial \beta_i}\log Z\,.
\end{equation}
In fact,
\begin{align}
  - \frac{\partial}{\partial \beta_i} \log Z = - \frac{1}{Z}\,\frac{\partial}{\partial \beta_i} \int_{\mathcal X} dx \, \exp \left(-\sum_i \beta_i \, O_i(x) \right) = \\
  = \int_{\mathcal X} dx \, \frac{1}{Z} \exp \left(-\sum_j \beta_j \, O_j(x) \right) O_i(x) = \langle O_i \rangle
\end{align}
Similarly, the variance of the observable \(O_i\) can be computed as:
\begin{equation}
  \langle O_i^2 \rangle - \langle O_i \rangle^2 \;=\; \frac{\partial^2}{\partial \beta_i^2} \log Z
\end{equation}
And the Shannon entropy \(S\) of the distribution \(p(x)\) can be computed as:
\begin{equation}
  S = \log Z + \sum_i \beta_i \langle O_i \rangle = \log Z - \sum_i \beta_i \frac{\partial}{\partial \beta_i} \log Z
\end{equation}
Unfortunately, computing the partition function analytically is a hard task in general, because it requires summing over exponentially
many configurations. This can be done in closed form in some models, or it can be computed numerically with algorithms like
Belief Propagation.

\subsubsection{Phase Transitions and Concentration of Measure}

In statistical mechanics, we often consider systems with a large number of degrees of freedom: we take the so-called \emph{thermodynamic limit} \(N\!\to\!\infty\).
This has radical consequences. First of all, as we briefly discussed, in the limit of a large system we can recover macroscopic determinism. This is because, in many situations,
measure concentrates and the fluctuations of observables become negligible as the number of degrees of freedom increases. This often happens, for example, for the energy, which is
normalized to be \emph{extensive} (i.e. linear in \(N\)): usually, both the mean and the variance of the energy are \(\mathcal O(N)\), and relative fluctuations
shrink like \(1/\sqrt{N}\).

Another important consequence of the thermodynamic limit is the emergence of \emph{phase transitions}, abrupt changes in the macroscopic properties of a system
as an \emph{order parameter} is varied. Phase transitions are characterized by non-analytic behavior of the free energy \(F\), and thus they cannot take place in finite
systems. Each phase is characterized by a different microscopic organization, which is reflected in the macroscopic properties of the system. 
Understanding the \emph{phase diagram} of a system is therefore crucial to understand its behavior. For example, in the theory of average-case complexity, 
it was shown that the random k-SAT problem exhibits different phases, characterized by the arrangement of the 
solutions into one, few, or many connected clusters. Polynomial-time algorithms seem to exist only when the
system presents so-called 'unfrozen' clusters: extensive and connected clusters of solutions in which at most a sub-linear number of variables 
have a fixed value \cite{mezardAnalyticAlgorithmicSolution2002}.

\subsubsection{Disorder and Self-Averaging Quantities}

In many problems, especially in computer science, there is a source of randomness in the definition of the energy function, which is often called \emph{quenched disorder}.
This shouldn't be confused with the \emph{thermal disorder} that arises from the Boltzmann distribution. In problems with quenched disorder, if no simplifications took place,
we would need to study each individual realization of the disorder separately. Luckily, in the thermodynamic limit, some observables tend to heavily concentrate around their
average value, implying that we can describe the typical behavior of the system by simply averaging over the disorder. These observables are called \emph{self-averaging}:
specifically, we require that there exists a function \(\epsilon(N)\) with \(\epsilon(N) \to 0\) as \(N \to \infty\), such that the observable is within \(\epsilon(N)\) 
of its average with probability \(1 - \epsilon(N)\).

Usually the partition function is not self-averaging, because it fluctuates exponentially with \(N\). Therefore, the free energy obtained
by taking the logarithm of its disorder average, called \emph{annealed free energy}, does not in general capture the typical behavior of the system.
However, the logarithm of the partition function is an extensive quantity and therefore, in the thermodynamic limit, it tends to be
self-averaging. If we can compute the disorder average of the logarithm of the partition function, we obtain a quantity, called the \emph{quenched free energy},
which well describes the typical behavior of the system in many cases of interest, in the limit \(N \to \infty\). Note that, by Jensen's inequality, we have:
\begin{equation}
  f_{ann} = - \frac{1}{\beta N} \log \langle Z \rangle \leq - \frac{1}{\beta N} \langle \log Z \rangle = f
\end{equation}

Unfortunately, computing the quenched free energy is a much harder task than computing the annealed one. Doing
so requires a non-rigorous yet remarkably successful technique called the \emph{replica method}, which we will review in the next section.

\subsection{The Replica Method}

The \emph{replica method} is a non-rigorous yet remarkably successful technique for analyzing random high-dimensional optimization
and inference problems. It allows to compute the quenched free energy of a system with quenched disorder, by exploiting the so-called \emph{replica trick}:
\begin{equation}
\label{eq:replica_trick}
  \langle \log Z \rangle
  = \lim_{n\to 0}\frac{\langle Z^n\rangle-1}{n}
  = \lim_{n\to 0}\frac{1}{n}\log \langle Z^n\rangle,
\end{equation}
The quantities \(\langle Z^n \rangle\) are not any easier to compute than \(\langle \log Z \rangle\), unless \(n\) is an integer:
in that case, \(\langle Z^n \rangle\) can be computed as the product of \(n\) \emph{replicas} of the original partition function \(Z\).
This is a laborious task, but it is often feasible. Once \(\langle Z^n \rangle\) has been computed for all integer \(n\), 
the limit \(n \to 0\) in \eqref{eq:replica_trick} can be performed by means of an analytic continuation.

A canonical replica calculation proceeds through four phases:

\begin{enumerate}
  \item \textbf{Introducing replicas and Disorder average:} Express \(\langle Z^n\rangle\) as the disorder average of the product of \(n\) replicas of the partition function. Use field-theoretic techniques, like the Hubbard-Stratonovich transformation and the Fourier representation of the delta function, to decouple the disorder variables in the expectation, and average out the disorder. Typically, this comes at the cost of having to introduce a set of auxiliary integration variables and their Fourier conjugates. It also often produces an expression where the replicas of the system are coupled with each other.
  \item \textbf{Introduce Order parameters:} Define a set of order parameters (often, some form of weighted overlaps between the states of the replicas) which capture the dependency of \(\langle Z^n\rangle\) on the states of the replicas. Use delta functions to enforce their definition, introducing integration variables for the order parameters and their conjugates. Typically, after introducing order parameters, the components of the states of the replicas become decoupled and the integral can be factorized over them.
  \item \textbf{Saddle-point method:} The result of the previous step is an expression for \(\langle Z^n\rangle\) as an integral over the order parameters. The integrand is typically an exponential of the form \(\exp(-N\mathcal F)\), where \(\mathcal F\) is a function of the order parameters. In the thermodynamic limit, the integral can be evaluated by the steepest-descent method. This yields a set of saddle-point equations for the order parameters.
  \item \textbf{Ansatz and analytic continuation:} The saddle-point equations are typically nonlinear and coupled, thus they are difficult to solve. To reduce the complexity of the problem, we impose a plausible structure on the order parameters by making an ansatz. Then, we can perform the analytic continuation \(n \to 0\), solve the saddle-point equations, and compute the free energy associated with the saddle-point solution.
\end{enumerate}

There are some subtleties and mathematical caveats in this procedure. For example, in principle, the order of the limits \(N \to \infty\) and \(n \to 0\) is not guaranteed to commute.
Also, the analytic continuation \(n \to 0\) could be problematic in some cases. Putting these issues aside for now, at the end of the procedure we obtain a saddle-point solution for the order parameters.
While we cannot guarantee that, by making a more expressive ansatz, we would not find a different saddle point with a lower free energy, there are some checks that we can perform to at least corroborate
the results. First, we can check that the saddle-point solution is stable, by computing the eigenvalues of the Hessian of the action with respect to the order parameters (without ansatz!). If it isn't,
then we can be sure that a more expressive ansatz is needed. Second, we can check that the saddle-point solution leads to predictions that are consistent with physical constraints. Finally, we can compare
the predictions of the saddle-point solution with numerical simulations of the system.

We haven't yet made any recommendations on the choice of the ansatz. Consider the simple case where the order parameters are an \(n \times n\) matrix \(Q_{ab}\) of overlaps between the states of the replicas.
Typically, the first guess is \emph{replica symmetry} (RS): all off-diagonal overlaps are equal, $Q_{ab}=q$ for $a\!\neq\! b$.
If this ansatz is insufficient, then we need to break replica symmetry. There is a hierarchy of options to do so:
\begin{itemize}
  \item \textbf{1RSB} (one-step RSB): two overlap values, describing a clustered but not hierarchically nested state space.
  \item \textbf{Finite-$k$ RSB}: a hierarchical block structure with $k$ distinct overlap levels.
  \item \textbf{Full RSB}: an infinite hierarchy characterized by a continuous $q(x)$, giving
        the exact solution of paradigmatic models such as the Sherrington-Kirkpatrick (SK) spin glass \cite{parisiSequenceApproximatedSolutions1980}.
\end{itemize}
Physically, RS tends to match phases with a convex-like landscape where efficient algorithms often succeed, whereas RSB often corresponds
to rugged or glassy landscapes with exponentially many metastable states and algorithmic slow-downs.

Despite not being on rigorous mathematical grounds, there are several instances in which the predictions of the replica method
have been confirmed rigorously, most notably in the SK model \cite{guerraBrokenReplicaSymmetry2003, talagrandParisiFormula2006}. The method will be crucial for the theoretical analysis presented in Chapter \ref{ch:analytical}.

\subsection{Beyond the Equilibrium Description} \label{sec:local_entropy}

In the previous sections, we have presented the formalism of statistical mechanics as an analytical tool to probe the structure
of the energy landscape of a system. When facing an optimization problem, studying the Boltzmann distribution, with the energy
being the cost function, allows to gain considerable insight into the structure of the solution space, especially in the limit
\(\beta \to \infty\), which focuses the distribution on the global minima of the loss. However, this approach has an important limitation: it makes no distinction between different states, beyond their
energy. Except in some special cases like, for example, Simulated Annealing, which for long times and slow cooling schedules
approximately samples from the Boltzmann distribution, optimization algorithms have biases due to their dynamics that make
them more likely to sample certain types of solutions than others.

An important example of this phenomenon, with connections to average-case complexity and hardness in random optimization problems, is the so-called 
\emph{overlap gap property} (OGP) \cite{gamarnikOverlapGapProperty2021}. In simple terms, a random optimization
problem exhibits the OGP if, with high probability over the disorder realization, near-optimal configurations have a clustered structure, with the
diameter of each cluster being smaller than the distance between any pair of clusters. The onset of the OGP has been linked with
the onset of algorithmic hardness in many random structures including, for example, random k-SAT, the problem of finding cliques
in Erdős-Rényi graphs, and the problem of finding the ground states of spin glasses like the Sherrington-Kirkpatrick model.
For a large class of 'stable' algorithms exhibiting small sensitivity to perturbations of the problem instance, which includes Simulated Annealing, Stochastic
Gradient Descent, and Belief Propagation, to name a few, the presence of the OGP provably prevents them from finding near-optimal solutions \cite{gamarnikOverlapGapProperty2021}.

For this reason, to understand the behavior of optimization algorithms, we need to go beyond the equilibrium description of the
energy landscape, and introduce a bias towards the configurations that are accessible to algorithms and thus relevant for learning.
In this section, we review a recently developed analytical approach, first introduced in \cite{baldassiSubdominantDenseClusters2015} and then
extended in \cite{baldassiUnreasonableEffectivenessLearning2016, baldassiShapingLearningLandscape2020, baldassiUnveilingStructureWide2021},
that allows to enhance the statistical weight of (near-optimal) solutions surrounded by a large number of other solutions compared
to the equilibrium measure. For concreteness, we focus the exposition on the binary perceptron learning problem, a simple non-convex
constraint satisfaction problem. This framework allows to leverage the tools introduced in
the previous sections to detect the existence and characterize the properties of extensive and connected clusters of solutions,
which tend to be rare but accessible to algorithms and, in light of the OGP as well as a wealth of empirical evidence, turn out to be the
relevant ones for learning.

\subsubsection{Binary Perceptron Learning}

A binary perceptron with weights \(W \in \{\pm 1\}^N\) maps any \(N\)-dimensional input vector \(\boldsymbol{\xi} \in \{\pm 1\}^N\) to a label
\(\tau(W,\boldsymbol{\xi}) = \Sign (W \cdot \boldsymbol{\xi})\). Given a dataset of \(N\)-dimensional input patterns
\(\{\boldsymbol{\xi}^{\mu}\}_{\mu=1}^{\alpha N}\) and their corresponding labels \(\{\sigma^{\mu}\}_{\mu=1}^{\alpha N}\),
the perceptron learning problem consists in finding a vector of weights \(W^*\) such that \(\tau(W^*, \boldsymbol{\xi}^{\mu}) = \sigma^{\mu}\) 
for every \(\mu = 1, 2, \ldots, \alpha N\).

Two scenarios are usually studied: in the \emph{storage} setting, labels and patterns
are statistically independent, while in the \emph{teacher-student} setting, the labels are provided by an unknown teacher perceptron with the same architecture as the student.
Both cases are known to display a phase transition in the thermodynamic limit \(N \to \infty\) as \(\alpha = P/N\) crosses a critical threshold.
In the storage setting, at \(\alpha_c \approx 0.833\) there is a SAT/UNSAT transition where the probability that a random instance is solvable drops
from 1 to 0. In the teacher-student setting, below \(\alpha_{TS} \approx 1.245\) there are exponentially many solutions, while above \(\alpha_{TS}\) the only solution 
to the typical problem is the teacher itself \cite{baldassiSubdominantDenseClusters2015}. For concreteness, we focus here on the teacher-student setting, but all the results can be adapted to the storage case as well. \\

Despite it being hard in the worst case, there exist at least four algorithms which, at least outside of a window \(1.0 \le \alpha_c \le 1.5\) around the critical point,
can solve the perceptron learning problem in the teacher-student setting: Reinforced Belief Propagation \cite{braunsteinLearningMessagePassing2006}, Reinforced Max-Sum \cite{baldassiMaxSumAlgorithmTraining2015}, SBPI \cite{baldassiEfficientSupervisedLearning2007}, and CP+R \cite{baldassiGeneralizationLearningPerceptron2009}.
Theoretically, the problem can be analyzed using the replica method, revealing that typical solutions are isolated for all values
of \(\alpha\), and that the teacher is also isolated and indistinguishable from the other solutions (except for its generalization error). However, some
predictions of the analysis are not consistent with what is found empirically by sampling solutions with the algorithms mentioned above. In fact,
the solutions they sample tend to generalize better, and to be surrounded by a larger number of other solutions, compared to the predictions of the replica method.
This means that the algorithms cannot be sampling from the equilibrium measure: their dynamics favors certain types of solutions over others. \\

\subsubsection{The Local Entropy}

Motivated by the discrepancy between the equilibrium description of the landscape of
the binary perceptron and the behavior of algorithms, and inspired by the observation that solutions
sampled by algorithms tend to not be isolated, the authors of \cite{baldassiSubdominantDenseClusters2015} introduced
the following large deviation free energy density:
\begin{align}
\label{eq:local_entropy_free_energy}
\mathcal F(d,y) \;=\; -\frac{1}{N\,y}\;\log\Biggl(\sum_{\{\tilde{W}\}} 
   X_{\xi}(\tilde{W})\,\bigl[\mathcal N(\tilde{W},d)\bigr]^{y}\Biggr), \\
\mathcal N(\tilde{W},d) \;=\; \sum_{\{W\}} 
   X_{\xi}(W)\;\delta\!\bigl(W\!\cdot\!\tilde{W},\,N(1 - 2d)\bigr),
\end{align}
where \(X_{\xi}(W)\) is 1 if \(W\) is a solution and 0 otherwise.
This ensemble increases the importance of solutions surrounded by a large number of other solutions compared
to the equilibrium measure, and it is governed by a formal energy function given by:
\begin{equation}
  \varepsilon(\tilde{W}) \;=\; -\frac{1}{N}\;\log \mathcal N(\tilde{W},d) ,
\end{equation}
with the corresponding inverse temperature \(y\). Two key quantities are the \emph{local entropy}:
\begin{equation}
  S_I(d, y) = \frac{1}{N} \left\langle \log \mathcal N(\tilde{W},d) \right\rangle _{\tilde W, \, \xi}
\end{equation}
and the \emph{external entropy}:
\begin{equation}
  S_E(d, y) = - y \left( \mathcal F(d, y) + S_I(d, y) \right)
  \label{eq:external_entropy}
\end{equation}

The authors of \cite{baldassiSubdominantDenseClusters2015} have studied this free energy in the RS ansatz, which provides
self-consistent predictions up to a maximum inverse temperature \(y^*(\alpha, d)\) beyond which the external entropy
becomes negative.
We summarize here their key findings, since qualitatively similar phenomena are observed in several other models \cite{baldassiUnreasonableEffectivenessLearning2016, baldassiShapingLearningLandscape2020, baldassiUnveilingStructureWide2021}, including
the one considered later in this thesis. Results are reported at the inverse temperature \(y^*(\alpha, d)\):
\begin{itemize}
  \item Below \(\alpha_{TS}\), \(S_I(d)\) is positive in a neighborhood of \(d = 0\), signaling the existence of exponentially large clusters of solutions. Their size reduces as \(\alpha\) increases, and they disappear at \(\alpha_{TS}\).
  \item Around \(d = 0\), the curves \(S_I(d, \alpha)\) are essentially indistinguishable from \(S_I(d, 0)\), meaning that the clusters are extremely dense at their core. At large enough \(d\), \(S_I(d)\) collapses onto the equilibrium entropy.
  \item The \(S_I(d)\) curves are monotonic up to a critical \(\alpha_U \approx 1.1\), after which geometric bounds are violated, indicating that the RS ansatz is no longer valid. It is conjectured that this breakdown signals a fracture in the space of solutions, with the dense core becoming isolated and hard to find.
  \item The generalization error of solutions dominating the large-deviation ensemble is significantly lower than that of the typical equilibrium solutions.
\end{itemize}

The picture that emerges is the following: in the landscape of the binary perceptron, there exist rare, extensive clusters of solutions,
which are extremely dense at their core and accessible to algorithms despite being subdominant. The exponentially more numerous
equilibrium solutions, on the other hand, are isolated and thus hard to sample, despite dominating the equilibrium measure: they are not
the relevant ones for learning. There is increasing empirical and theoretical evidence 
that a similar phenomenology is relevant well beyond the binary perceptron \cite{baldassiSubdominantDenseClusters2015, baldassiUnreasonableEffectivenessLearning2016, gamarnikOverlapGapProperty2021, baldassiShapingLearningLandscape2020, baldassiUnveilingStructureWide2021, chaudhariEntropySGDBiasingGradient2017a, pittorinoEntropicGradientDescent2021}.
For example, the authors of \cite{baldassiShapingLearningLandscape2020} have adapted the local entropy analysis to shallow
(one- and two-layer) neural networks with continuous weights, showing that rare, extremely wide and flat minima coexist with
much more numerous narrow minima and critical points. Empirical studies have demonstrated, even in deeper networks, that these
minima possess superior generalization properties compared to the more numerous narrow minima \cite{jiangFantasticGeneralizationMeasures2019, pittorinoEntropicGradientDescent2021}.

% todo: mention parity machine (appendix E of \cite(baldassiShapingLearningLandscape2020) and the connection with local entropy (appendix F same paper)) (?)

\subsubsection{Optimizing the Local Entropy}

The large-deviation free energy density \(\mathcal F(d,y)\) can be softened by introducing Lagrange multipliers \(\beta\), \(\beta'\) and \(\lambda\)
to relax the energy and distance constraints. For the binary perceptron:
\begin{align}
  Z(\beta, \beta', \lambda, y) = \sum_{\tilde W} \exp\left(-\beta \, E(\tilde W) + y \, \phi(\tilde W, \lambda, \beta')\right) \\
  \phi(\tilde W, \lambda, \beta') = \log \sum_W \exp\left(-\beta' \, E(W) - \frac{\lambda}{2} \, d(W, \tilde W)\right)
\end{align}
As the inverse temperature \(y\) increases, the ensemble becomes more and more biased towards the minimizers of \(\phi(\cdot, \lambda, \beta')\),
a soft version of \(\mathcal N(\cdot, d)\) where the role of the distance \(d\) is played by the Lagrange multiplier \(\lambda\).

For integer values of \(y\), simple algebraic manipulations allow to rewrite the partition function as:
\begin{equation}
Z(\beta, \beta', \lambda, y) = \sum_{\tilde W, \{W^a\}} \exp\left(-\beta E(\tilde W) - \beta' \sum_{a=1}^y E(W^a) - \frac{\lambda}{2} \sum_{a=1}^y d(\tilde W, W^a)\right).
\label{eq:replicated_partition_function}
\end{equation}
This partition function describes a system of \(y + 1\) interacting \emph{real} replicas with an interaction mediated
by the reference configuration  \( \tilde W\). These replicas should not be confused with the \emph{virtual} replicas
introduced in the replica method: here we are describing a system that is literally made of \(y + 1\)
interacting replicas of the original system. \\

This representation is particularly useful because it allows to derive a general algorithmic scheme for optimization
problems: take the original system, replicate it \(y\) times, and allow each replica to optimize its own cost function
freely subject to the interaction with a reference replica \(\tilde W\). The inverse temperature \(\beta'\) is often
set to 0, so that the reference acts as a sort of barycenter for the replicas. In the limit of large \(y\), this procedure
will tend to sample from high local entropy regions. The value of \(\lambda\) can be decreased over time, effectively
exploring the energy landscape at increasingly finer resolution. In practice, already a few replicas (3-5) are enough
to observe significant benefits. Algorithms like replicated simulated annealing, replicated stochastic gradient descent, 
and replicated belief propagation can all be derived from this framework \cite{baldassiUnreasonableEffectivenessLearning2016}.

% todo: mention barbier paper 2025 (?)

\section{Attractors and Chaos in Neural Networks}

\subsection{Associative Memories and Hopfield-type Models}\label{sec:hopfield}

The Hopfield model \cite{hopfieldNeuralNetworksPhysical1982} is a simple model of associative memory that could be regarded as the
"fruit fly" or "hydrogen atom" of neural network theory \cite{hertzIntroductionTheoryNeural2018}, because it illustrates, in the
simplest possible setting, the way in which collective computation can emerge. Associative memories are tasked with storing
a set of patterns in memory in such a way that, presented with a partial or corrupted version of one of the patterns, the network
is able to correct errors and retrieve it. In this section, we review the fundamental concepts underlying the Hopfield model, as
well as some of its generalizations and recent developments. For a more thorough introduction to the topic, we refer the reader
to \cite{hertzIntroductionTheoryNeural2018}.

\subsubsection{The Hopfield Model}

In its original formulation, the Hopfield model is a recurrent neural network of \(N\) binary neurons \(x_i\!\in\!\{\pm 1\}\), 
which are able to store and retrieve patterns through Hebbian plasticity. Given a set of \(P\) patterns \(\{\boldsymbol{\xi}^{\mu}\}_{\mu=1}^{P}\),
with each \(\xi^\mu_i\) extracted independently and uniformly at random from \(\{\pm 1\}\), synaptic couplings \(J_{ij}\) between neurons are defined in such a way as to capture the correlations between the activations of pairs
of neurons across the dataset:
\begin{equation}
  J_{ij}\;=\;\frac{1}{P}\sum_{\mu=1}^{P}\xi^{\mu}_{i}\,\xi^{\mu}_{j}, \quad i \neq j.
  % \qquad J_{ii}=0,\;J_{ij}=J_{ji},
  \label{eq:Hebb}
\end{equation}
It is usually assumed that learning has already taken place, resulting in the symmetric synaptic efficacies given by \eqref{eq:Hebb},
and the network behavior is studied through the tools of statistical mechanics.

Many dynamical rules can be considered for updating the state of the network over time. A common choice is to use an asynchronous update rule 
which aligns neurons with the sign of their local field, according to:
\begin{equation}
  x_{k}(t+1)\;=\;\Sign\!\Bigl(\sum_{j}J_{kj}\,x_{j}(t)\Bigr)
  \label{eq:update}
\end{equation}
Because of the symmetry of the couplings, and under the assumption that \(J_{ii}=0\), it can be shown that the dynamics \eqref{eq:update} is
governed by a Lyapunov (energy) function, defined as:
\begin{equation}
  \mathcal{E}(\mathbf{x})\;=\;- \frac{1}{2} \sum_{i,j}J_{ij}x_i x_j
  \label{eq:energy}
\end{equation}
In fact, the change in energy due to the update of neuron \(k\) can easily be computed and equals:
\begin{equation}
  \Delta\mathcal{E} = -(\Sign h_k-x_k)h_k,
  \qquad h_k = \sum_{j\neq k}J_{kj}x_j
  \label{eq:delta_energy}
\end{equation}
It is straightforward to check that \(\Delta\mathcal{E}\le 0\), implying that the asynchronous dynamics
descends the energy landscape defined by \eqref{eq:energy}. If one is consistent with the behavior when the local field \(h_k\) is zero
(e.g. only flip when \(h_k\!\neq\!0\)), \eqref{eq:delta_energy} implies that the asynchronous dynamics \eqref{eq:update} avoids nontrivial 
cycles and converges asymptotically to a local minimum of the energy.

Once this is established, understanding the behavior of the Hopfield model as an associative memory reduces to the problem of
understanding the structure of the energy landscape defined by \eqref{eq:energy}. There are three main types of ground states of the energy, depending on the loading ratio \(\alpha\!=\!P/N\):
\begin{enumerate}
  \item \emph{Retrieval states}: these are strongly correlated to the stored patterns $\boldsymbol{\xi}^{\mu}$ (with a small, \(\alpha\)-dependent discrepancy of about 1-2\%).
  \item \emph{Spurious minima}: these correspond to finite linear combinations of an odd number of stored patterns $\boldsymbol{\xi}^{\mu}$ (or their mirror states $- \boldsymbol{\xi}^{\mu}$). That is, they are states \(\boldsymbol{\xi}\) of the form
  \(\xi_i = \Sign( \sum_{k = 1}^{2m + 1} \pm \, \xi^{\mu_k}_i )\), where \(m\) is an integer, \(\{\mu_k\}\) a set of pattern indices, and the choice of signs must be consistent across all neurons \(i\).
  \item \emph{Glassy states}: these do not correspond to any finite linear combination of stored patterns.
\end{enumerate}

This is the prototypical problem in the study of disordered systems, and it can be tackled analytically with the replica method.
In the thermodynamic limit \(N\!\to\!\infty\), the system undergoes a first‐order phase transition at a critical loading ratio
\(\alpha_c \approx 0.138\), sometimes referred to as "blackout catastrophe". Below this critical point, the retrieval states are stable attractors of the dynamics, and the system
behaves like an associative memory, recovering stored patterns if the dynamics is initialized sufficiently close to one of them. For \(\alpha > \alpha_c\),
instead, interference among the stored patterns destroys these stable states, leading to a proliferation of spurious minima and glassy states.
For this reason, it is said that the Hopfield model has a storing capacity of \(\alpha_c N\), scaling linearly with the system size \cite{amitStoringInfiniteNumbers1985}.

\subsubsection{Some Generalizations of the Hopfield Model}

The limited capacity of the original Hopfield model has motivated several generalizations. Among them, the dense associative memory
model \cite{krotovDenseAssociativeMemory2016} has achieved a capacity scaling polynomially in \(N\) by going beyond pairwise interactions.
They consider a generalized energy function of the form:
\begin{equation}
  \mathcal{E}(\mathbf{x}) = - \frac{1}{2} \sum_{\mu = 1}^P F(\sum_{i=1}^N \xi_i^\mu x_i)
\end{equation}
and consider an asynchronous update rule that updates the state of a neuron \(x_k\) in such a way as to minimize
the resulting energy after the update, i.e.:
\begin{align}
  x_k(t+1) = \Sign\Bigl( \sum_{\mu = 1}^P F(\xi^\mu_k + \sum_{i \neq k} \xi^\mu_i x_i(t)) - \sum_{\mu = 1}^P F(- \xi^\mu_k + \sum_{i \neq k} \xi^\mu_i x_i(t)) \Bigr)
\end{align}
By taking \(F(x) = x^n\) for \(n \ge 2\), the model can store a number of patterns scaling as \(N^{n-1}\) \cite{krotovDenseAssociativeMemory2016} by leveraging
\(n\)-way interactions between neurons. For \(n = 2\), we recover the original Hopfield model.
In \cite{demircigilModelAssociativeMemory2017}, the authors show that by taking \(F(x) = \exp(x)\), the model is able to store
an exponential number of patterns in \(N\), with basins of attraction almost as large as in the Hopfield model. A biological implementation of higher-order
interactions between neurons to achieve memory has been proposed in \cite{kozachkovNeuronAstrocyteAssociativeMemory2024}, but it remains speculative.

More recently, the authors of \cite{ramsauerHopfieldNetworksAll2021} have proposed a continuous variant of the dense associative memory model with
\(F(x) = \exp(x)\), equipped with an update rule that guarantees convergence to a stationary point of the energy. They show that this continuous
model retains the exponential capacity and is able to recover stored patterns in a single update step. Interestingly, their update rule has strong similarities
with the attention mechanism used in Transformers \cite{vaswaniAttentionAllYou2023}.

The above generalizations, and the whole notion of capacity, are grounded in an understanding of the Hopfield model that is
based on the equilibrium description of the energy landscape. While this is a natural starting point, it neglects the dynamical
aspects of retrieval. Very recently, the author of \cite{clarkTransientDynamicsAssociative2025} challenged this view by studying
the transient dynamics of dense associative memories using a Dynamic Mean Field Theory (DMFT) approach. They showed that, even
above capacity, stored memories can still be transiently retrieved thanks to the presence of slow regions in the energy landscape,
surviving traces of the fixed points that exist below capacity.

% todo: references for Hopfield from transient dynamics paper (?)

\subsubsection{Effects of Asymmetry in the Synapses}

The Hopfield model has been extremely influential in the field of neural networks. However, from a biological perspective, it
is a very crude model. The most problematic aspect is the definition of the synaptic couplings \(J_{ij}\) in \eqref{eq:Hebb}.
In fact, in addition to neglecting the learning process completely, it assumes that the synaptic connections are symmetric, i.e.
\(J_{ij} = J_{ji}\) for all \(i\) and \(j\). This assumption is important for the statistical
mechanics analysis, since it guarantees the existence of an energy function that is descended by the asynchronous dynamics. However,
it is not realistic: in the cortex, synapses are known to be one-way, with each neuron tending to form either excitatory or inhibitory
connections with its neighbors. This is the main motivation for considering asymmetry in the couplings, whose consequences we
now briefly review.

Without symmetry there is no Lyapunov function for the asynchronous dynamics, and the trajectory can exhibit complex behavior, including cycles and chaos \cite{sompolinskyChaosRandomNeural1988}.
In the Hopfield model, the results of \cite{crisantiDynamicsSpinSystems1987, feigelmanStatisticalPropertiesHopfield1986, feigelmanAugmentedModelsAssociative1987, singhFixedPointsHopfield1995} show
that the introduction of random asymmetry in the synaptic connections has the effect of destabilizing the glassy spurious attractors which characterize the symmetric Hopfield model.
Retrieval states, on the other hand, remain stable with moderate amounts of asymmetry. Asymmetry can be introduced, for example, by
considering a mixture of the Hebbian connectivity matrix in \eqref{eq:Hebb} with a random non-symmetric component. This could be thought to reflect a 
tabula-non-rasa scenario \cite{toulouseSpinGlassModel1986}, in which Hebbian learning takes place on top of an
existing random initial state. By studying the number of fixed points of such a model, the authors of \cite{singhFixedPointsHopfield1995}
found that there is a critical level of asymmetry, dependent on the ratio of patterns to neurons, after which the expected number of fixed points is 0.

It was also argued in \cite{parisiAsymmetricNeuralNetworks1986} that significant levels of asymmetry might serve a useful computational purpose,
in addition to being more realistic. The author proposed a synaptic plasticity rule that can be used to modify the synapses
of a Hopfield network while its dynamics is running. Assuming continuous time, it reads:
\begin{align}
  \frac{d J_{ij}(t)}{dt} &= \left[ - \lambda J_{ij}(t) + \bar S_i(t) \, \bar S_j(t) \right] f(\bar S_i(t) \, \bar S_j(t)), \\
  \bar S_i(t) &= \frac{1}{T} \int_0^T S_i(t - t') dt'
  \label{eq:parisi_rule}
\end{align}
Here, \(S_i(t)\) is the activity of neuron \(i\) at time \(t\), \(\bar S_i(t)\) is its average activity over a time window of length \(T\) in the immediate past, and \(f(x)\) is
a function that should be close to a step function with a discontinuity at a threshold \(x_t\). With this rule, the update of the
synaptic strengths only happens if the neurons show a strong correlation in their time-averaged activity, which can only
happen if they do not flip from one state to the other too quickly relative to \(T\). Assume that we had a network such that, when the initial
state does not lead to a retrieval state, the state of the network becomes time-dependent in a chaotic way. Then, memorization would
effectively only happen when retrieval is successful (reinforcing existing memories), or when the network is forced into
new patterns by external inputs (learning new memories), but not when the network is 'confused' and its activity is chaotic.
This view highlights a potential advantage, from the computational standpoint, of chaotic spontaneous activity in an associative
memory undergoing continuous learning with a rule like \eqref{eq:parisi_rule}. As was argued in \cite{parisiAsymmetricNeuralNetworks1986},
and later studied in more detail by \cite{sompolinskyChaosRandomNeural1988}, chaotic behavior is a signature of networks with asymmetric connectivity,
where the dynamics is not governed by an energy function.

We review some works studying the onset of chaos in random asymmetric neural networks in the next section.

\subsection{Chaos in Random Neural Networks}\label{sec:chaos_random}

\subsubsection{Random Asymmetric Neural Networks}

In a seminal paper, the authors of \cite{sompolinskyChaosRandomNeural1988} used a Dynamic Mean Field Theory approach to
analyze the emergence of chaos in a fully-connected rate-based recurrent neural network model with asymmetric random couplings.
They considered a network of \(N\) neurons with continuous states, evolving in time according to the following equation:
\begin{equation}
      \dot h_i=-h_i+\sum_{j}J_{ij}\phi(h_j),
      \label{eq:rate_network}
\end{equation}
with i.i.d. off-diagonal couplings \(J_{ij}\!\sim\!\mathcal N(0,J^{2}/N)\) and a sigmoidal transfer function \(\phi(x) = \tanh(gx)\).
Diagonal couplings are assumed to be 0. To perform the mean-field analysis, they approximated the second term in the RHS of \eqref{eq:rate_network} 
with a Gaussian distribution, whose time-dependent statistics are determined self-consistently. In the thermodynamic limit \(N\!\to\!\infty\), the
approximation becomes exact, and there appears a phase transition at a critical gain \(gJ = 1\): for \(gJ<1\) the system relaxes to the trivial fixed point \(h_i=0\),
while for \(gJ>1\) that fixed point becomes unstable and the system enters a chaotic phase, characterized by a positive maximal Lyapunov exponent \(\lambda_{\max}\). The transition 
is sharp in the thermodynamic limit, but numerical studies revealed the existence, for finite \(N\), of an intermediate regime where
the system exhibits long chaotic transients before settling into a stable state \cite{sompolinskyChaosRandomNeural1988}. \\

In \cite{kadmonTransitionChaosRandom2015}, these results were extended to two more realistic neural network models.
First, the authors considered rate-based networks composed of several subpopulations with randomly diluted connections. These
architectures were shown to exhibit the same transition to chaos as in the model of \cite{sompolinskyChaosRandomNeural1988},
under some conditions on the single-neuron transfer function.
In the same paper, the authors also showed that spiking neural networks 
with leaky integrate-and-fire neurons can undergo a transition to chaos, provided that the synaptic time constants are slow relative to the mean inverse firing rates \cite{kadmonTransitionChaosRandom2015}.
In this regime, as the gain crosses a critical threshold, the network switches from a phase of fast spiking fluctuations with constant rates to a phase where the firing rates exhibit chaotic fluctuations \cite{kadmonTransitionChaosRandom2015}. \\

Another notable development, which is related to the model studied in this thesis, is represented by \cite{sternDynamicsRandomNeural2014}. 
The authors introduced a variant of the rate-based model studied in \cite{sompolinskyChaosRandomNeural1988}, which includes
a self-coupling term in the dynamics of each neuron:
\begin{equation}
  \dot x_i=-x_i+s\,\phi(x_i)+g\sum_{j}J_{ij}\phi(x_j)
  \label{eq:stern_model}
\end{equation}
From the biological perspective, self-interacting units might represent clusters of neurons with strong local connectivity;
the random couplings \(J_{ij}\) would then reflect long-range interactions between different clusters.

A mean field analysis of this model in the thermodynamic limit reveals a rich phase diagram in the \((g,s)\) plane, with three macroscopic regimes:
\begin{enumerate}
  \item For \(s + g < 1\), the system trajectories decay to the trivial fixed point \(x_i=0\).
  \item For \(s < 1 + O(\log g)\) and \(s + g > 1\), the system exhibits persistent chaotic activity, consistently with the findings of \cite{sompolinskyChaosRandomNeural1988}. 
  \item For \(s > 1 + O(\log g)\), the system exhibits transient chaotic activity, which after a time growing exponentially with \(N\) converges to one of exponentially many stable fixed points.
\end{enumerate}

What is most striking is that, because of the interplay of the bi-stabilizing self-couplings with the random connectivity, there
exists a region in the \((g,s)\) plane where the system combines features most often seen independently, i.e. chaotic evolution
and multiple fixed point attractors. Along chaotic trajectories, especially for \(s > 1\) and moderate \(g\), the authors report
that the activity is characterized by relatively infrequent flips between fluctuations around the two solutions of \(-x + s \tanh(x) = 0\),
with a log-normal distribution of inter-flip times \cite{sternDynamicsRandomNeural2014}. The average inter-flip time increases with \(s\)
and decreases with \(g\).
It is also interesting to note that, while individual units become bistable for \(s>1\), network activations already show signs of bimodality
for \(s\) well below 1.

\subsubsection{Harnessing Chaos for Learning}
\label{harnessing_chaos}

Considering the results summarized in the previous section, it is natural to wonder what the implications are for learning of
chaotic spontaneous activity in recurrent neural networks. This question is even more interesting in light of the observation
that, in the brain, spontaneous activity tends to exhibit statistical and dynamical features typical of chaotic systems \cite{matteraChaoticRecurrentNeural2025}.
The dynamics of chaotic systems is rich and complex, which seems promising for learning, but it is also hard to control. In fact, the
strong irregularity in their activity, and their sensitive dependence on perturbations, make chaotic recurrent networks hard to train.
Furthermore, chaotic RNNs are particularly vulnerable to the exploding gradient problem when trained with backpropagation through time (BPTT) \cite{mikhaeilDifficultyLearningChaotic2022}.

Despite these challenges, there exist two main lines of research which have achieved some success in training chaotic RNNs.
Both directions have stemmed from reservoir computing (RC) \cite{jaegerHarnessingNonlinearityPredicting2004a, schrauwenOverviewReservoirComputing2007, tanakaRecentAdvancesPhysical2019},
a framework in which the rich (but usually not chaotic) dynamics of a recurrent neural network, called reservoir, is exploited to perform a task by training a linear readout layer
on top of its states. In RC, the most straightforward training approach is to simply perform a regression from the reservoir states to the target outputs.
In architectures where the reservoir receives feedback from the readout, it is common to clamp the readout to the true output during training,
a procedure known as teacher forcing \cite{jaegerHarnessingNonlinearityPredicting2004a}. To avoid causing the norm of the trainable
readout weights to explode, noise is often added to the activity of the reservoir when employing teacher forcing \cite{matteraChaoticRecurrentNeural2025}. \\

The first approach for learning in chaotic RNNs was initiated by \cite{sussilloGeneratingCoherentPatterns2009}, which introduced the First-Order Reduced and Controlled Error learning (FORCE) algorithm for supervised learning.
FORCE was initially developed for an architecture inspired by liquid- and echo-state networks \cite{buonomanoTemporalInformationTransformed1995, jaegerAdaptiveNonlinearSystem2002, jaegerAdaptiveNonlinearSystem2002}, with a
recurrent reservoir and a readout layer which also provides feedback to the reservoir. However, there are two key differences with respect to most RC approaches:
the reservoir is chosen to exhibit chaotic spontaneous activity, and teacher forcing and noise injection are avoided. While the richness
of the reservoir dynamics can be an asset, chaos during training should be kept under control. This is achieved in FORCE by
very strongly and rapidly adjusting the readout weights during the early stages of training. This way, the feedback on the
reservoir becomes strong enough to suppress chaos, and additionally it approximates teacher forcing. Furthermore, noise injection
can be avoided because, due to the small discrepancies between the readout state and the true output, the network is forced to sample
and learn to stabilize fluctuations in the reservoir activity. To achieve this, a number of optimization algorithms can be employed,
with the most successful being a recursive least-squares algorithm \cite{sussilloGeneratingCoherentPatterns2009}.

To evaluate the impact of chaotic spontaneous activity on learning performance, the authors of \cite{sussilloGeneratingCoherentPatterns2009} introduced a
gain parameter to control the spontaneous dynamics at initialization. They observed that, for values of the gain small enough that chaos can still be suppressed
by FORCE, larger values of the gain lead to faster and more accurate learning, as well as greater resilience to noise. This could be taken
as an indication that learning in RNNs happens optimally "at the edge of chaos", where spontaneous activity is rich but not overwhelmingly irregular \cite{sussilloGeneratingCoherentPatterns2009}.

While some follow-up works have improved on the original FORCE by remaining within the RC framework \cite{asabukiInteractiveReservoirComputing2018, zhengRFORCERobustLearning2025, tamuraTransferRLSMethodTransferFORCE2021, liCompositeFORCELearning2022},
the algorithm has proven to be flexible
enough that it could be adapted to learn the recurrent weights of the reservoir as well \cite{depasqualeFullFORCETargetbasedMethod2018, yinImprovingFullFORCEDynamical2021, tamuraPartialFORCEFastRobust2021, manneschiSpaRCeImprovedLearning2023}. This requires solving the credit assignment problem,
which is not trivial because there are no explicit targets for the reservoir states. One way to achieve this is to use a separate target network
to provide targets for the internal states of the reservoir. This is the approach taken, for example, in Full-FORCE, 
which has allowed to perform tasks with fewer recurrent units and greater robustness to noise than the original FORCE algorithm \cite{depasqualeFullFORCETargetbasedMethod2018}. \\

The second line of research stemmed from \cite{hoerzerEmergenceComplexComputational2014},
which introduced a reinforcement learning algorithm based on a three-factor Hebbian learning rule \cite{fremauxNeuromodulatedSpikeTimingDependentPlasticity2016},
called Reward-modulated Hebbian learning (RMHL). Also RMHL has been developed in a similar closed-loop architecture as FORCE \cite{hoerzerEmergenceComplexComputational2014},
using a spontaneously chaotic reservoir and strong readout feedback to tame chaos. The network undergoes stochastic perturbations of the reservoir
and readout states, and the output is evaluated at each step. Synaptic change happens according to a three-factor rule that tries
to adjust the weights in such a way as to replicate the outcomes of successful perturbations:
\begin{equation}
  \Delta W^{out} = \eta \, s(t) \, \left( z(t) - \bar z(t) \right) \, M(t).
\end{equation}
Here, \(s(t)\) is the instantaneous (pre-synaptic) activity of the reservoir, \(z(t)\) is the instantaneous (postsynaptic) 
activity of the readout, \(\bar z(t)\) is the time-averaged readout activity, and \(M(t)\) is a modulation signal that
equals 1 when the reward is higher than its recent average, and 0 otherwise.

\section{Biologically Plausible Learning in Artificial Neural Networks}

Backpropagation, an instance of reverse-mode automatic differentiation, has been a workhorse of the deep learning revolution. By
allowing to efficiently compute gradients on arbitrary computational graphs, it has provided a robust solution to the credit assignment
problem in deep neural networks.
For a long time, however, backpropagation has been considered implausible as a model of learning in the cortex, due to its necessity of
global coordination, long-range communication, and the symmetry of the forward and backward weights. This has motivated the
search for alternative, biologically plausible learning algorithms for Artificial Neural Networks (ANNs), which could serve as models
of learning in the cortex, enable learning with low-power neuromorphic hardware, or even inspire innovation in machine learning.

We begin this section by recalling the backpropagation algorithm, since most research in this area has developed in contrast to,
or directly as an alternative for, backpropagation. Then, we present a set of criteria for biological plausibility, and discuss what makes backpropagation
implausible. Finally, we provide an overview of the most prominent existing alternatives to gradient-based optimization for 
training ANNs, considering both feed-forward and recurrent architectures.

% Artificial Neural Networks (ANNs) trained with stochastic gradient descent have achieved unprecedented
% performance in a wide range of tasks, from image classification to natural language processing and
% robotics. These successes, largely driven by experimentation, have been enabled by the availability
% of large datasets, powerful hardware, and software libraries providing efficient implementations of automatic
% differentiation. Indeed, by far the most popular and successful approach to training ANNs is by
% employing first-order gradient-based methods to minimize a surrogate objective function. Backpropagation, 
% an instance of reverse-mode automatic differentiation, has been a workhorse of the deep learning revolution 
% by providing an efficient and flexible dynamic programming approach for computing the required gradients.
% However, for a long time the backpropagation algorithm has been considered implausible as a model of
% learning in biological neural networks, due to reasons including the necessity of long-range coordination
% and communication, and the symmetry of the forward and backward propagation weights. This has
% motivated the search for alternative, biologically plausible learning algorithms for ANNs, which could
% serve as models of learning in the cortex, enable learning with low-power neuromorphic hardware, 
% or even inspire further innovation in machine learning.

\subsection{Error Backpropagation and Biological Plausibility}

Backpropagation is an instance of reverse-mode automatic differentiation. For concreteness, we recall the algorithm description
in the simple case of a feed-forward multilayer perceptron performing supervised learning. However, the same principles can
be used to efficiently compute the gradient of any leaf node in a directed acyclic computational graph with respect to any
of its ancestors.

The algorithm consists of two distinct phases: a \emph{forward pass}, during which information is forward-propagated through the network
to compute the output and loss function, and neuron activations are cached; and a \emph{backward pass}, during which the error signal is back-propagated in accordance with the chain rule.

\paragraph{Forward pass.}
For an input--target pair $(\mathbf{x},\mathbf{y})$, the forward pass can be described as follows:
\begin{align}
\mathbf{a}^{0} &= \mathbf{x}, \\
\mathbf{z}^{\ell} &= \mathbf{W}^{\ell}\mathbf{a}^{\ell-1} + \mathbf{b}^{\ell}, && \ell = 1,\dots,L,\\
\mathbf{a}^{\ell} &= \sigma\!\bigl(\mathbf{z}^{\ell}\bigr), && \ell = 1,\dots,L-1, \\
\hat{\mathbf{y}} &= \mathbf{z}^{L},
\end{align}
where $\mathbf{W}^{\ell}\!\in\!\mathbb{R}^{m_\ell\times m_{\ell-1}}$, $\mathbf{b}^{\ell}\!\in\!\mathbb{R}^{m_\ell}$ and $\sigma(\cdot)$ is an element-wise non-linearity.  
Let the loss be $\mathcal{L}\bigl(\hat{\mathbf{y}},\mathbf{y}\bigr)$. During the forward pass, all intermediate values, i.e. all the values \(\mathbf{a}_\ell, \, \mathbf{z}_{\ell+1}\), \(\ell = 0, \dots, L-1\), are stored for use in the backward pass.
Define the shorthand \(f_\ell(\mathbf{a}^{\ell -1}) = \sigma \left( \mathbf{W}^{\ell}\mathbf{a}^{\ell-1} + \mathbf{b}^{\ell} \right)\) for \(\ell = 1,\dots,L-1\) and \(f_L(\mathbf{a}^{L-1}) = \mathbf{W}^{L}\mathbf{a}^{L-1} + \mathbf{b}^{L}\).

\paragraph{Backward pass.}
Denote with \( \boldsymbol{\delta}^{\ell} \) the gradient of the loss with respect to the pre-activations \(\mathbf{z}^{\ell}\) at layer \(\ell\). We
can define a recursion as follows:
\begin{align}
    \boldsymbol{\delta}^{L} 
    &= \nabla_{\mathbf{z}^{L}}\mathcal{L} \\
    \boldsymbol{\delta}^{\ell} 
    &= \bigl(\mathbf{W}^{\ell+1}\bigr)^{\!\top}\!\boldsymbol{\delta}^{\ell+1}\;\odot\;\sigma'\!\bigl(\mathbf{z}^{\ell}\bigr), \quad \ell=L-1,\dots,1
    \label{eq:backprop_backwards}
\end{align}
where $\odot$ denotes element-wise multiplication.
The gradients of the loss with respect to the network parameters can then be computed as follows:
\begin{align}
  \frac{\partial\mathcal{L}}{\partial\mathbf{W}^{\ell}} &= \boldsymbol{\delta}^{\ell}\,(\mathbf{a}^{\ell-1})^{\!\top}, \\
  \frac{\partial\mathcal{L}}{\partial\mathbf{b}^{\ell}} &= \boldsymbol{\delta}^{\ell}, \qquad \ell = 1,\dots,L.
\end{align}
This completes the description of the backpropagation algorithm for an MLP; the same principles apply to arbitrary directed acyclic 
computational graphs by traversing nodes in reverse topological order and accumulating gradients via the chain rule. \\

Now, we turn to the question of biological plausibility. Throughout this chapter, we will consider the following set of criteria, formalized in \cite{whittingtonTheoriesErrorBackPropagation2019},
to assess the biological plausibility of learning algorithms for ANNs:
\begin{itemize}
    \item \textbf{Locality of Dynamics}. Each neuron should perform computations based solely on the states of neighboring neurons, and on the strength of the synapses connecting them to it.
    \item \textbf{Locality of Plasticity}. The changes in synaptic efficacy should depend only on the activity of the pre- and post-synaptic neurons, and possibly on a global signal (neuromodulators).
    \item \textbf{Minimality of External Control}. There should be as little external control, routing information in different ways at different times, as possible.
    \item \textbf{Plausibility of Neural Architecture}. The patterns of connectivity between neurons should be compatible with known constraints from connectomics.
\end{itemize}
With these criteria in mind, it is clear that the classical formulation of backpropagation as described above is biologically implausible.
There are several reasons for this; we summarize the most compelling arguments below, and we refer the reader to \cite{whittingtonTheoriesErrorBackPropagation2019} and \cite{lillicrapBackpropagationBrain2020}
for a more detailed discussion.
\begin{itemize}
    \item \textbf{Non-locality of synaptic plasticity}. The synaptic plasticity rule in backpropagation depends on the activity of all downstream neurons, which conflicts with
    evidence suggesting that plasticity in the cortex depends only on pre- and post-synaptic activity, possibly modulated by a global neuromodulator \cite{pawlakTimingNotEverything2010}.
    \item \textbf{Global coordination}. Backpropagation requires two distinct phases, and the ability to store activation values during the forward pass for use in the backward pass.
    This requires the network to exhibit global coordination and synchrony, which is in stark contrast with the asynchronous, continuously operating nature of the cortex \cite{crickRecentExcitementNeural1989}.
    \item \textbf{Weight transport problem}. Backpropagation requires the feedback pathway to use the exact transpose of the forward synaptic weights. However, despite bidirectional connections being significantly more common in cortex than expected by chance,
    they are not always present and, even when they are, the forward and backward weights are not precisely matched in strength or sign \cite{HighlyNonrandomFeatures}.
\end{itemize}

These issues highlight the implausibility of backpropagation as a credit assignment mechanism in the cortex, at least in its classical formulation.
Motivated by this, in the next section, we will provide an overview of the most prominent algorithms that
seek to provide biologically plausible alternatives to backpropagation. 

\subsection{Overview of Existing Approaches}

Biologically inspired learning algorithms for ANNs can be organized, at a high level, into two classes, based on their
attitude towards backpropagation (BP). The first group explicitly attempts to approximate the behavior of BP.
These algorithms retain the goal of achieving credit assignment by propagating an error signal throughout the network, but they replace the biologically problematic ingredients
with a local and distributed dynamics, or with feedback pathways designed to be compatible with known cortical constraints.
Among these, some predominant approaches are Target Propagation, Feedback Alignment, Equilibrium Propagation, and Predictive Coding.
In the second group, there are methods that propose a different approach for credit assignment, without explicitly trying to 
follow the gradient of the objective function. We now review each approach.

\paragraph{Target Propagation} \cite{bengioHowAutoEncodersCould2014, zhangTargetPropagation2015, leeDifferenceTargetPropagation2015, meulemansTheoreticalFrameworkTarget2020, ernoultScalingDifferenceTarget2022}.

Target propagation is an algorithm for training feed-forward architectures that consists of a forward and a backward pass, like
BP. Instead of computing gradients in the backward pass, in each layer a learned approximate inverse map $g_\ell\!\approx\!f_\ell^{-1}$
sends a \emph{target activation} $t_{\ell-1}=g_{\ell-1}(t_\ell)$ to the previous layer, with \(t_L\) set to the target output \(\mathbf{y}\).

Each forward module $f_\ell$ learns to perform a local regression $x_{\ell-1}\!\to\!t_\ell$, while \(g_\ell\) can be 
parametrized as an autoencoder and be learned with a local loss \cite{bengioHowAutoEncodersCould2014}. This approach sidesteps
the weight-transport problem and makes the plasticity rule local. However, it introduces considerable additional complexity,
and still requires two distinct phases with a clear direction in the propagation of information.

An important variant is Difference Target Propagation (DTP) \cite{leeDifferenceTargetPropagation2015},
which uses the targets $t_{\ell-1}=x_{\ell-1}+g_{\ell-1}(t_\ell)-g_{\ell-1}(x_\ell)$. If the inverse mapping was perfect,
then this would reduce to vanilla target propagation, but this formula stabilizes the optimization considerably when the
inverse mapping is only approximate \cite{leeDifferenceTargetPropagation2015}. DTP is closely related with Gauss-Newton
optimization \cite{meulemansTheoreticalFrameworkTarget2020}. A variant of it, called Local Difference Reconstruction Loss (L-DRL), 
has been shown to approximate BP, and to achieve performance almost on par with it on benchmarks like Cifar10 and ImageNet 32x32 \cite{ernoultScalingDifferenceTarget2022}.
          
\paragraph{Feedback Alignment} \cite{liaoHowImportantWeight2016, noklandDirectFeedbackAlignment2016, xiaoBiologicallyplausibleLearningAlgorithms2018, launayDirectFeedbackAlignment2020, akroutDeepLearningWeight2020}.

Feedback Alignment (FA) was designed to train feed-forward architectures, and it is similar in spirit to Target Propagation. It uses a forward
and a backward pass like in BP, but it substitutes \((W^{\ell + 1})^T\) with a fixed random matrix \(B_\ell\) in \eqref{eq:backprop_backwards}.
Thus, weight updates use the pseudo-error $\delta_\ell=B_{\ell+1}\delta_{\ell+1}$. The authors of \cite{liaoHowImportantWeight2016} observed that
during learning the forward weights self-organize so that $W_\ell^{\!\top}B_{\ell+1}$ becomes partially aligned with the identity: despite not being
in the exact direction of the gradients, the updates still tend to decrease the loss.

From the biological perspective, FA brings similar improvements over BP as Target Propagation: plasticity is local,
and the weight transport problem is solved. However, it is much simpler, avoiding the complexity of parametrizing
and learning layer-wise approximate inverses.

Variants of this approach include Sign-Symmetry \cite{liaoHowImportantWeight2016}, that 
maintains the feedback weights equal to the sign of the transposed forward weights, and Direct Feedback Alignment \cite{noklandDirectFeedbackAlignment2016},
which connects the output layer to each hidden layer with a random feedback matrix directly. The Sign-Symmetry variant has been
shown to approach the performance of BP on ImageNet and MS COCO \cite{xiaoBiologicallyplausibleLearningAlgorithms2018}.

\paragraph{Equilibrium Propagation} \cite{bengioEarlyInferenceEnergyBased2016, scellierEquilibriumPropagationBridging2017, ernoultEquilibriumPropagationContinual2020}.

Equilibrium Propagation (EP) is a learning framework for energy-based models that approximately computes gradients of an objective function through its dynamics. The
algorithm requires that the underlying model converges to a steady state asymptotically. 
Let \(\theta\) denote the network parameters, \(s\) the network state, \(v\) the state of the external world, \(E(\theta, v, s)\) the
energy function. Given a cost function \(C(\theta, v, s)\), the objective is to minimize the cost of the fixed point produced
by the network dynamics:
\begin{align}
  \min_{\theta, s} \quad &C(\theta, v, s) \\
  s.t. \quad &\frac{\partial E}{\partial s}(\theta, v, s) = 0.
\end{align}

During training, each example \(v\) is presented to the network in two phases. In the free phase, the network dynamics is run until convergence to
a \emph{free fixed point} \(s^0_{\theta, v}\). During the nudged phase, the dynamics continues from the free fixed point until a \emph{nudged fixed point} \(s^\beta_{\theta, v}\), governed by:
\begin{equation}
  F(\theta, v, s, \beta) = E(\theta, v, s) + \beta C(\theta, v, s)
\end{equation}
Then, the update
\begin{equation}
  \Delta \theta \propto -\frac{1}{\beta} \left[ \frac{\partial F}{\partial \theta}(\theta, v, s^\beta_{\theta, v}, \beta) - \frac{\partial F}{\partial \theta}(\theta, v, s^0_{\theta, v}, 0) \right]
\end{equation}
recovers the gradient of the objective function in the limit of infinitesimal nudging \(\beta \to 0\). This is a kind of
contrastive update rule, in which the energy of the free fixed point is increased and the energy of the
weakly-clamped fixed point is decreased.

In practice, the authors of \cite{scellierEquilibriumPropagationBridging2017} consider a continuous Hopfield model with symmetric
couplings to ensure convergence to a fixed point. Their cost is a MSE loss between ground truth targets and some designated output
units. With these choices, the nudged phase makes an infinitesimal perturbation of the output units towards the true target,
and then "back-propagates" its influence through the dynamics. The synaptic plasticity instead becomes the sum of a Hebbian step
at the nudged equilibrium and an anti-Hebbian step at the free equilibrium.

From the biological point of view, Equilibrium Propagation is appealing because it uses a local plasticity rule with a Hebbian
and an anti-Hebbian contribution, and it propagates the error signal through the same dynamics used for inference. However,
the need for two separate phases, with a gentle nudge of the output units during the second, requires significant external
control which is hard to imagine in the cortex. Furthermore, the update rule, despite being local in space, is not local in time.
This last concern has been addressed in an extension of the EP approach, Continual Equilibrium Propagation \cite{ernoultEquilibriumPropagationContinual2020},
which restores the locality of plasticity in time by having neuron and synapse dynamics occur simultaneously throughout the second phase, while maintaining the guarantees
of convergence to the exact gradient.

In terms of performance, the original approach struggled to scale beyond toy datasets because the gradient estimate is
biased when the update is not infinitesimal. By introducing an additional negative nudging phase, pushing
away from the true target, and symmetrizing the finite difference, the authors of \cite{laborieuxScalingEquilibriumPropagation2020}
were able to drastically reduce this bias and approach BP performance on Cifar10. Another limitation is the long training and
inference time due to the need to simulate the dynamics, which cannot be parallelized across time. The authors of \cite{oconnorINITIALIZEDEQUILIBRIUMPROPAGATION2019}
partially addressed this, at the cost of increased complexity, by introducing an additional feed-forward network that learns to predict the asymptotic state of the
network via a local learning rule, and using it at inference time as a substitute for the dynamics.

\paragraph{Predictive Coding} \cite{raoPredictiveCodingVisual1999, whittingtonApproximationErrorBackpropagation2017, millidgePredictiveCodingApproximates2020, millidgePredictiveCodingTheoretical2022}.

\label{sec:predictive_coding}

Predictive Coding (PC) is a framework for learning in a special type of RNNs which exhibit a
hierarchical structure, with each layer attempting to predict the activity of the one below it,
and two types of neurons: to each \emph{value neurons} there is associated an \emph{error neuron}, 
which explicitly encodes the difference between the value neuron's state and its prediction
by the layer below it.

In the simplest case of MLP-like architectures, PC is derived assuming that the DAG defined
by the network topology underlies a probabilistic model, where, denoting with \(x_i^\ell\)
the state of the \(i\)-th value neuron at layer \(\ell\), we have:
\begin{equation}
  x_i^\ell | x^{\ell + 1} \sim \mathcal{N}(\sum_j J_{ij}^{\ell + 1} f(x_j^{\ell + 1}), \Sigma_i^\ell).
\end{equation}
Here, \(J_{ij}^{\ell + 1}\) is the synaptic weight from neuron \(j\) in layer \(\ell + 1\) to neuron \(i\) in layer \(\ell\),
and \(f(\cdot)\) is a non-linear activation function. The error neurons are defined as:
\begin{equation}
  \epsilon_i^\ell = \frac{x_i^\ell - \sum_j J_{ij}^{\ell + 1} f(x_j^{\ell + 1})}{\Sigma_i^\ell}.
\end{equation}
The neuronal dynamics is defined as gradient ascent on the log-likelihood of this Gaussian model:
\begin{equation}
  \dot x_i^\ell = - \epsilon_i^\ell + f'(x_i^\ell) \sum_j \epsilon_j^{\ell - 1} J_{ji}^\ell.
\end{equation}
During training, the input and output value neurons are clamped to their external values, and the network relaxes
until convergence (which is guaranteed due to concavity of the objective function). Then, the synaptic weights
are updated according to a Hebbian rule (which is also gradient ascent on the log-likelihood, but with respect to the synaptic weights):
\begin{equation}
  \Delta J_{ij}^\ell \propto \epsilon_i^{\ell-1} f(x_j^\ell).
\end{equation}
During inference, only the input value neurons are clamped, and the network is allowed to relax.
Under some conditions, the synaptic updates of PC have been shown to approximate the backpropagation updates \cite{whittingtonApproximationErrorBackpropagation2017}.

The seemingly bizarre alternation between gradient ascent steps on the same objective function with
respect to neurons and synapses has a natural interpretation in the variational inference framework \cite{millidgePredictiveCodingTheoretical2022a}.
In fact, one can take an abstract view of PC as variational inference in a Gaussian probabilistic model, and in that setting
the PC dynamics can be interpreted as performing a form of Expectation Maximization (EM) on the variational free energy, by taking
gradient descent steps \cite{whittingtonApproximationErrorBackpropagation2017, millidgePredictiveCodingTheoretical2022a}. The
Gaussianity assumption allows to recover the neuronal instantiation with value and error neurons, and the interpretation of PC
as minimization of local prediction errors \cite{millidgePredictiveCodingTheoretical2022a}.

The variational inference perspective on PC has allowed a generalization of the method to an arbitrary directed acyclic
computational graph: in this more general case, the probabilistic model has one variable for each vertex of the graph,
and the distribution factorizes according to the DAG underlying the computational graph \cite{millidgePredictiveCodingApproximates2020}.
Again, the model should be Gaussian to recover the neuronal instantiation described above. Also in this more general case, 
it has been shown that the PC dynamics can approximate backpropagation under appropriate conditions.
With the flexibility afforded by an arbitrary computational graph, the authors of \cite{millidgePredictiveCodingApproximates2020} 
have shown that PC can be used to train a range of deep learning architectures, including CNNs, RNNs, and LSTMs,
achieving performance on par with backpropagation on benchmarks like Cifar10 and Cifar100.

Predictive Coding is arguably the most mature among the existing approaches for biologically plausible learning 
in ANNs. It proposes a unifying framework to understand cortical function based on the minimization of local prediction errors,
with premises that are compatible with cortical anatomy \cite{raoPredictiveCodingVisual1999, millidgePredictiveCodingTheoretical2022},
a local and Hebbian plasticity rule, and a single dynamics for both inference and learning. Among its weaknesses, there is
the need for symmetric interactions between layers, and the elevated computational cost of inference and training
due to the need to run the dynamics until convergence.

\paragraph{Methods that do not approximate Backpropagation}

A second group of models departs more radically from backpropagation, exploring alternative credit assignment mechanisms that do not attempt to approximate the gradients of a global loss function.
Among these, some notable examples are:
\begin{itemize}
    \item \textbf{Three-factor Learning} \cite{kosslynFrontiersCognitiveNeuroscience1995, mazzoniMoreBiologicallyPlausible1991, williamsSimpleStatisticalGradientfollowing1992, unnikrishnanAlopexCorrelationBasedLearning1994, hoerzerEmergenceComplexComputational2014, ThreeFactorLearningSpiking}.
          In addition to considering the pre- and post-synaptic activity, these approaches use a global neuromodulatory signal to gate local plasticity. Early approaches
          introduced randomness in the behavior of synapses or neurons, and use correlations with a global error signal to do reinforcement learning \cite{kosslynFrontiersCognitiveNeuroscience1995, mazzoniMoreBiologicallyPlausible1991, williamsSimpleStatisticalGradientfollowing1992, unnikrishnanAlopexCorrelationBasedLearning1994}.
          However, it has been shown that learning in these models requires very small learning rates and does not scale well with model size \cite{werfelLearningCurvesStochastic2003}. in the interest of further
          biological realism, there have been attempts to extend three-factor learning to spiking neural networks, using Spike Timing Dependent Plasticity (STDP) and a global neuromodulator to guide plasticity \cite{williamsSimpleStatisticalGradientfollowing1992, unnikrishnanAlopexCorrelationBasedLearning1994, ThreeFactorLearningSpiking},
          but these approaches still face challenges with scalability and are very sensitive to hyperparameters \cite{ThreeFactorLearningSpiking}.
          Also the RMHL and related approaches, introduced in Section \ref{harnessing_chaos}, use a three factor learning rule for plasticity. The strength of these approaches is their extreme
          biological realism compared to the alternatives, especially in their spiking neural network variants, and their ability to learn from very limited supervision. However,
          in general they tend to struggle to keep up with BP in complex tasks, since credit assignment based on a scalar reward is much noisier than what gradients afford \cite{ThreeFactorLearningSpiking}.

    \item \textbf{FORCE algorithm} \cite{sussilloGeneratingCoherentPatterns2009, depasqualeFullFORCETargetbasedMethod2018}
          FORCE and related algorithms are very interesting from the biological perspective because, unlike most other methods,
          they are able to achieve learning in RNNs exhibiting spontaneous chaotic activity, by employing local synaptic plasticity
          rules. As is argued in \cite{sussilloGeneratingCoherentPatterns2009}, these models provide an avenue to explore the idea that
          brain function might arise from the reorganization of spontaneous activity. For an overview of the FORCE algorithm, please
          see Section \ref{harnessing_chaos}.

    \item \textbf{Forward-Forward algorithm} \cite{hintonForwardForwardAlgorithmPreliminary2022, gandhiExtendingForwardForward2023, ororbiaPredictiveForwardForwardAlgorithm2023}.
          The forward-forward algorithm is a recent proposal that replaces the backward pass in backpropagation with a second forward pass, done with negative examples.
          It has similarities with the Contrastive Divergence algorithm used to train Boltzmann machines, but replaces the intractable free energy with a simple local objective.
          Each layer in a feed-forward architecture is fed the normalized activations of the previous layer, and attempts to separate positive and negative examples by means of 
          a goodness function which could be, for instance, the norm of its activations. The maximization of the difference in goodness between the two phases provides a simple,
          local update rule for the weights. The approach has shown promise on simple tasks, and can be extended to spiking neural networks \cite{ghaderBackpropagationfreeSpikingNeural2025},
          but there is still an open question about scalability.

    \item \textbf{No-Prop algorithm} \cite{liNoPropTrainingNeural2025}.
          The no-prop algorithm is a recent proposal which dispenses with forward and backward passes entirely. Instead,
          they propose that each hidden layer be interpreted as an intermediate step in a denoising diffusion process, with an objective
          function derived from the ELBO that can be optimized independently for each layer. At training time, each layer is fed a noisy
          label at a different noise level, together with the input. Hidden layers learn to denoise the noisy label, while the final one
          minimizes a cross-entropy loss. At inference time, each layer is used to denoise the noisy label output by the previous one,
          starting from white noise.
\end{itemize}

% todo: dendritic error models (insieme a PC?) (?)

\chapter{Dense Clusters of Fixed Points in a Random RNN}
\label{ch:analytical}
In this chapter, we present novel theoretical results on the phase diagram of a random asymmetric neural network similar to the one
studied in \cite{sompolinskyChaosRandomNeural1988}. Differently from their model, we consider a network of binary neurons, and we
introduce a self-coupling term in the spirit of \cite{sternDynamicsRandomNeural2014}. Our analysis employs tools from the statistical
mechanics of disordered systems to study the number and the structure of the fixed points of the network, as a function of the self-coupling
strength. We first use the replica method in the Replica-Symmetric (RS) ansatz to derive the entropy of fixed points as a function of 
the self-coupling strength. Similar to the continuous case studied in \cite{sternDynamicsRandomNeural2014}, we find an exponential
number of fixed points appearing for positive values of the self-coupling strength. Then, we use the 1RSB formalism to
perform a large-deviation analysis that enhances the statistical weight of fixed points surrounded by a large number of other fixed
points, using techniques first introduced in \cite{baldassiSubdominantDenseClusters2015} and \cite{baldassiUnreasonableEffectivenessLearning2016}.
This way, we find evidence of a phase transition as the self-coupling strength crosses a critical value, beyond which large, connected clusters
of fixed points, which are rare and thus subdominant in the RS analysis, appear. The existence of similar structures has been found to be very relevant
for learning in constraint satisfaction problems \cite{baldassiSubdominantDenseClusters2015, baldassiUnreasonableEffectivenessLearning2016} as
well as in neural networks \cite{baldassiShapingLearningLandscape2020, baldassiUnveilingStructureWide2021}, since they tend to be accessible
to local search algorithms \cite{gamarnikOverlapGapProperty2021}. We will investigate their implications for learning in the next chapter.

\section{Model Definition}

% 1. Neuron states
We consider a network of \(N\) coupled binary neurons with asymmetric interactions:
\begin{equation}
    s_i \in \{\pm1\}, \quad i = 1,\dots,N.
    \label{eq:states}
\end{equation}

% 2. Synaptic weights
The synaptic weight matrix \(J\in\mathbb{R}^{N\times N}\) has independent standard Gaussian off-diagonal entries, and a uniform self-coupling \(J_D\) which is a control parameter:
\begin{equation}
    J_{ij} \sim\mathcal{N}\!\Bigl(0,\tfrac1N\Bigr), \; J_{ii} = J_D; \quad i,j = 1,\dots,N, \quad i\neq j.
    \label{eq:weights}
\end{equation}

% 3. Energy function
We define an energy function \(E_J(\mathbf{s})\) that counts the number of neurons that are not aligned with their local field:
\begin{equation}
    E_J(\mathbf{s})
    =\sum_{i=1}^{N}\Bigl[\,1 - s_i\,\Sign\!\Bigl(J_D\,s_i + \sum_{j\neq i}J_{ij}\,s_j\Bigr)\Bigr],
\end{equation}
This can be interpreted as the number of violated fixed-point constraints, for the dynamics that tends
to align the state of each neuron with its local field. Note that when \(J_D \to \infty\), the self-coupling
makes all network states fixed points.

% 4. Partition function
At inverse temperature \(\beta\), the partition function is
\begin{equation}
    Z_J = \sum_{\{s_i\}}
       \exp\!\bigl(-\beta\,E_J(\mathbf{s})\bigr)
     = e^{-\beta N}
       \sum_{\{s_i\}}
       \exp\!\Bigl(\beta\sum_{i=1}^N
         \Sign\!\bigl(J_D + s_i\textstyle\sum_{j\neq i}J_{ij}\,s_j\bigr)\Bigr).
    \label{eq:partition}
\end{equation}
Here, the summation is understood to be over all configurations of \(s\). We are going to study 
the phase diagram of this model as a function of \(J_D\) and \(\beta\), investigating
the existence, entropy, and structure of the fixed points (zero-energy states).

\section{Replica Formalism}

\subsection{Disorder Average}

% 1. Fourier representation of the delta–constraint
We introduce the local field \(h_i = \sum_{j\neq i}J_{ij}s_is_j\) and its conjugate \(\hat h_i\) using the identity
\begin{equation}
    1 = \int_{-\infty}^{\infty} \!d h_i\;
    \delta\!\Bigl(h_i - \sum_{j\neq i}J_{ij}s_is_j\Bigr)
    = \int_{-\infty}^{\infty}\!d h_i \int_{-i\infty}^{i\infty} \frac{d\hat h_i}{2\pi i}\,
    \exp\!\Bigl(- \hat h_i h_i + \hat h_i \sum_{j\neq i}J_{ij}s_is_j\Bigr).
    \label{eq:delta_identity}
\end{equation}

% 2. Integral representation of the partition function
Substituting \eqref{eq:delta_identity} into the definition \eqref{eq:partition} of \(Z_J\) gives
\begin{align}
    Z_J = e^{-\beta N}
    \int \prod_{i=1}^N \frac{d h_i\,d\hat h_i}{2\pi i}
    &\exp\Bigl[\, - \sum_{i=1}^N\hat h_i\,h_i + \beta \sum_{i=1}^N \Sign\Bigl( J_D + h_i \Bigr) \Bigr] \nonumber \\
        \times  \sum_{\{s_i\}} &\exp \Bigl( \, \sum_{i=1}^N\hat h_i s_i \sum_{j\neq i}J_{ij}s_j \Bigr).
    \label{eq:Z_integral}
\end{align}

% 3. Replicated partition function
Now, we introduce \(n\) replicas of the partition function, indexed by \(a \in \{1,\dots,n\}\):
\begin{align}
    Z_J^{n} = \prod_{a=1}^n Z_J^{(a)} = e^{-\beta n N} &\int \prod_{i, a} \frac{d h_i^a \, d\hat h_i^a }{2\pi i} 
    \exp\Bigl[\, - \sum_{i, a}\hat h_i^a \,h_i^a + \beta \sum_{i, a} \Sign\Bigl( J_D + h_i^a \Bigr) \Bigr] \nonumber\\
        \times &\sum_{\{s_i^a\}} \exp \Bigl( \, \sum_{i, a} \hat h_i^a s_i^a \sum_{j\neq i}J_{ij}s_j^a \Bigr)
    \label{eq:Z_replicated}
\end{align}
Here, the sum over the states is over all configurations of \(s_i^a\) for each replica \(a\) and site \(i\). 
Furthermore, indices \(i\) and \(j\) implicitly run over neurons 1, \dots, \(N\), while the replica index \(a\) 
runs from 1 to \(n\).

% 4. Disorder average over the couplings
Now, we perform the disorder average. Note that only the last term in the integrand of \eqref{eq:Z_integral} depends on the random couplings \(J_{ij}\).
The average factorizes since the \(J_{ij}'s\) are iid and they are decoupled in the exponential:
\begin{align}
\mathbb{E}_J \Bigl[\exp \bigl(\sum_{i, a}\sum_{j \ne i} J_{ij}\,\hat h_i^a s_i^a s_j^a\bigr)\Bigr]
    &= \prod_{i\neq j}\int_{-\infty}^{\infty} \!dJ_{ij}\;
    \sqrt{\frac{N}{2\pi}}\,e^{-\frac{1}{2}N J_{ij}^2}
    \, \exp \Bigl(J_{ij}\sum_{a}\hat h_i^a s_i^a s_j^a\Bigr)\nonumber\\
    &= \prod_{i\neq j}
    \exp\Bigl[\frac{1}{2N}\bigl(\sum_{a}\hat h_i^a s_i^a s_j^a\bigr)^2\Bigr] \nonumber\\
    &= \exp\Bigl(\frac{1}{2N}\sum_{i\neq j}\sum_{a,b}
      \hat h_i^a \hat h_i^b \,s_i^a s_i^b\,s_j^a s_j^b\Bigr).
    \label{eq:disorder_average}
\end{align}
Where we have used the identity:
\begin{equation}
    \int_{-\infty }^{\infty }\exp \left(-{1 \over 2}ax^{2}+Jx\right)\,dx =\left({2\pi  \over a}\right)^{1 \over 2}\exp \left({J^{2} \over 2a}\right)
    \label{eq:gaussian_integral}
\end{equation}

\subsection{Order Parameters}

% 5. Rearranging the disorder average
Since \(s_i^a s_i^b s_j^a s_j^b = 1\) when \(i = j\), the (missing) term having \(i = j\) in the sum over \(i\neq j\) in \eqref{eq:disorder_average} is of order 1.
Therefore, it is negligible in the limit of large \(N\) since the exponent is of order \(N\). With this observation, we can rearrange \eqref{eq:disorder_average},
introduce the missing term \(i = j\), and substitute into the partition function to obtain:
\begin{align}
    \mathbb{E}_J Z_J^{n} = e^{-\beta n N} \int &\prod_{i, a} \frac{d h_i^a \, d\hat h_i^a }{2\pi i} 
    \exp\Bigl[\, - \sum_{i, a}\hat h_i^a \,h_i^a + \beta \sum_{i, a} \Sign\Bigl( J_D + h_i^a \Bigr) \Bigr] \nonumber\\
    \times \sum_{\{s_i^a\}} &\exp\Bigl[\frac{1}{2N}\sum_{a,b} \Bigl( \sum_i \hat h_i^a \hat h_i^b \,s_i^a s_i^b \Bigr) \Bigl( \sum_j \,s_j^a s_j^b \Bigr) \Bigr].
    \label{eq:Z_disorder_average}
\end{align}

% 6. Defining the order parameters
Now, we introduce the order parameters:
\begin{align}
    Q^{ab} &= \frac{1}{N}\sum_{i} s^{a}_{i} s^{b}_{i}
    \quad &\text{for }a<b \\
    R^{ab} &= \frac{1}{N}\sum_{i} \hat h^{a}_{i}\,\hat h^{b}_{i}\,s^{a}_{i}\,s^{b}_{i}
    \quad &\text{for }a\le b
    \end{align}
    As we did for \(h_i^a\), we can introduce the conjugate variables \(\hat Q^{ab}\) and \(\hat R^{ab}\) using the Fourier representation of the delta function:
    \begin{align}
    1 \;&=\;
    \int \!\prod_{a<b} \frac{dQ^{ab}\,d\hat Q^{ab}}{2 \pi i}\;
    \exp \left[\sum_{a < b}\hat Q^{ab}\Bigl(\sum_{i}s^{a}_{i}s^{b}_{i}-NQ^{ab}\Bigr)\right] \\
    1 \;&=\;
    \int \!\prod_{a\le b} \frac{dR^{ab}\,d\hat R^{ab}}{2 \pi i}\;
    \exp \left[\sum_{a \le b}\hat R^{ab}\Bigl(\sum_{i}\hat h^{a}_{i}\hat h^{b}_{i}s^{a}_{i}s^{b}_{i}-N R^{ab}\Bigr)\right]
\end{align}

% 7. Introducing the order parameters
For convenience, further define \(Q^{aa} = 1\) and \(\hat Q^{aa} = 0\) for all \(a\). Thus, we can write:
\begin{align}
    &\; \exp \Bigl[\frac{1}{2N} \sum_{a,b} \Bigl( \sum_i \hat h_i^a \hat h_i^b \,s_i^a s_i^b \Bigr) \Bigl( \sum_j \,s_j^a s_j^b \Bigr) \Bigr] = \nonumber \\
    = &\int \prod_{a<b} \frac{dQ^{ab}\,d\hat Q^{ab}}{2 \pi i} \prod_{a\le b} \frac{dR^{ab}\,d\hat R^{ab}}{2 \pi i} \exp\Bigl[\tfrac12 N \sum_{a, b} \Bigl( Q^{ab}R^{ab} - Q^{ab} \hat Q^{ab} - R^{ab} \hat R^{ab} \Bigr) \Bigr] \nonumber \\
    &\quad \times \exp \Bigl[ \tfrac12 \sum_{a, b, i} \Bigl( \hat Q^{ab} s_i^a s_i^b + \hat R^{ab} \hat h_i^a \hat h_i^b s_i^a s_i^b \Bigr) + \tfrac12 \sum_a \Bigl( - N R^{aa} \hat R^{aa} + \hat R^{aa} \sum_i (\hat h_i^a)^2 \Bigr) \Bigr]
    \label{eq:order_parameters}
\end{align}
% TODO: here, i missed the diagonal term for \(Q^{ab} R^{ab}\). Check and fix (here, in 1RSB, and in replicated)! --> sarà il nostro piccolo segreto

% 8. Factorization of the integral
Looking at \eqref{eq:Z_disorder_average} and \eqref{eq:order_parameters}, the neurons \(i\) have become decoupled from each other. Therefore, the integral factorizes over the neurons,
and we can drop the \(i\) dependence. We can rewrite the \(n\)-th moment of the partition function as:
\begin{align}
    \mathbb{E}_{J}Z_{J}^{n} = e^{-\beta n N}
    \int \prod_{a<b} \frac{dQ^{ab}\,d\hat Q^{ab}}{2 \pi i} \prod_{a\le b} \frac{dR^{ab}\,d\hat R^{ab}}{2 \pi i}\,
    &e^{N F(Q^{ab}, \hat Q^{ab},R^{ab}, \hat R^{ab}) + N \Psi(\hat Q^{ab},\hat R^{ab})}
\end{align}
Where
\begin{align}
    &F(Q^{ab}, \hat Q^{ab},R^{ab}, \hat R^{ab}) = \tfrac12 \sum_{a,b}\bigl(Q^{ab}R^{ab}-\hat Q^{ab}Q^{ab}-\hat R^{ab}R^{ab}\bigr) - \tfrac12 \sum_{a} \hat R^{aa} R^{aa} \\
    &\Psi(\hat Q^{ab},\hat R^{ab}) = \log \int \prod_{a} \frac{dh^{a}\,d\hat h^a}{2 \pi i} \sum_{\{s^a\}} \exp \Bigl(-\sum_{a}\hat h^{a}h^{a} +\beta\sum_{a}\Sign(J_{D}+h^{a}) \Bigr) \nonumber \\
    &\quad\quad\quad\quad\quad\quad\quad\quad\quad \exp \Bigl( \tfrac12\sum_{a, b}\hat Q^{ab}s^{a}s^{b} + \tfrac12\sum_{a,b}\hat R^{ab}\hat h^{a}\hat h^{b}s^{a}s^{b} + \tfrac12 \sum_a \hat R^{aa} (\hat h^a)^2 \Bigr)
    \label{eq:Z_moment}
\end{align}
Note that after the factorization, the sum over neuron states is now over \(s^a \in \{\pm 1\}\) for each replica \(a\).
% \begin{align}
%     \exp \Bigl[&\frac{1}{2N} \sum_{a,b} \Bigl( \sum_i \hat h_i^a \hat h_i^b \,s_i^a s_i^b \Bigr) \Bigl( \sum_j \,s_j^a s_j^b \Bigr) \Bigr] =
%     \int \prod_{a<b} \frac{dQ^{ab}\,d\hat Q^{ab}}{2 \pi i} \prod_{a\le b} \frac{dR^{ab}\,d\hat R^{ab}}{2 \pi i} \nonumber \\
%     &\times \exp\Bigl[\tfrac12 N \sum_{a,b} Q^{ab} R^{ab} - N \sum_{a < b} Q^{ab} \hat Q^{ab} - N \sum_{a \le b} R^{ab} \hat R^{ab} 
%     + \sum_{i, a < b} \hat Q^{ab} s_i^a s_i^b + \sum_{i, a \le b} \hat R^{ab} \hat h_i^a \hat h_i^b s_i^a s_i^b \Bigr] = \nonumber \\
%     &\int d[Q,\hat Q,R,\hat R]\, \exp\Bigl[\tfrac12 N \sum_{a, b} \Bigl( Q^{ab}R^{ab} - Q^{ab} \hat Q^{ab} - R^{ab} \hat R^{ab} + 
%     \sum_i \hat Q^{ab} s_i^a s_i^b + \sum_i \hat R^{ab} \hat h_i^a \hat h_i^b s_i^a s_i^b \Bigr) \nonumber \\
%     &\quad + \tfrac12 \sum_a \Bigl( R^{aa} \hat R^{aa} + \hat R^{aa} \hat h_i^a \hat h_i^b s_i^a s_i^b \Bigr) \Bigr]
% \label{eq:order_parameters}
% \end{align}

\section{Replica-Symmetric Analysis}

\subsection{RS Ansatz}

We impose a replica-symmetric (RS) ansatz, which assumes that the order parameters are independent of the replica indices:
\begin{align}
    Q^{ab} &= q \;,\quad \hat Q^{ab} = \hat q \quad (a\neq b)\\
    R^{ab} &= r \;,\quad \hat R^{ab} = \hat r \quad (a\neq b)\\
    R^{aa} &= u \;,\quad \hat R^{aa} = \hat u          
\end{align}
Recall that \(Q^{aa} = 1\) and \(\hat Q^{aa} = 0\). The expressions for \(f\) and \(\Psi\) simplify to:
\begin{align}
    &F(q, \hat q, r, \hat r, u, \hat u) = \tfrac12 n \bigl[ (n-1)qr - (n-1)q \hat q - (n-1)r \hat r + u - 2 u \hat u \bigr] \\
    &\Psi(\hat q,\hat r, \hat u) = \log \int \prod_{a} \frac{dh^{a}\,d\hat h^{a}}{2 \pi i} \sum_{\{s^a\}} \exp \Bigl(-\sum_{a}\hat h^{a}h^{a} +\beta\sum_{a}\Sign(J_{D}+h^{a}) \Bigr) \nonumber \\
    &\quad\quad\quad\quad\quad\quad\quad\quad \exp \Bigl( \tfrac12\sum_{a \ne b}\hat q s^{a}s^{b} + \tfrac12\sum_{a \ne b}\hat r \hat h^{a}\hat h^{b}s^{a}s^{b} + \sum_a \hat u (\hat h^a)^2 \Bigr) \label{label:Psi}
\end{align}

We want to factorize the integral in \eqref{label:Psi} with respect to the replicas. To do so, we first use the identities:
\begin{align}
    \tfrac{1}{2} \sum_{a \ne b} \hat q s^{a}s^{b} &= \tfrac{1}{2} \hat q (\sum_{a} s^{a})^2 - \tfrac{1}{2} n \hat q \\
    \tfrac{1}{2} \sum_{a \ne b} \hat r \hat h^{a}\hat h^{b}s^{a}s^{b} &= \tfrac{1}{2} \hat r (\sum_{a} s^{a} \hat h^{a})^2 - \tfrac{1}{2} \hat r \sum_{a} (\hat h^{a})^2
    \label{eq:sum_identities}
\end{align}
Then, we use the Hubbard–Stratonovich transformation, followed by a change of variables, to obtain the identities:
\begin{align}
    \exp\!\Bigl(\tfrac12\,\hat q\bigl(\sum_a s^a\bigr)^2\Bigr)
    &=\int_{-\infty}^{\infty}\frac{dz}{\sqrt{2\pi}}\,
    \exp\!\Bigl(-\tfrac{z^2}{2}\Bigr)\,
    \exp\!\Bigl(\sqrt{\hat q}\,z\sum_a s^a\Bigr) \\
    \exp\!\Bigl(\tfrac12\,\hat r\bigl(\sum_a \hat h^a s^a\bigr)^2\Bigr)
    &=\int_{-\infty}^{\infty}\frac{dy}{\sqrt{2\pi}}\,
    \exp\!\Bigl(-\tfrac{y^2}{2}\Bigr)\,
    \exp\!\Bigl(\sqrt{\hat r}\,y\sum_a \hat h^a s^a\Bigr)
\end{align}

The neuron states are now decoupled, so the average factorizes across replicas:
\begin{align}
    \sum_{\{s^a\}} \exp\Bigl(\sqrt{\hat q}\,z\sum_a s^a + \sqrt{\hat r}\,y\sum_a \hat h^a s^a\Bigr) &= \\
    &= \prod_{a=1}^n \sum_{s^a=\pm1} \exp\Bigl(\sqrt{\hat q}\,z s^a + \sqrt{\hat r}\,y\hat h^a s^a\Bigr) = \\
    &= \prod_{a=1}^n \Bigl(2\cosh\bigl(\sqrt{\hat q}\,z + \sqrt{\hat r}\,y\hat h^a\bigr)\Bigr)
    \label{eq:states_average}
\end{align}

Looking at \eqref{label:Psi}, \eqref{eq:sum_identities}, and \eqref{eq:states_average}, we see that the integrals
are now factorized over the replicas. Thus, we can drop the replica index \(a\) and reassemble the pieces to obtain:
\begin{align}
    \Psi(\hat q,\hat r, \hat u) &= \log \int \frac{dz}{\sqrt{2\pi}} \, \frac{dy}{\sqrt{2\pi}} e^{- \frac{z^2 + y^2}{2}} e^{-n \frac{\hat q}{2}} \\
    \Bigl[\; \sum_{s = \pm 1} \int dh \, d\hat h \; &\exp\Bigl(-h \hat h + \beta \Sign(h + J_D) + (\hat u - \tfrac12 \hat r) \hat h^2 + s(\sqrt{\hat q}\,z + y\,\sqrt{\hat r}\,\hat h)\Bigr) \Bigr]^n \nonumber
\end{align}

We can now integrate out the variables \(h\) and \(\hat h\). The integral over \(\hat h\) is Gaussian, and we can use the identity \eqref{eq:gaussian_integral},
with \(a = \hat r - 2 \hat u\) and \(J = - h + s y \sqrt{\hat r}\):
\begin{align}
    &\quad \int dh \, d\hat h \; \exp\Bigl(-h \hat h + \beta \Sign(h + J_D) - \tfrac12 (\hat r - 2 \hat u) \hat h^2 + s(\sqrt{\hat q}\,z + y\,\sqrt{\hat r}\,\hat h)\Bigr) = \nonumber \\
    &= \sqrt\frac{2 \pi}{\hat r - 2 \hat u} e^{s \sqrt {\hat q}\,z} \int dh \, \exp\Bigl[ \beta \Sign(h + J_D) + \frac{(h - sy\sqrt{\hat r})^2}{2 (\hat r - 2 \hat u)} \Bigr]
\end{align}

For the integral over \(h\), we can split the domain of integration into two parts, depending on the sign of \(h + J_D\),
and express the result in terms of the error function.
\begin{align}
    I \;&=\; \int_{-\infty}^{\infty} dh\;
       \exp\!\Bigl[
         \beta\,\operatorname{sgn}(h+J_D)
         - \tfrac12 \frac{(h-sy\sqrt{\hat r})^{2}}{(2\hat u - \hat r )}
       \Bigr] = \nonumber\\
    &= e^{-\beta}\!\int_{-\infty}^{-J_D} dh\;
    \exp\!\Bigl[- \tfrac12 \tfrac{(h-sy\sqrt{\hat r})^{2}}{2 \hat u - \hat r}\Bigr]
    \;+\;
    e^{\beta}\!\int_{-J_D}^{\infty} dh\;
    \exp\!\Bigl[- \tfrac12 \tfrac{(h-sy\sqrt{\hat r})^{2}}{2 \hat u - \hat r}\Bigr].
\end{align}
Define \(\Delta = 2 \hat u - \hat r\) and make the change of variables \(x = \frac{h-sy\sqrt{\hat r}}{\sqrt{\Delta}}\),
\(dh = \sqrt \Delta dx\). The threshold becomes \(x_* = \frac{-J_D-sy\sqrt{\hat r}}{\sqrt{\Delta}}\), and:
\begin{align}
    I &= e^{-\beta} \int_{-\infty}^{x_*} e^{- \tfrac12 x^2} \sqrt{\Delta} \, dx + e^{\beta} \int_{x_*}^{\infty} e^{- \tfrac12 x^2} \sqrt{\Delta} \, dx = \nonumber\\
    &= \sqrt{2\pi\,\Delta}\,
    \Bigl[
        e^{-\beta}\,H(-x_*)
        +e^{\beta}\,H(x_*)
    \Bigr],
\end{align}
where \(H(x) = \frac12\,\operatorname{erfc}\!\Bigl(\frac{x}{\sqrt2}\Bigr) = \int_{x}^\infty \frac{dh}{\sqrt{2 \pi}} e^{-h^2 / 2}\). Thus,
\begin{align}
    \Psi(\hat q,\hat r, \hat u) &= \log \int \frac{dz}{\sqrt{2\pi}} \, \frac{dy}{\sqrt{2\pi}} e^{- \frac{z^2 + y^2}{2}} e^{-n \frac{\hat q}{2}} \\
    &\quad\quad\quad\; \Bigl[ \, \sum_{s = \pm 1} e^{s \sqrt {\hat q}\,z}\,
    \Bigl[
        e^{-\beta}\,H(-x_*)
        +e^{\beta}\,H(x_*)
    \Bigr] \Bigr]^n
\end{align}

\subsection{Saddle Point Equations}

We want to exploit the replica trick:
\begin{equation}
    \mathbb{E}_J \log Z_J = \lim_{n\to 0} \frac{\log \mathbb{E}_J Z_J^n}{n}
\end{equation}
In the thermodynamic limit \(N \to \infty\), the logarithm of the \(n\)-th moment of the partition function can 
be evaluated using the saddle point method. 

We can expand \(F(q, \hat q, r, \hat r, u, \hat u)\) and \(\Psi(\hat q, \hat r, \hat u)\) in the \(n\to 0\) limit, obtaining:
\begin{align}
    F(q, \hat q, r, \hat r, u, \hat u) &= \tfrac12 n \bigl[ -qr + q \hat q + r \hat r + u - 2 u \hat u \bigr] + O(n^2) \\
    \Psi(\hat q, \hat r, \hat u) &= n \int Dz \, Dy \, \log \Bigl[\, \sum_{s = \pm 1} e^{s \sqrt {\hat q}\,z}\,
    \Bigl[
        e^{-\beta}\,H(-x_*)
        +e^{\beta}\,H(x_*)
    \Bigr]\Bigr] \nonumber \\ 
    &\quad - n \frac{\hat q}{2} + O(n^2)
\end{align}
where we are using the short-hand notation \(Dz = \frac{dz}{\sqrt{2\pi}} e^{-\frac{1}{2}z^2}\) and \(Dy = \frac{dy}{\sqrt{2\pi}} e^{-\frac{1}{2}y^2}\).

Thus, we obtain:
\begin{align}
    \frac{1}{N} \, \mathbb{E}_J \log Z_J = - \beta + f(q^*, \hat q^*, r^*, \hat r^*, u^*) + \psi(\hat q^*, \hat r^*, \hat u^*)
    \label{eq:log_Z}
\end{align}
where we have defined:
\begin{align}
    f(q, \hat q, r, \hat r, u, \hat u) &= \lim_{n \to 0} \frac{F(q, \hat q, r, \hat r, u, \hat u)}{n} = \tfrac12 \bigl[ -qr + q \hat q + r \hat r + u - 2 u \hat u \bigr] \\
    \psi(\hat q, \hat r, \hat u) &= \lim_{n \to 0} \frac{\Psi(\hat q, \hat r, \hat u)}{n} = - \frac{\hat q}{2} + \int Dz \, Dy \, \log \Bigl[ \\ 
    &\quad \sum_{s = \pm 1} e^{s \sqrt {\hat q}\,z}\,
    \Bigl[
        e^{-\beta}\,H\!\bigl(\frac{J_D+sy\sqrt{\hat r}}{\sqrt{2 \hat u - \hat r}}\bigr)
        +e^{\beta}\,H\!\bigl(- \frac{J_D+sy\sqrt{\hat r}}{\sqrt{2 \hat u - \hat r}}\bigr)
    \Bigr]\Bigr] \nonumber
\end{align}
and \(q^*, \hat q^*, r^*, \hat r^* u^*, \hat u^*\) are the global maximizers of \(f + \psi\).

To find the saddle point, we need to solve the saddle point equations. \(\psi(\hat q, \hat r, \hat u)\) is independent
of \(q, r, u\), so we immediately obtain:
\begin{align}
    \frac{\partial f}{\partial q}
    &=\tfrac12(-\,r + \hat q)=0 
    \;\;\Longrightarrow\;\;\hat q = r, \\
    \frac{\partial f}{\partial r}
    &=\tfrac12(-\,q + \hat r)=0 
    \;\;\Longrightarrow\;\;\hat r = q, \\
    \frac{\partial f}{\partial u}
    &=\tfrac12(1 - 2\,\hat u)=0 
    \;\;\Longrightarrow\;\;\hat u = \tfrac12.
\end{align}

% Deriving with respect to the other variables gives:
% \begin{equation}
%     \frac{1}{2}q + \frac{\partial \psi}{\partial \hat q} = 0, \quad
%     \frac{1}{2}r + \frac{\partial \psi}{\partial \hat r} = 0, \quad
%     - u + \frac{\partial \psi}{\partial \hat u} = 0.
% \end{equation}

We can simplify the objective function \(f + \psi\) using the relations we have found:
\begin{align}
    &\frac{1}{N} \, \mathbb{E}_J \log Z_J = \max_{q, r} \left[ - \beta + \frac{1}{2}(q-1)r + \phi(r, q) \right] \\
    &\phi(r, q) = \int Dz \, Dy \, \log \Bigl[ \, \sum_{s = \pm 1} e^{s \sqrt {r}\,z}\,
    \Bigl[
        e^{-\beta}\,H\!\bigl(\frac{J_D+sy\sqrt{q}}{\sqrt{1 - q}}\bigr)
        +e^{\beta}\,H\!\bigl(- \frac{J_D+sy\sqrt{q}}{\sqrt{1 - q}}\bigr)
    \Bigr]\Bigr] \nonumber
\end{align}

Now, we are left with only two order parameters \(q\) and \(r\) to optimize. To lighten the notation, 
define:
\begin{align}
    H_{\tau\tau'} \;&=\; H\!\biggl(\frac{\tau\,J_{D} \;+\; \tau'\,y\sqrt{q}}{\sqrt{1 - q}}\biggr) \\
    E_{\tau\tau'} \;&=\; \frac{1}{\sqrt{2\pi}}\bigl[\tau J_{D}\sqrt{q} \;+\; \tau' y \bigr]\;
    \exp\!\Biggl(-\frac{\bigl(\tau J_{D} \;+\; \tau'\,y\sqrt{q}\bigr)^{2}}{2\,(1 - q)}\Biggr)
\end{align}
Recall that \(H(x) = \frac12\,\operatorname{erfc}\!\Bigl(\frac{x}{\sqrt2}\Bigr)\).
Taking derivatives of the objective with respect to \(q\) and \(r\), we obtain the saddle point equations:
\begin{align}
    q \;&=\; \int Dz\,Dy
    \left[
    \frac{
    e^{\,z\sqrt{r}}\Bigl(H_{++}\,e^{-\beta} \;+\; H_{--}\,e^{\beta}\Bigr)
    \;-\;
    e^{-\,z\sqrt{r}}\Bigl(H_{+-}\,e^{-\beta} \;+\; H_{-+}\,e^{\beta}\Bigr)
    }{
    e^{\,z\sqrt{r}}\Bigl(H_{++}\,e^{-\beta} \;+\; H_{--}\,e^{\beta}\Bigr)
    \;+\;
    e^{-\,z\sqrt{r}}\Bigl(H_{+-}\,e^{-\beta} \;+\; H_{-+}\,e^{\beta}\Bigr)
    }
    \right]^2 \nonumber \\
    r \;&=\; \frac{1}{\sqrt{q(1 - q)^3}}
    \int Dz\,Dy\,
    \frac{
    e^{\,z\sqrt{r}}\Bigl(E_{++}\,e^{-\beta} \;+\; E_{--}\,e^{\beta}\Bigr)
    \;+\;
    e^{-\,z\sqrt{r}}\Bigl(E_{+-}\,e^{-\beta} \;+\; E_{-+}\,e^{\beta}\Bigr)
    }{
    e^{\,z\sqrt{r}}\Bigl(H_{++}\,e^{-\beta} \;+\; H_{--}\,e^{\beta}\Bigr)
    \;+\;
    e^{-\,z\sqrt{r}}\Bigl(H_{+-}\,e^{-\beta} \;+\; H_{-+}\,e^{\beta}\Bigr)
    } \nonumber
\end{align}

These equations can be solved numerically to find the order parameters \(q\) and \(r\). To this end,
we employ a fixed-point iteration method with a small damping coefficient to mitigate oscillations, 
varying the control parameters \(\beta\) and \(J_D\) in a wide range of values.
We focus the numerical analysis on the region \(J_D \ge 0\), since this is the most interesting regime
for the design of algorithms mapping memories to fixed points. Fixed-point iteration reveals that \(q^* = r^* = 0\)
satisfies both equations and thus is a saddle point for all values of \(\beta\) and \(J_D \ge 0\). 
This can also be checked analytically by direct substitution, considering the limit \(q \to 0^+\) to handle the 
singularity in the denominator of the second equation. By doing this, it can be seen that \(q^* = r^* = 0\)
is also a saddle point in the region \(J_D < 0\). 
We have not been able to find any other saddle points in the interesting regime \(J_D \ge 0\).

\begin{figure}[htbp]
  \centering
  \begin{subfigure}[b]{0.45\textwidth}
    \centering
    \includegraphics[width=\linewidth]{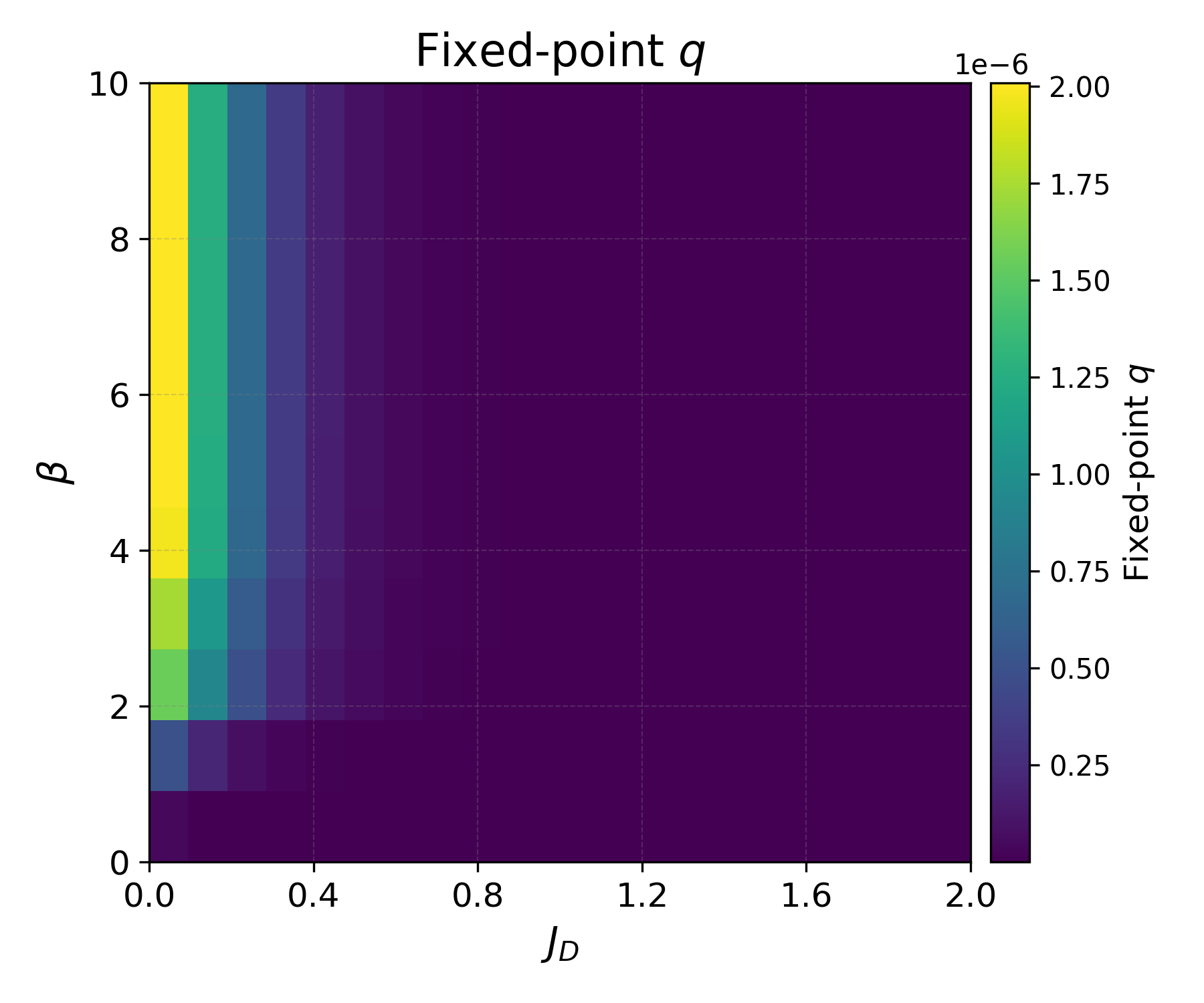}
    % \caption{Caption for figure 1}
    \label{fig:1}
  \end{subfigure}
  \hfill
  \begin{subfigure}[b]{0.45\textwidth}
    \centering
    \includegraphics[width=\linewidth]{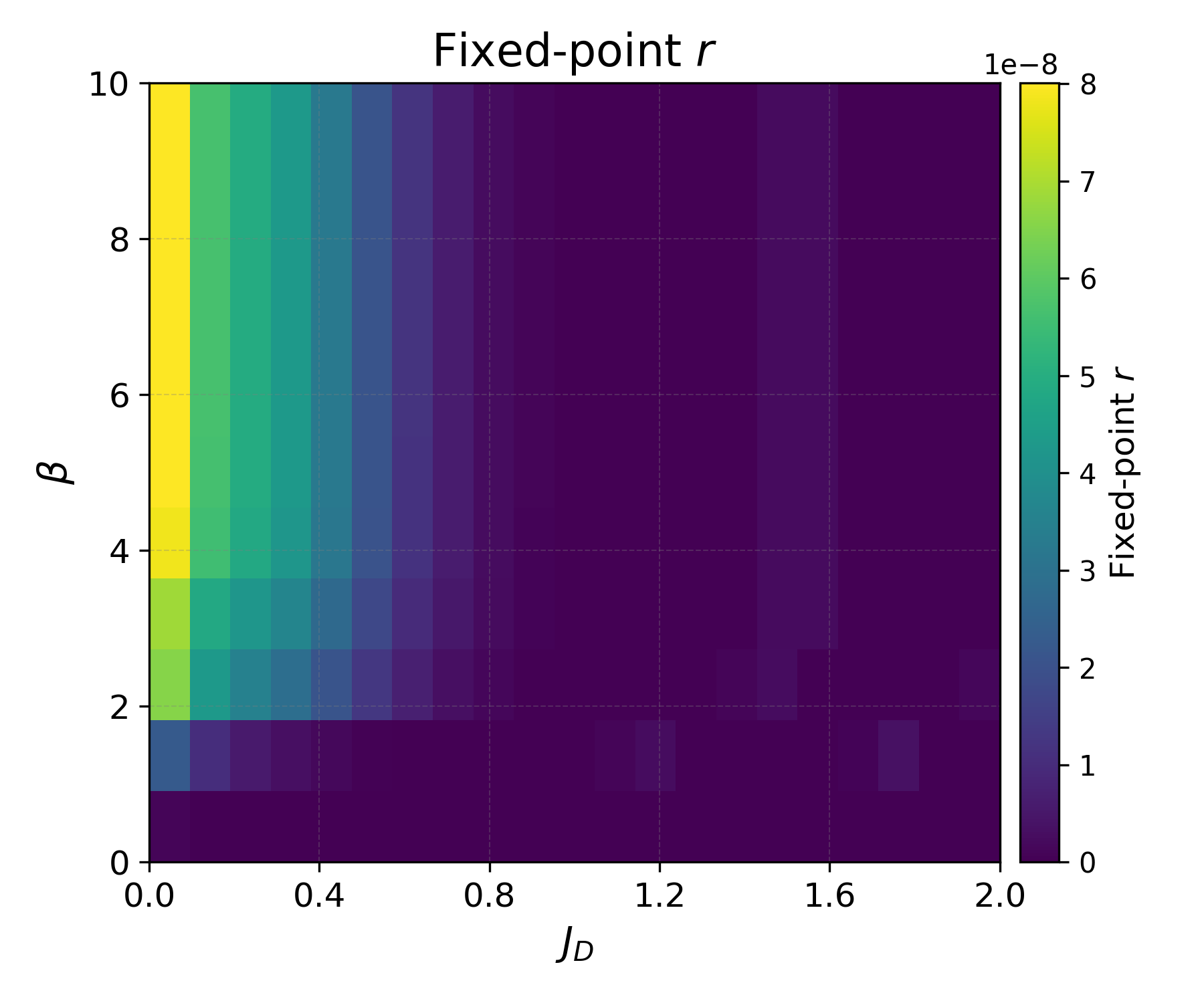}
    % \caption{Caption for figure 2}
    \label{fig:2}
  \end{subfigure}
  \caption{Solution of the saddle point equations for \(q\) and \(r\) in the RS ansatz. Fixed point iteration with damping coefficient \(\alpha = 0.9\). Iteration is terminated when the relative change in
  the order parameters is less than \(10^{-6}\). \textbf{Left:} \(q^*\) as a function of \(\beta\) and \(J_D\). \textbf{Right:} \(r^*\) as a function of \(\beta\) and \(J_D\).}
  \label{fig:both}
\end{figure}

\subsection{Computing the Number of Fixed Points}

Finally, we can compute the free energy associated with the saddle point \(q^* = r^* = 0\):
\begin{align}
    f_{RS}(q^* = r^* = 0, \beta, J_D) &= - \frac{1}{\beta N} \mathbb{E}_J \log Z_J = \\
    &= 1 - \frac{1}{2\beta}(q^*-1)r^* - \frac{1}{\beta} \phi(r^*, q^*) = \\
    &= 1 - \frac{1}{\beta}\, \log \Bigl[ 2 e^{-\beta} H(J_D) + 2 e^{\beta} H(-J_D) \Bigr] = \\
    &= - \frac{\log 2}{\beta}  - \frac{1}{\beta}\, \log \Bigl[ 1 + (e^{-2\beta} - 1) H(J_D) \Bigr]
\end{align}
Incidentally, it can be checked that this expression is equal to the annealed free energy \(f_{ann}(\beta, J_D) = - \frac{1}{\beta N} \log \mathbb{E}_J Z_J \) of the system.
And in fact, by considering explicitly the first and second moments of the partition function over the disorder, which we have computed, one can show that the ratio between the variance and the second moment
of the partition function vanishes in the limit of a large system. This implies that, in the thermodynamic limit, its fluctuations are negligible and Jensen's inequality is tight for this system:
\begin{equation}
    \lim_{N \to \infty} \frac{1}{N} \mathbb{E}_J \log Z_J \;=\; \lim_{N \to \infty} \frac{1}{N} \log \mathbb{E}_J Z_J
\end{equation}

Having obtained the free energy, using standard thermodynamic relations, we can compute the average energy and entropy densities as a function of the control parameters:
\begin{align}
    u(\beta,J_D) \;&=\; \frac{1}{N}\langle E\rangle \;=\; \frac{\partial\bigl[\beta\,f_{RS}(q^*=r^*=0,\beta,J_D)\bigr]}{\partial \beta} = \\
    &= \frac{\partial}{\partial \beta}\Biggl[-\ln 2 \;-\; \ln\Bigl(1 + (e^{-2\beta}-1)\,H(J_D)\Bigr)\Biggr] \\[1ex]
    &= \frac{2\,e^{-2\beta}\,H(J_D)}{\,1 + (e^{-2\beta}-1)\,H(J_D)\,}\!,
\end{align}
\begin{align}
    s(\beta,J_D) \;&=\; \frac{S(\beta, J_D)}{N} \;=\; \beta\bigl(u(\beta, J_D) - f_{RS}(q^*=r^*=0,\beta,J_D)\bigr) \\
    &= \beta\Biggl[\frac{2\,e^{-2\beta}\,H(J_D)}{\,1 + (e^{-2\beta}-1)\,H(J_D)\,} 
    + \frac{\ln 2}{\beta} + \frac{1}{\beta}\,\ln\Bigl(1 + (e^{-2\beta}-1)\,H(J_D)\Bigr)\Biggr] \\[1ex]
    &= \ln 2 \;+\; \ln\Bigl(1 + (e^{-2\beta}-1)\,H(J_D)\Bigr) 
    \;+\; \frac{2\,\beta\,e^{-2\beta}\,H(J_D)}{\,1 + (e^{-2\beta}-1)\,H(J_D)\,}.
\end{align}

In the limit \(\beta \to \infty\), we have \(u(\beta, J_D) \to 0\) and \(s(\beta, J_D) \to \ln 2 + \ln\bigl(1 - H(J_D)\bigr)\). The
solution is unphysical in the region \(J_D < 0\), where the entropy density becomes negative. This suggests that there are no fixed
points in the typical case in this region. To obtain the correct description of the system in this region, we would need to seek a 
different saddle point.

\begin{figure}[htbp]
  \centering
  \includegraphics[width=0.8\linewidth]{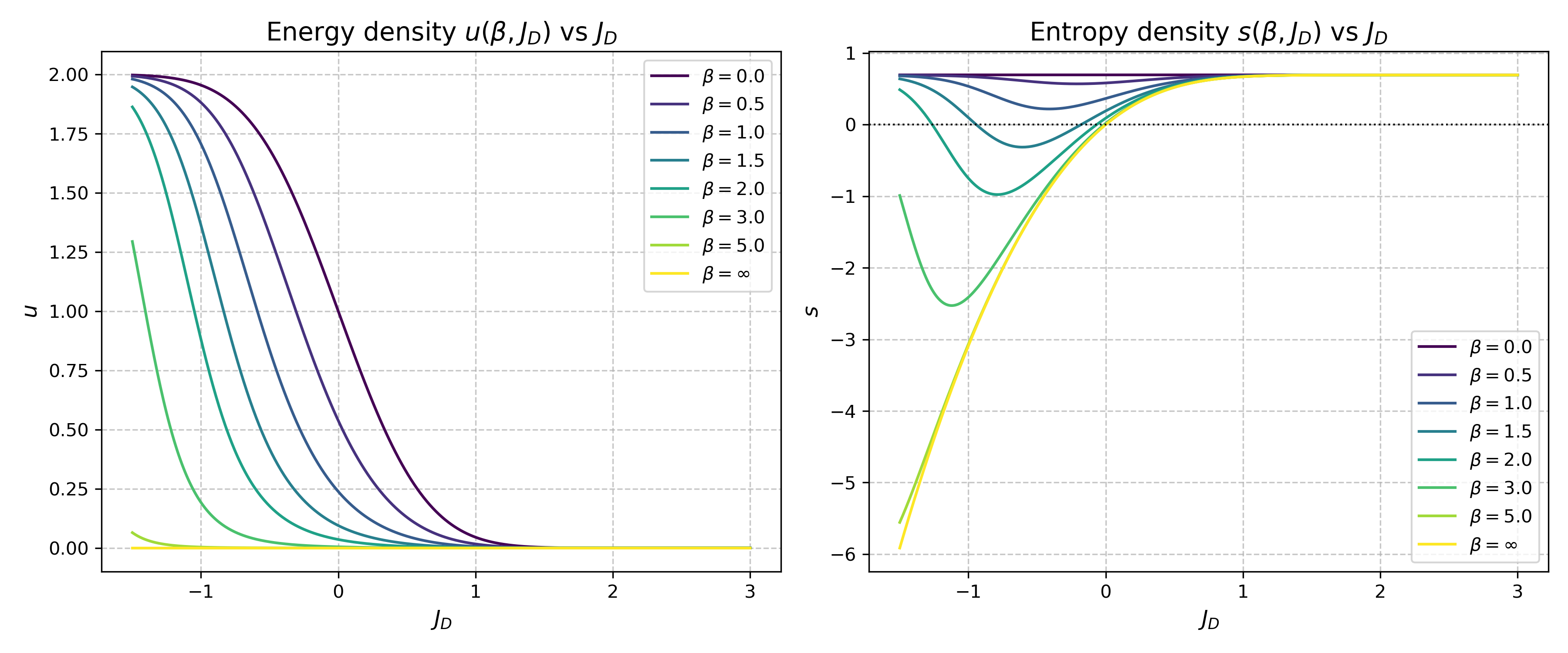}
  \caption{Entropy density and Energy density as a function of \(J_D\) for several values of \(\beta\).}
  \label{fig:entropy_vs_JD}
\end{figure}

\begin{figure}[htbp]
  \centering
  \includegraphics[width=0.8\linewidth]{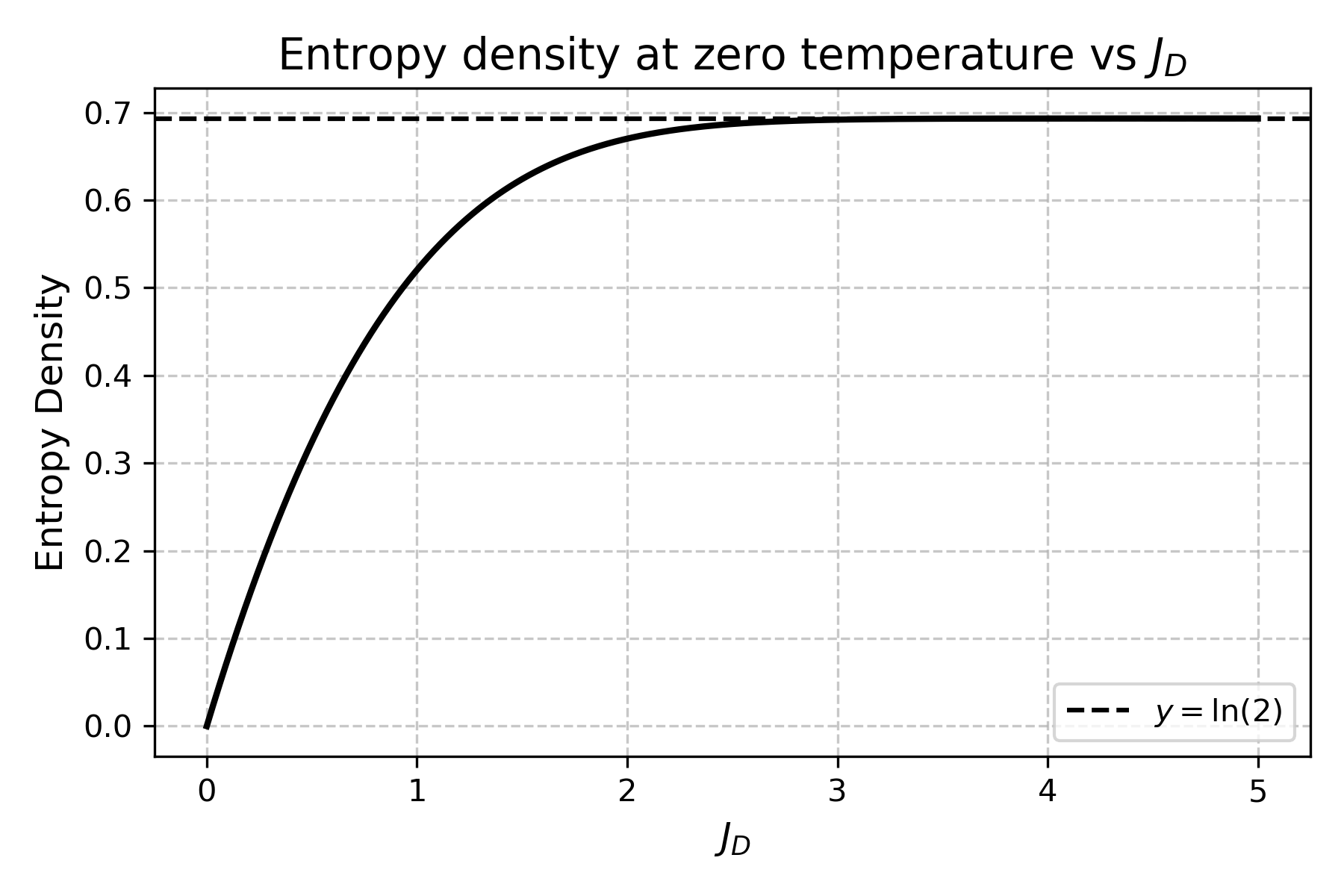}
  \caption{Logarithm of the number of fixed points, normalized by \(N\), as a function of \(J_D\).}
  \label{fig:entropy_vs_JD_zero_temp}
\end{figure}

When \(J_D\) is large enough, all configurations 
are fixed points and the entropy density becomes approximately \(\ln 2\).
In Figure \ref{fig:entropy_vs_JD}, we plot the entropy density and energy density as a function of \(J_D\) for several
values of \(\beta\), as well as in the limit \(\beta \to \infty\). In Figure \ref{fig:entropy_vs_JD_zero_temp}, we also show a zoom of 
the region \(J_D > 0\) in the zero temperature limit. That curve represents the logarithm of the number of fixed points
of the system, normalized by the number of neurons \(N\), in the large \(N\) limit.

% Its behavior as a function of \(J_D\) is plotted in Figure \ref{fig:entropy_vs_JD}.

% Note the following. One could wonder whether the entropy is negative for \(J_D < 0\) because we are considering the wrong saddle point.
% However, as we remarked, the free energy of the saddle point \(q^* = r^* = 0\) is equal to the annealed free energy, and the annealed
% free energy equals the quenched free energy in the limit \(N \to \infty\) for this system. Thus, the saddle point \(q^* = r^* = 0\) provides
% the correct description of the typical behavior of the system in the thermodynamic limit.

One can do a stability analysis by expanding \(F + \Psi\) to second order around the saddle point \(q^* = r^* = 0\), and
computing the eigenvalues of the Hessian matrix. If the saddle point is stable, it means that breaking replica symmetry
does not lower the free energy, at least in the vicinity of the saddle point (but there may still be other saddle points
with lower free energy that were ruled out by the RS ansatz). On the other hand, if the solution is unstable, we can be
sure that replica symmetry breaking is necessary to lower the free energy. It can be checked that the saddle point \(q^* = r^* = 0\)
is stable for \(J_D \ge -0.178\). Below this value of \(J_D\), the RS solution is surely not the correct one. However, the conclusion
that there are no fixed points remains valid. In fact, the quenched entropy is bounded from above by the annealed entropy, which coincides
with that computed using the saddle point \(q^* = r^* = 0\). Since the annealed entropy is negative in the region where instabilities
occur, the quenched entropy must also be negative there.

\section{Replica Symmetry Breaking}

\subsection{1RSB Ansatz}

In the one‑step replica‑symmetry‑breaking (1‑RSB) scheme the $n$ replicas are partitioned into $n/m$ blocks ("states") of equal size $m$. Let $\mathcal{B}(a)$ denote the block containing replica $a$. The ansatz introduces two off‑diagonal overlap levels: an \emph{intra‑state} value (subscript 1) and an \emph{inter‑state} value (subscript 0).

\begin{align}
    Q^{ab} &=
    \begin{cases}
    1         & a=b;\\
    q_{1}     & a\neq b, \quad \mathcal{B}(a)=\mathcal{B}(b);\\
    q_{0}     & a\neq b, \quad \mathcal{B}(a)\neq \mathcal{B}(b);
    \end{cases}\\
    \widehat{Q}^{ab} &=
    \begin{cases}
    0             & a=b;\\
    \widehat{q}_{1} & a\neq b, \quad \mathcal{B}(a)=\mathcal{B}(b);\\
    \widehat{q}_{0} & a\neq b, \quad \mathcal{B}(a)\neq \mathcal{B}(b);
    \end{cases}\\
    R^{ab} &=
    \begin{cases}
    u       & a=b;\\
    r_{1}   & a\neq b, \quad \mathcal{B}(a)=\mathcal{B}(b);\\
    r_{0}   & a\neq b, \quad \mathcal{B}(a)\neq \mathcal{B}(b);
    \end{cases}\\
    \widehat{R}^{ab} &=
    \begin{cases}
    \widehat{u}   & a=b;\\
    \widehat{r}_{1} & a\neq b, \quad \mathcal{B}(a)=\mathcal{B}(b);\\
    \widehat{r}_{0} & a\neq b, \quad \mathcal{B}(a)\neq \mathcal{B}(b).
    \end{cases}
    \label{eq:1rsb_ansatz}
\end{align}

We can simplify the terms appearing in \eqref{eq:Z_moment} using the ansatz:
\begin{align}
    \sum_{a,b} Q^{ab} R^{ab} &= n \bigl[ (m-1)q_1r_1 + (n - m)q_0r_0 + u \bigr] \\
    \sum_{a,b} Q^{ab} \hat Q^{ab} &= n \bigl[ (m-1)q_1\hat q_1 + (n - m)q_0\hat q_0 \bigr] \\
    \sum_{a,b} R^{ab} \hat R^{ab} &= n \bigl[ (m-1)r_1\hat r_1 + (n - m)r_0\hat r_0 + \hat u u \bigr] \\
    \sum_{a} R^{aa} \hat R^{aa} &= n \hat u u
\end{align}

Thus, we can rewrite \(F\) as:
\begin{align}
    F(q_0, q_1, \hat q_0, \hat q_1, r_0, r_1, \hat r_0, \hat r_1, u, \hat u) &= \tfrac12 n \, \Bigl[ (m-1) (q_1 r_1 - q_1 \hat q_1 - r_1 \hat r_1) + u \nonumber \\
    &\quad\quad\quad + (n-m) (q_0 r_0 - q_0 \hat q_0 - r_0 \hat r_0) - 2 \hat u u \Bigr]
\end{align}

Now we turn to \(\Psi\). Start by rewriting:
\begin{align}
    \sum_{a, b} \hat Q^{ab} s_i^a s_i^b &= \hat q_0 \left[ \left( \, \sum_a s^a \right)^2 - \sum_{B=1}^{n/m} \left( \, \sum_{a \in B} s^a \right)^2 \right] + \hat q_1 \left[ \, \sum_{B=1}^{n/m} \left( \, \sum_{a \in B} s^a \right)^2 - n \right] + 0 = \nonumber \\
    &= \hat q_0 \left( \, \sum_a s^a \right)^2 + (\hat q_1 - \hat q_0) \sum_{B=1}^{n/m} \left( \, \sum_{a \in B} s^a \right)^2 - n \, \hat q_1 \nonumber \\
    \sum_{a, b} \hat R^{ab} \hat h^a \hat h^b s^a s^b &= \hat r_0 \left( \, \sum_a \hat h^a s^a \right)^2 + (\hat r_1 - \hat r_0) \sum_{B=1}^{n/m} \left( \, \sum_{a \in B} \hat h^a s^a \right)^2 + (\hat u - \hat r_1) \sum_{a} (\hat h^a)^2 \nonumber \\
    \sum_a \hat R^{aa} (\hat h^a)^2 &= \hat u \sum_{a} (\hat h^a)^2
\end{align}

We can linearize the squares using the Hubbard–Stratonovich transformation plus a change of variables.
This introduces a pair of standard Gaussian variables for each block \(B\), plus another pair for the global sum:
\begin{align}
    \exp\left(\tfrac12 \sum_{a, b} \hat Q^{ab} s_i^a s_i^b\right) &= e^{- \tfrac12 \hat q_1 n} \int Dz_0 \, \exp\left(z_0 \sqrt{\hat q_0} \sum_a s^a\right) \nonumber \\
    &\quad\quad\times \prod_{B=1}^{n/m} \int Dz_B \, \exp\left(z_B \sqrt{\hat q_1 - \hat q_0} \sum_{a \in B} s^a\right) \\
    \exp\left(\tfrac12 \sum_{a, b} \hat R^{ab} \hat h^a \hat h^b s^a s^b\right) &= e^{\tfrac12 (\hat u - \hat r_1) \sum_a (\hat h^a)^2} \int Dy_0 \, \exp\left(y_0 \sqrt{\hat r_0} \sum_a \hat h^a s^a\right) \nonumber \\
    &\quad\quad\times \prod_{B=1}^{n/m} \int Dy_B \, \exp\left(y_B \sqrt{\hat r_1 - \hat r_0} \sum_{a \in B} \hat h^a s^a\right)
\end{align}

% Putting the pieces together, we can rewrite \(\Psi\) as:
% \begin{align}
%     &\quad \Psi(\hat q_0, \hat q_1, \hat r_0, \hat r_1, \hat u) = - \tfrac12 n \, \hat q_1 \; + \nonumber \\
%     &+ \, \log \Bigl[ \int Dz_0 \, Dy_0 \prod_B Dz_B \, Dy_B \prod_a dh^a \, d \hat h^a \, e^{ - \hat h^a h^a + \beta \Sign(h^a + J_D) + \tfrac12 (2 \hat u - \hat r_1) (\hat h^a)^2 } \nonumber \\
%     &\sum_{\{s^a\}} \prod_B \exp\left( \left(z_0\sqrt{\hat q_0} + z_B \sqrt{\hat q_1 - \hat q_0} \right) \sum_{a \in B} s^a + \left(y_0\sqrt{\hat r_0} + y_B \sqrt{\hat r_1 - \hat r_0} \right) \sum_{a \in B} s^a \hat h^a \right)\Bigr] \nonumber
% \end{align}
Note that after these manipulations, the integrals in \(\Psi\) are factorized over the replicas. We can write \(\Psi\) as:
\begin{equation}
    \Psi(\hat q_0, \hat q_1, \hat r_0, \hat r_1, \hat u) = - \tfrac12 n \, \hat q_1 + \log \Bigl[ \int Dz_0 \, Dy_0 \prod_B Dz_B \, Dy_B \prod_{a \in B} I_{a, B} \Bigr]
\end{equation}
where, given a replica index \(a\) belonging to block \(B\), \(I_{a, B}\) is given by:
\begin{align}
    I_{a, B} = \sum_{s^a = \pm 1} \int \frac{dh^a \, d \hat h^a}{2 \pi i} \exp\left[ - \hat h^a h^a + \beta \Sign(h^a + J_D) + \tfrac12 (2 \hat u - \hat r_1) (\hat h^a)^2 \right] \nonumber \\
    \exp\left[\left(z_0 \sqrt{\hat q_0} + z_B \sqrt{\hat q_1 - \hat q_0}\right) s^a + \left(y_0 \sqrt{\hat r_0} + y_B \sqrt{\hat r_1 - \hat r_0}\right) s^a \hat h^a \right]
\end{align}

The integral over \(\hat h^a\) is Gaussian, and we can use the identity \eqref{eq:gaussian_integral} with \(a = \hat r_1 - 2 \hat u\) and \(J = - h^a + s^a y_0 \sqrt{\hat r_0} + s^a y_B \sqrt{\hat r_1 - \hat r_0}\):
\begin{align}
    \int dh^a \, d \hat h^a e^{ - \hat h^a h^a - \tfrac12 (- 2 \hat u + \hat r_1) (\hat h^a)^2 + \left(y_0 \sqrt{\hat r_0} + y_B \sqrt{\hat r_1 - \hat r_0}\right) s^a \hat h^a } = \nonumber \\
    = \sqrt{\frac{2 \pi}{\hat r_1 - 2 \hat u}} \exp\left(\frac{h^a - s^a y_0 \sqrt{\hat r_0} - s^a y_B \sqrt{\hat r_1 - \hat r_0}}{2(\hat r_1 - 2 \hat u)}\right)^2
\end{align}
Then, like in the RS case, we can split the domain of integration, change variables, and express the integral in terms of the error function. Let \(\Delta = 2 \hat u - \hat r_1\), and \(\mu = s^a y_0 \sqrt{\hat r_0} + s^a y_B \sqrt{\hat r_1 - \hat r_0}\). Then:
\begin{align}
    I_{a, B} &= \sum_{s^a = \pm 1} \sqrt{\frac{2 \pi}{- \Delta}} e^{\left(z_0 \sqrt{\hat q_0} + z_B \sqrt{\hat q_1 - \hat q_0}\right) s^a} \, \int \frac{d h^a}{2 \pi i} \,
    \exp\left(\beta \Sign(h^a + J_D) - \frac{(h^a - \mu)^2}{2 \Delta}\right) = \nonumber \\
    &= \sum_{s^a = \pm 1} e^{\left(z_0 \sqrt{\hat q_0} + z_B \sqrt{\hat q_1 - \hat q_0}\right) s^a} \left[ e^{-\beta} H(-x_*) + e^\beta H(x_*) \right]
\end{align}
where \(x_* = \frac{-J_D - \mu}{\sqrt{\Delta}}\).

At this point, we can drop the replica index \(a\) and the block index \(B\), and rewrite \(\Psi\) as:
% \begin{align}
%     \Psi(\hat q_0, \hat q_1, \hat r_0, \hat r_1, \hat u) = - \tfrac12 n \, \hat q_1 + \log \Bigl[ \int Dz_0 \, Dy_0 \int \prod_B Dz_B \, Dy_B \\
%     \times \left( \sum_{s = \pm 1} e^{\left(z_0 \sqrt{\hat q_0} + z_B \sqrt{\hat q_1 - \hat q_0}\right) s} \left[ e^{-\beta} H(-x_*) + e^\beta H(x_*) \right] \right)^m \Bigr]
% \end{align}
\begin{align}
    \Psi(\hat q_0, \hat q_1, \hat r_0, \hat r_1, \hat u) &= - \tfrac12 n \, \hat q_1 + \log \int Dz_0 \, Dy_0 \Bigl[ \int Dz \, Dy \nonumber \\
    &\quad \times \Bigl( \sum_{s = \pm 1} e^{\left(z_0 \sqrt{\hat q_0} + z \sqrt{\hat q_1 - \hat q_0}\right) s} \left[ e^{-\beta} H(-x_*) + e^\beta H(x_*) \right] \Bigr)^m \, \Bigr]^{\frac{n}{m}}
    \label{eq:Psi_simplified_1rsb}
\end{align}
where we have redefined:
\begin{equation}
    x_* = \frac{-J_D - s y_0 \sqrt{\hat r_0} - s y \sqrt{\hat r_1 - \hat r_0}}{\sqrt{2 \hat u - \hat r_1}}    
\end{equation}

\subsection{Saddle Point Equations}

Now, we take the limit \(n \to 0\). First for \(F\):
\begin{align}
    f(q_0, q_1, \hat q_0, \hat q_1, r_0, r_1, \hat r_0, \hat r_1, u, \hat u) = \lim_{n \to 0} \, \frac{1}{n} F(q_0, q_1, \hat q_0, \hat q_1, r_0, r_1, \hat r_0, \hat r_1, u, \hat u) = \nonumber \\
    = \tfrac12 (m-1) (q_1 r_1 - q_1 \hat q_1 - r_1 \hat r_1) - \tfrac12 m (q_0 r_0 - q_0 \hat q_0 - r_0 \hat r_0) - \tfrac{1}{2} u (2 \hat u - 1)
\end{align}
Then for \(\Psi\). Use \(A\) to denote the expression that is raised to the power \(m\) in \eqref{eq:Psi_simplified_1rsb}. We can expand:
\begin{align}
    \log \int Dz_0 \, Dy_0 &\Bigl[ \int Dz \, Dy \, A^m \Bigr]^{\frac{n}{m}} = \\
    &= \log \int Dz_0 \, Dy_0 \exp\left(\frac{n}{m} \log \int Dz \, Dy \, A^m\right) = \\
    &= \log \Bigl[ 1 + \frac{n}{m} \int Dz_0 \, Dy_0 \left( \log \int Dz \, Dy \, A^m + O(n) \right) \Bigr] = \\
    &= \frac{n}{m} \int Dz_0 \, Dy_0 \log \int Dz \, Dy \, A^m + O(n^2)
\end{align}
Thus, we have:
\begin{align}
    &\psi(\hat q_0, \hat q_1, \hat r_0, \hat r_1, \hat u) = \lim_{n \to 0} \, \frac{1}{n} \Psi(\hat q_0, \hat q_1, \hat r_0, \hat r_1, \hat u) = - \tfrac12 \hat q_1 \, + \\
    &+ \frac{1}{m} \int Dz_0 \, Dy_0 \log \int Dz \, Dy \, \left( \sum_{s = \pm 1} e^{\left(z_0 \sqrt{\hat q_0} + z \sqrt{\hat q_1 - \hat q_0}\right) s} \left[ e^{-\beta} H(-x_*) + e^\beta H(x_*) \right] \right)^m \nonumber
\end{align}

We have obtained that, in the 1RSB ansatz, the free entropy is given by:
\begin{align}
    \frac{1}{N} \, \mathbb{E}_J \log Z_J = - \beta + f(q_0^*, q_1^*, \hat q_0^*, \hat q_1^*, r_0^*, r_1^*, \hat r_0^*, \hat r_1^*, u^*, \hat u^*) + \psi(\hat q_0^*, \hat q_1^*, \hat r_0^*, \hat r_1^*, \hat u^*)
\end{align}
where \(q_0^*, q_1^*, \hat q_0^*, \hat q_1^*, r_0^*, r_1^*, \hat r_0^*, \hat r_1^*, u^*, \hat u^*\) are the global maximizers of \(f + \psi\).

Since \(\psi\) is independent of \(q_0, q_1, r_0, r_1, u\), we can immediately deduce:
\begin{align}
    \frac{\partial f}{\partial q_0} &= \tfrac12 m \hat q_0 - \tfrac12 m r_0 = 0  \;\;\Longrightarrow\;\; \hat q_0 = r_0 \\
    \frac{\partial f}{\partial q_1} &= \tfrac12 m r_1 - \tfrac12 m \hat q_1 = 0  \;\;\Longrightarrow\;\; \hat q_1 = r_1 \\
    \frac{\partial f}{\partial r_0} &= - \tfrac12 m q_0 + \tfrac12 m \hat r_0 = 0  \;\;\Longrightarrow\;\; \hat r_0 = q_0 \\
    \frac{\partial f}{\partial r_1} &= - \tfrac12 m \hat r_1 + \tfrac12 m q_1 = 0  \;\;\Longrightarrow\;\; \hat r_1 = q_1 \\
    \frac{\partial f}{\partial u} &= - \hat u + \tfrac12 = 0  \;\;\Longrightarrow\;\; \hat u = \tfrac12
\end{align}
Taking these relations into account, we can simplify the objective function:
\begin{align}
    \frac{1}{N} \, \mathbb{E}_J \log Z_J &= \max_{q_0, q_1, r_0, r_1} \left[ - \beta + \frac{1}{2} (1-m) r_1 q_1 + \phi(r_0, q_0, r_1, q_1) \right] \\
    \phi(r_0, q_0, r_1, q_1) &= - \tfrac12 r_1 \, + \nonumber \\
    + \frac{1}{m} \int Dz_0 \, &Dy_0 \log \int Dz \, Dy \, \left( \sum_{s = \pm 1} e^{\left(z_0 \sqrt{r_0} + z \sqrt{r_1 - r_0}\right) s} \left[ e^{-\beta} H(-x_*) + e^\beta H(x_*) \right] \right)^m \nonumber
\end{align}
where
\begin{equation}
    x_* = \frac{-J_D - s y_0 \sqrt{q_0} - s y \sqrt{q_1 - q_0}}{\sqrt{1 - q_1}}
\end{equation}

This can be further simplified if we assume, like in the RS case, that \(q_0 = r_0 = 0\). Then, we have:
\begin{align}
    &\frac{1}{N} \, \mathbb{E}_J \log Z_J = \max_{q_1, r_1} \left[ - \beta + \frac{1}{2} (1-m) r_1 q_1 + \phi(r_1, q_1) \right] \\
    &\phi(r_1, q_1) = - \tfrac12 r_1 + \frac{1}{m} \log \int Dz \, Dy \, \left( \sum_{s = \pm 1} e^{s z \sqrt{r_1}} \left[ e^{-\beta} H(-x_*) + e^\beta H(x_*) \right] \right)^m
    \label{eq:1rsb_free_entropy}
\end{align}
and
\begin{equation}
    x_* = \frac{-J_D - s y \sqrt{q_1}}{\sqrt{1 - q_1}}
\end{equation}
We can write down the saddle point equations, extremizing the action with respect to \(q_1\), \(r_1\), and \(m\), and solve them numerically
using the same methods as in the RS case. We again recover, in the region \(J_D > 0\), the annealed free energy density.

We have obtained a description of the number of fixed points as a function of the self‑interaction strength \(J_D\).
In the next section, we will use the results of the 1RSB analysis, mapping them onto a variant of the large-deviation
introduced in Section \ref{sec:local_entropy}, to probe the existence of dense clusters of fixed points.

\section{Large-Deviation Analysis}

\subsection{Studying the Local Entropy through 1RSB}

We can use the results of the 1RSB analysis to study a variant of the large-deviation
ensemble introduced in Section \ref{sec:local_entropy} within the RS ansatz. In this section, we show exactly
why this is possible, and how the mapping should be performed.

We start from a variant of \eqref{eq:replicated_partition_function} where \(y\) \emph{real}
replicas of the system interact pairwise without mediation through the reference configuration,
and the distance constraint is hard:
\begin{equation}
    Z_J(\beta, D, y) = \sum_{\{s^c_i\}} \exp\left(- \beta \sum_{c=1}^y E_J(s^c)\right) \prod_{c<d} \delta(d(s^c, s^d) - N D)
    \label{eq:large_deviation_partition_function}
\end{equation}
where \(d(s^c, s^d)\) is the Euclidean distance between configurations \(s^c\) and \(s^d\), and \(D\) is an external parameter. To lighten
the notation we are denoting with \(\beta\) what used to be \(\beta'\) in \eqref{eq:replicated_partition_function}.
Note that here \(y\) is constrained to be an integer, but we can always perform an analytic continuation to go back to an arbitrary real
positive inverse temperature once we have an expression for all integers values of \(y\).
We denote with letters \(c, d\) the indexes of the real replicas, and we will use \(a, b\) for the virtual replicas
of the replica trick, consistently with the notation of the previous sections.

We can introduce fields \(h^c_i = \sum_{j \neq i} J_{ij} s^c_i s^c_j\) and their conjugates \(\hat h^c_i\) to rewrite
the partition function as (compare with \eqref{eq:Z_integral}):
\begin{align}
    Z_J(\beta, D, y) = e^{-\beta y N}
    &\int \prod_{c, i} \frac{d h_i^c\,d\hat h_i^c}{2\pi i}
    \exp\Bigl[\, - \sum_{c, i} \hat h_i^c\,h_i^c + \beta \sum_{c, i} \Sign\Bigl( J_D + h_i^c \Bigr) \Bigr] \nonumber\\
        &\times  \sum_{\{s_i^c\}} \exp \Bigl( \, \sum_{c, i} \hat h_i^c s_i^c \sum_{j\neq i}J_{ij}s_j^c \Bigr).
\end{align}
Replicating the partition function \(n\) times, and performing the Gaussian integral over the disorder as in 
\eqref{eq:disorder_average}, we obtain (neglecting terms of order \(O(1/N)\)):
\begin{align}
    \mathbb{E}_J Z_J(\beta, D, y)^n = e^{-\beta n y N} 
    &\int \prod_{a,c,i} \frac{d h_i^{ac}\,d\hat h_i^{ac}}{2\pi i}
    \exp\Bigl[\, - \sum_{a,c,i} \hat h_i^{ac}\,h_i^{ac} + \beta \sum_{a,c,i} \Sign\Bigl( J_D + h_i^{ac} \Bigr) \Bigr] \nonumber\\
        &\times \sum_{\{s_i^{ac}\}} \exp \Bigl( \, \frac{1}{2N} \sum_{a,b,c,d} (\sum_i \hat h_i^{ac} \hat h_i^{bd} s_i^{ac} s_i^{bd})
        (\sum_j s_j^{ac} s_j^{bd}) \Bigr) \nonumber\\
        &\times \prod_a \prod_{c<d} \delta\Bigl( d(s^{ac}, s^{ad}) - N D \Bigr).
\end{align}
Compare this with \eqref{eq:Z_disorder_average}.

Now, we can introduce order parameters \(Q^{ac, bd}\), \(R^{ac, bd}\) and their conjugates. Analogously to the non-replicated case,
we define:
\begin{align}
    Q^{ac, bd} &= \frac{1}{N} \sum_i s_i^{ac} s_i^{bd} \quad \text{for every pair} \;(ac, bd) \; \text{unordered}, \; ac \neq bd; \\
    R^{ac, bd} &= \frac{1}{N} \sum_i \hat h_i^{ac} \hat h_i^{bd} s_i^{ac} s_i^{bd} \quad \text{for every pair} \;(ac, bd) \; \text{unordered}.
\end{align}
and \( \hat Q^{ac, bd}, \hat R^{ac, bd}\) their conjugates. We also set \(Q^{ac, ac} = 1\) and \(\hat Q^{ac, ac} = 0\) for every \(a, c\). Let \(A = \{(ac, bd) \text{ unordered } | \, ac \neq bd\}\) and \(B = \{(ac, bd) \text{ unordered } \}\).
We can rewrite (compare this with \eqref{eq:order_parameters}): 
\begin{align}
    &\; \exp \Bigl( \, \frac{1}{2N} \sum_{a,b,c,d} (\sum_i \hat h_i^{ac} \hat h_i^{bd} s_i^{ac} s_i^{bd})
        (\sum_j s_j^{ac} s_j^{bd}) \Bigr) = \int \prod_{(ac,bd) \in A} \frac{dQ^{ac,bd}\,d\hat Q^{ac,bd}}{2 \pi i} \nonumber \\
    &\quad \times \int \prod_{(ac,bd) \in B} \frac{dR^{ac,bd}\,d\hat R^{ac,bd}}{2 \pi i} \exp\Bigl[\tfrac12 N \sum_{a,b,c,d} \Bigl( Q^{ac,bd}R^{ac,bd} - Q^{ac,bd} \hat Q^{ac,bd} - R^{ac,bd} \hat R^{ac,bd} \Bigr) \Bigr] \nonumber \\
    &\quad \times \exp \Bigl[ \tfrac12 \sum_{a,b,c,d,i} \Bigl( \hat Q^{ac,bd} s_i^{ac} s_i^{bd} + \hat R^{ac,bd} \hat h_i^{ac} \hat h_i^{bd} s_i^{ac} s_i^{bd} \Bigr) \Bigr] \nonumber \\
    &\quad \times \exp \Bigl[ \tfrac12 \sum_{a,c} \Bigl( - N R^{ac,ac} \hat R^{ac,ac} + \hat R^{ac,ac} \sum_i (\hat h_i^{ac})^2 \Bigr) \Bigr]
\end{align}
And:
\begin{align}
    \; &\prod_a \prod_{c<d} \delta\Bigl( d(s^{ac}, s^{ad}) - N D \Bigr) = \nonumber \\
    = &\prod_a \prod_{c<d} \delta\Bigl( N (Q^{ac, ac} + Q^{ad, ad} - 2 Q^{ac, ad}) - N D \Bigr) 
\end{align}

At this point, we can write down the ansatz (assuming replica symmetry):
\begin{align}
    Q^{ac, ac} &= 1 \quad \text{for every } a, c; \\
    Q^{ac, ad} &= q_1 \quad \text{for every } a, c, d \text{ with } c \neq d; \\
    Q^{ac, bd} &= q_0 \quad \text{for every } a, c, b, d \text{ with } a \neq b \\
    R^{ac, ac} &= u \quad \text{for every } a, c; \\
    R^{ac, ad} &= r_1 \quad \text{for every } a, c, d \text{ with } c \neq d; \\
    R^{ac, bd} &= r_0 \quad \text{for every } a, c, b, d \text{ with } a \neq b; \\
    \hat Q^{ac, ac} &= 0 \quad \text{for every } a, c; \\
    \hat Q^{ac, ad} &= \hat q_1 \quad \text{for every } a, c, d \text{ with } c \neq d; \\
    \hat Q^{ac, bd} &= \hat q_0 \quad \text{for every } a, c, b, d \text{ with } a \neq b; \\
    \hat R^{ac, ac} &= \widehat{u} \quad \text{for every } a, c; \\
    \hat R^{ac, ad} &= \widehat{r}_1 \quad \text{for every } a, c, d \text{ with } c \neq d; \\
    \hat R^{ac, bd} &= \widehat{r}_0 \quad \text{for every } a, c, b, d \text{ with } a \neq b.
\end{align}

Now, stop and ponder. The expression for the \(n\)-th moment of the replicated partition function
has almost the same structure as that for the non-replicated case, except for these differences:
\begin{itemize}
    \item Instead of having \(n\) replicas, we have \(n y\) replicas.
    \item There is an extra factor given by the distance constraints.
\end{itemize}
The rest of the terms all look exactly the same as in the non-replicated case. Furthermore,
the structure of the RS ansatz we just made is exactly the same as the 1RSB ansatz in the non-replicated case.
Consider for example the ansatz for the overlaps \(Q^{ac, bd}\). We have \(n y\) terms whose value
is fixed by the norm constraints and equal to 1; we have \(n y (y-1)\) terms that share the same value \(q_1\);
and we have \(n (n-1) y^2\) terms that share the same value \(q_0\). It is clear that this is exactly the same situation
that we had in \eqref{eq:1rsb_ansatz}, mapping \(n\) to \(n y\) and \(m\) to \(y\).

The only difference comes from the distance constraints, which are absent in the 1RSB analysis of the non-replicated case.
Imposing the RS ansatz, we get:
\begin{align}
        &\prod_a \prod_{c<d} \delta\Bigl( N (Q^{ac, ac} + Q^{ad, ad} - 2 Q^{ac, ad}) - N D \Bigr) = \\
        = &\prod_a \prod_{c<d} \delta\Bigl( 2 N - 2 N q_1 - N D \Bigr)
\end{align}
Thus, the additional distance constraints are forcing the value of \(q_1\) to be \(q_1 = 1 - \frac{D}{2}\).
This removes a degree of freedom compared to the 1RSB ansatz: \(q_1\) becomes an external parameter, which
is linked to the distance \(D\) in the replicated partition function. And the Parisi parameter \(m\) corresponds
to the number of real replicas \(y\) in the replicated partition function.

In conclusion, we can study the large-deviation ensemble \eqref{eq:large_deviation_partition_function}
by extremizing \eqref{eq:1rsb_free_entropy} with respect to the order parameters, treating \(q_1 = 1 - \frac{D}{2}\)
as an external parameter, and with the correspondence \(m = y\).

\subsection{Understanding the Structure of Fixed Points}
\label{sec:structure_fixed_points}

We can now use the results of the 1RSB analysis to study the large-deviation ensemble \eqref{eq:large_deviation_partition_function}.
Redefine:
\begin{align}
    \phi(r_1, q_1, m, \beta, J_D) = &- \beta + \frac{1}{2} (1-m) r_1 q_1 - \frac{1}{2} r_1 + \nonumber \\
    &+ \frac{1}{m} \log \int Dz \, Dy \, \left( \sum_{s = \pm 1} e^{s z \sqrt{r_1}} \left[ e^{-\beta} H(-x_*) + e^\beta H(x_*) \right] \right)^m \\
    x_* = \frac{-J_D - s y \sqrt{q_1}}{\sqrt{1 - q_1}}\,&
\end{align}
Here, \(q_1, \beta, J_D\) are treated as external parameters, while \(r_1\) must be extremized. As for \(m\), we will seek the value \(m^*(q_1, \beta, J_D)\)
such that the external entropy \eqref{eq:external_entropy} becomes 0. This is sometimes called the zero-complexity condition.
In the simplest case \eqref{eq:local_entropy_free_energy},
the external entropy captures the number of reference configurations at a local entropy level which is determined
by the inverse temperature. Given our mapping onto the 1RSB formalism, this local entropy level is determined by \(m\):
the larger \(m\), the lower the temperature in the large-deviation ensemble, the higher the average local entropy. The external
entropy crossing 0 at \(m^*\) thus signals that there is only one reference configuration at the local entropy level corresponding to 
the value \(m^*\) for the Lagrange multiplier \(m\). We will then be interested in the value of the local entropy at
\(m^*\), which can be roughly interpreted as the local entropy of the most entropic cluster of fixed points in the landscape.
All of this depends on the choice of \(q_1\), which fixes the distance in the large-deviation ensemble, and of \(J_D\). We will
be most interested in the behavior of \(S_I(q_1, m^*(q_1, \beta, J_D))\) as \(q_1\) and \(J_D\) vary.

Call \(f(q_1, m)\) the free energy, obtained by extremizing \(\phi\) with respect to \(r_1\).
To achieve our goal, we will make use of the thermodynamic relations:
\begin{align}
    S_I(q_1, m) = - \frac{\partial \left[ m \, f(q_1, m) \right]}{\partial m} \\
    S_E(q_1, m) = - m (f(q_1, m) + S_I(q_1, m))
    \label{eq:thermodynamic_local_entropy}
\end{align}
These are standard, once one realized that \(S_I\) and \(S_E\) are respectively the negative of
the formal energy and the formal entropy of the large deviation ensemble (see also Section \ref{sec:stat_phys}).
We can solve numerically for \(m^*(q_1, \beta, J_D)\) by imposing the condition:
\begin{equation}
    f(q_1, m^*) = \frac{\partial \left[ m^* \, f(q_1, m^*) \right]}{\partial m}
\end{equation}
This allows us to immediately compute \( S_I(q_1, m^*) \) using \eqref{eq:thermodynamic_local_entropy}.

\begin{figure}[htbp]
  \centering
  \includegraphics[width=0.8\linewidth]{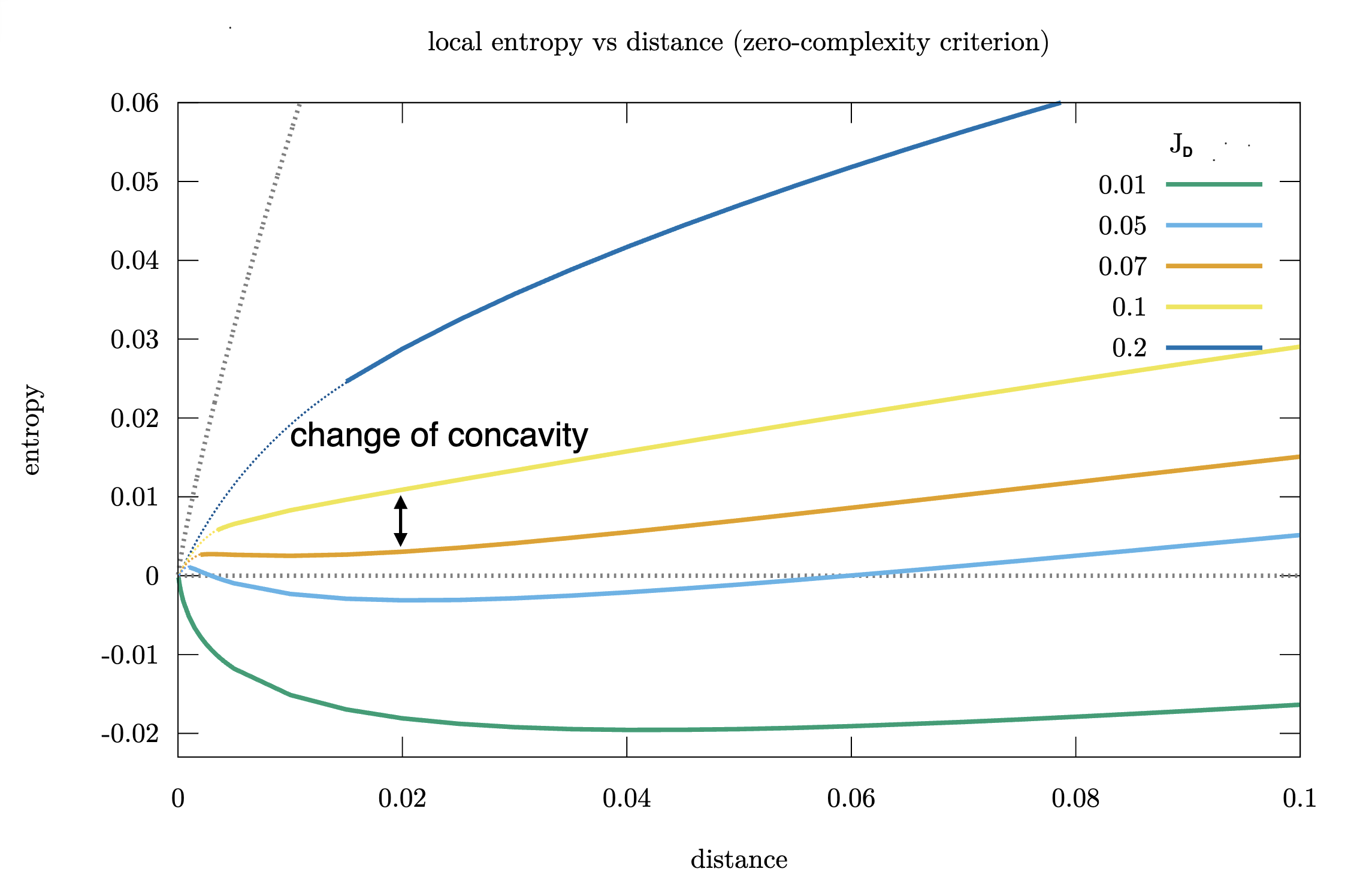}
  \caption{Local entropy curves \(S_I(q_1, m^*)\) as a function of the distance \(\frac{1 - q_1}{2}\) for different
  values of the self-interaction strength \(J_D\). \(m^*\) is chosen to satisfy the zero-complexity criterion. The steep
  dotted line corresponds to the limit \(J_D \to \infty\), in which all states are fixed points. Dotted parts of the curves
  are regions of numerical instability. The curves stop being monotonic for \(J_D\) between 0.07 and 0.1.}
  \label{fig:local_entropy_curves}
\end{figure}

By doing so for a few different values of \(J_D\), sweeping \(q_1\) between 0 and 0.1, we obtain the curves shown in
Figure \ref{fig:local_entropy_curves}. We observe that there is a critical value of \(J_c\) of \(J_D\), which at this granularity
can be estimated to be slightly lower than 0.1, at which the curves stop being monotonic and start exhibiting a maximum
at a certain value of \(q_1\). This signals that, for \(0 \le J_D \le J_c\), isolated solutions coexist with an
exponential number of clusters of fixed points. By computing the value of \(q_1\) at which the maximum occurs, which
belongs to the interval \([0.998, 1.0]\), we see that these clusters are very narrow: in the region below \(J_c\), the
overlap-gap property (OGP) is satisfied.

Above the critical value \(J_c\), instead, the local entropy curves continue increasing as the distance \(\frac{1 - q_1}{2}\) increases, signaling
the appearance of extended dense regions connecting the narrow clusters of fixed points.
Since the \(m^*\) satisfying the zero-complexity criterion is larger than 1, the non-isolated solutions are subdominant
in the equilibrium measure, which is dominated by the isolated solutions, and thus they do not contribute to the entropy
density of fixed points computed with the replica-symmetric analysis.

\subsubsection{Accessing the Fixed Points}

It is natural to wonder, given a realization of the disorder in a finite size network, under what conditions a 
dynamical rule can access the fixed points. To answer this question, we considered two dynamical rules. 
The first is a simple, asynchronous dynamics that aligns a neuron with the sign of its local field:
\begin{equation}
    s_i(t+1) = \Sign\left( J_D s_i(t) + \sum_{j \neq i} J_{ij} s_j(t) \right)
    \label{eq:simple_dynamics}
\end{equation}
The second is a more complex decoding dynamics called focusing Belief Propagation (fBP) \cite{baldassiUnreasonableEffectivenessLearning2016}, a variant of the BP algorithm
that has a bias towards highly locally entropic regions of the landscape.

In the region \(J_D < J_c\), the clusters of fixed points are too far apart compared to their diameter, and therefore
they tend to be inaccessible. In fact, neither the simple dynamics nor the fBP algorithm can find them: both produce
chaotic trajectories that do not converge to any fixed point.

In the region \(J_D > J_c\), the situation is different. In fact, in this case the fBP algorithm is able to converge
to the fixed points, by virtue of its explicit bias, while the simple dynamics still produces chaotic trajectories
and fails to converge.

\begin{figure}[htbp]
  \centering
  \includegraphics[width=0.8\linewidth]{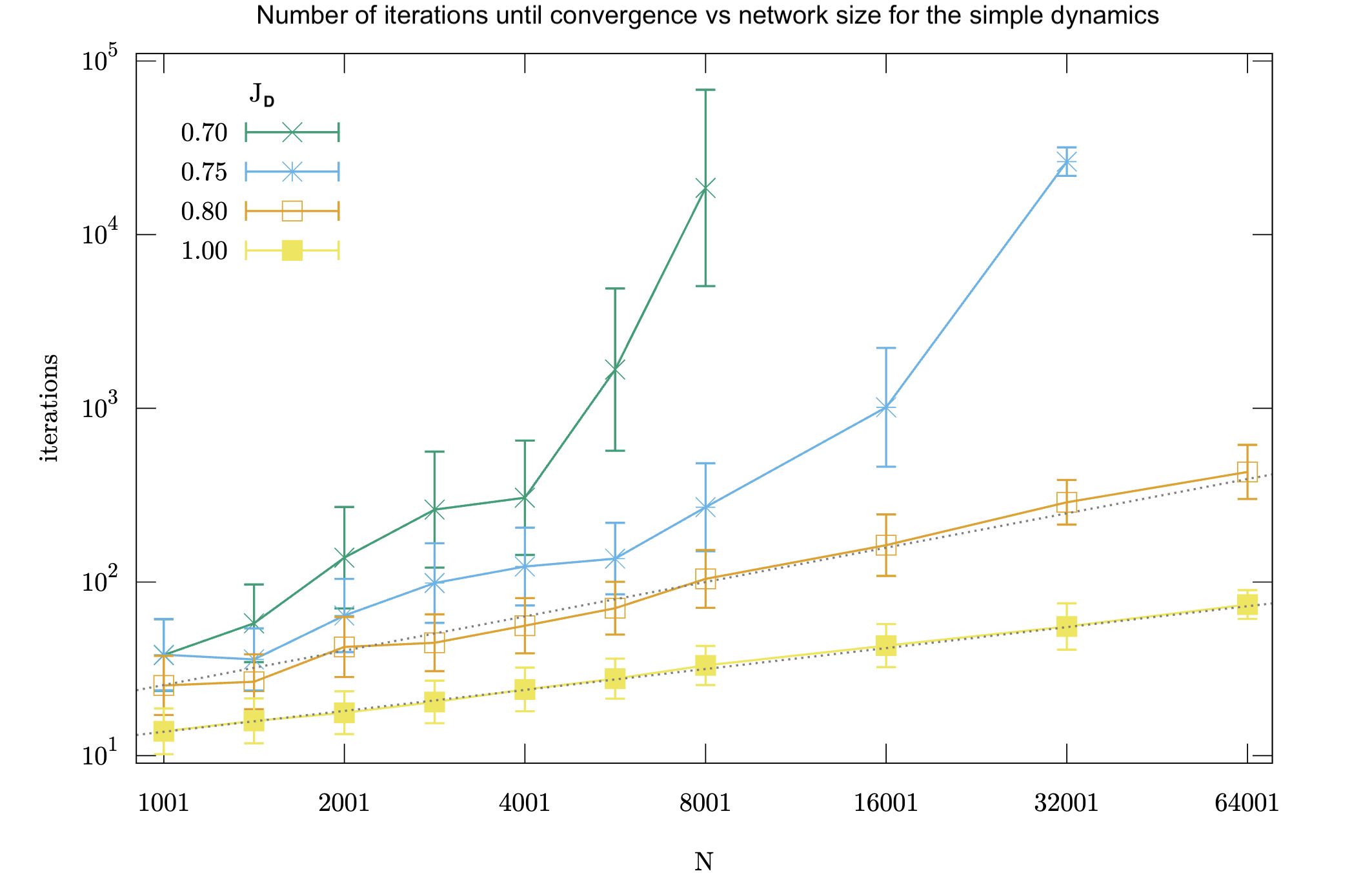}
  \caption{Number of iterations needed by the dynamics \eqref{eq:simple_dynamics} to converge to a fixed point as a function
  of the size \(N\) of the network, for different values of the self-interaction strength \(J_D\). For pairs of values \((N, J_D)\)
  for which we do not observe convergence, we do not plot the number of iterations.}
  \label{fig:simple_dynamics_convergence}
\end{figure}

Finally, there is a threshold value \(J_a(N)\), scaling slowly with the size \(N\) of the network, after which the simple
dynamics is able to converge to the fixed points. For values of \(N\) of the order of \(10^4\), this threshold
is around 0.75. This can be seen in Figure \ref{fig:simple_dynamics_convergence}, where we plot the number of iterations
needed by the simple dynamics to converge to a fixed point as a function of the size \(N\), for different values of \(J_D\):
at a fixed value of \(N\), when \(J_D\) becomes small enough the dynamics fails to converge.

\chapter{A Biologically-Plausible Learning Algorithm for RNNs}
\label{ch:numerical}
% - Benchmarks: MNIST aggrovigliato, Fashion MNIST, MNIST, CIFAR10, SVHN
% - 1 hidden layer performance (against baselines) and analysis
% - lambda same role as JD -> flexibility of architecture (hetero associations, multi-layer classification)
% NOTE: need to specialize algo to 1 hidden layer and to chained multi-layer. need to explain role of JD and lambda. This requires
% introducing some notation (layered structure, etc.) and rewriting update rules, essentially (keep pseudocode)

In this chapter, we introduce a novel, local and distributed algorithm for supervised learning in recurrent neural networks (RNNs).
The algorithm is based on the simple idea of mapping input-output pairs to fixed points of the network dynamics, using synaptic plasticity.
We show that the algorithm is very flexible, in that it can be successfully applied to a wide range of network architectures, and we
investigate its performance in two settings: classification and hetero-associations.

The skeleton of the algorithm bears some similarities with predictive coding, discussed in Section \ref{sec:predictive_coding}. During training,
the network is presented with an input-output pair, which influences its dynamics through external fields. The network is allowed
to relax to a fixed point under their influence, and a local synaptic plasticity rule adjusts the weights so that the stability
margin of the fixed point is increased. At inference time, the network is presented with just an input, the dynamics runs, and
the output is read from the internal state of the network.

It would seem, from this brief description, that the algorithm requires the network dynamics to converge to a fixed point at
initialization, which might be limiting. However, as it turns out, this is not the case: one can use the same plasticity rule 
with a network that initially does not converge, by truncating the dynamics after a number of steps, and over time the synaptic 
plasticity will ensure that the network starts converging consistently.

\section{Algorithm Description}
\label{sec:algorithm_description}

In this section, we provide a general description of the algorithm, which is applicable to a wide range of network architectures.
Later, we will specialize it to some topologies of interest.
For concreteness, here we describe the algorithm in the classification setting. We are provided with a dataset \(\{(x_i, y_i)\}_{i=1}^P\)
of input-output pairs, where \(x_i\) is a vector in \(\mathbb{R}^N\) and \(y_i\) is a one-hot vector in \(\mathbb{R}^C\) encoding the true label.
The network architecture is a recurrent neural network with an arbitrary connection topology. The network receives feedback from two sources:
the input \(x_i\), through a projection matrix \(W^{in}\), and a readout layer \(z \in \mathbb{R}^C\), through a projection matrix \(W_{back}\).
The readout layer is affected by the internal state of the network through a projection matrix \(W^{out}\). These projections in general only affect 
a subset of the neurons; however, in this abstract description, it will be convenient for notational purposes to consider \(W^{in}\) and \(W^{back}\)
projecting to all neurons, and \(W^{out}\) reading from all neurons (thus they will be sparse, in the notation of this section).

The recurrent network is composed of \(M\) neurons, each with a state variable \(s_i \) which, for simplicity we assume to be binary: \(s_i \in \pm 1\).
In future work, we will consider the extension to continuous states.
We consider a simple dynamical rule akin to
\eqref{eq:simple_dynamics}: denoting the synaptic weight from neuron \(j\) to neuron \(i\) with \(J_{ij}\),
each neuron aligns with its local field:
\begin{align}
    s_i(t+1) &= \Sign \left( \sum_{j=1}^M J_{ij} s_j(t) + \sum_{j=1}^N W^{in}_{ij} x_j + \sum_{c=1}^C W^{back}_{ic} z_c \right), \\
    z_c(t+1) &= \Sign \left( \sum_{i=1}^M W^{out}_{ic} s_i(t) + \lambda_y y_c \right).
\end{align}
This can be done asynchronously, as in the experiments of Section \ref{sec:structure_fixed_points}, or synchronously, updating all
neurons at once. The synchronous dynamics is advantageous in that it allows to exploit the parallelism of modern hardware, but
the asynchronous one is more appealing from a biological perspective, since it is completely local and distributed. Based on our experiments,
both are viable. The dynamics is terminated upon convergence, or after a maximum number of updates is reached.
The input and target affect the dynamics through the terms \(W^{in}_i x_i\) and \(\lambda_y y_c\), respectively.

Of course, the term \(\lambda_y y_c\) is only present during training, effectively clamping the readout layer to the true label,
as in Predictive Coding. During inference, the network is allowed to evolve freely, influenced only by the input \(x_i\),
and the readout reads the output from the internal states after convergence. To reduce the gap between training and inference,
we have found it beneficial to have a short warm-up phase in the dynamics, during which the importance of the feedback from
the readout is set to zero.

As for the synaptic plasticity rule, given the state of the network after the relaxation, denoted by an asterisk, we use the following
update rule:
\begin{equation}
    J_{ij} \leftarrow J_{ij} + \eta \, s_i^* \, s_j^* \; \mathbf{1}(s_i^* \, f_i \le k), \quad i,j = 1, \ldots, M
    \label{eq:plasticity_rule}
\end{equation}
This is inspired by the perceptron learning rule, interpreting each neuron as the output unit of a perceptron
which receives as input the states of all the neurons communicating with it. \(\mathbf{1}(\cdot)\) is the indicator function, 
and \(k, \eta\) are hyperparameters. As for \(f_i\), we consider two variants.
In the first one, during training, throughout the network dynamics, we anneal the importance
of the readout feedback from an initial value down to 0. In this case, \(f_i\) is simply the local field at the end
of the dynamics. In the second case, we avoid the annealing and ignore the contribution of the readout to \(f_i\).
In both cases, the product \(s_i^* \, f_i\) should be interpreted as the stability margin of neuron \(i\) after convergence,
and \(k\) as a desired stability threshold.

The first variant is more plausible from a biological perspective, since it treats all synapses, including connections
with the readout, in the same way. The annealing of the feedback importance could be motivated by interpreting the readout
in our model as a proxy for another, unmodelled network from another area of the brain. The supervisory signal could
then be thought of as unusual activity in that area due to a stimulus, which influences the dynamics of our network through
long-range interactions, of which the interaction with the readout is a caricature. With this interpretation, the annealing
would be motivated by the gradual return to baseline activity of the other area.

Note that our plasticity rule is completely local and distributed, a key ingredient for biological plausibility. Furthermore,
as we will see, we do not require the couplings between the neurons to be symmetric, contrary to approaches like Equilibrium Propagation.

We use the same plasticity rule \eqref{eq:plasticity_rule} also for the other synaptic weights:
\begin{align}
    W^{in}_{ij} \leftarrow W^{in}_{ij} + \eta \, s_i^* \, x_j \; \mathbf{1}(s_i^* \, f_i \le k), &\quad i = 1, \ldots, M, j = 1, \ldots, N \\
    W^{out}_{ci} \leftarrow W^{out}_{ci} + \eta \, z_c^* \, s_i^* \; \mathbf{1}(z_c^* \, f_c \le k), &\quad i = 1, \ldots, M, c = 1, \ldots, C \\
    W^{back}_{ic} \leftarrow W^{back}_{ic} + \eta \, s_i^* \, z_c^* \; \mathbf{1}(s_i^* \, f_c \le k), &\quad i = 1, \ldots, M, c = 1, \ldots, C \\
\end{align}
To help keep the norm of the couplings under control, we include a weight decay term in the update. For example, for the internal
couplings:
\begin{equation}
    J_{ij} \leftarrow (1 - \lambda_{wd}) J_{ij} + \eta \, s_i^* \, s_j^* \, \mathbf{1}(s_i^* \, f_i \le k), \quad i,j = 1, \ldots, M
    \label{eq:plasticity_rule_wd}
\end{equation}

The couplings are initialized in such a way that contributions to the local fields from different sources are of order 1. As for the
network state, we initialize it to a neutral state of all zeros. To further exploit hardware parallelism, we can perform
the dynamics in parallel for a batch of input-output pairs, and update the weights in parallel using an average of the single-pair updates.

The general structure of the algorithm, during training and inference, is summarized in Algorithm \ref{alg:training} and Algorithm \ref{alg:inference} respectively.
We consider the variant with annealing of the readout feedback, and synchronous dynamics. The annealing schedule \(\alpha(t)\) 
captures both the initial warm-up phase and the subsequent annealing of the feedback importance.
For simplicity, we assume a batch size of 1.

The plasticity rule could be replaced by the optimization of a local binary cross-entropy criterion, where we regard each neuron
as the head of a perceptron, and the neurons firing onto it as its inputs. The update is then performed by taking a local gradient
step with respect to this objective, aiming for the input neurons to 'correctly classify' (i.e. stabilize) the output neuron.
Concretely, one can define a Boltzmann distribution over the states of each output neuron:
\begin{equation}
    p(s_i = 1) = \frac{e^{\beta f_i / 2}}{e^{\beta f_i / 2} + e^{-\beta f_i / 2}}
\end{equation}
where \(f_i\) is computed as before. Then, a local gradient ascent step on the log-likelihood of the observed states \(s_i^*\) reads:
\begin{equation}
    J_{ij} \leftarrow J_{ij} + \eta \, \beta \, s_i^* s_j^* \left( 1 - \sigma(\beta \, s_i^* \, f_i) \right)
\end{equation}
where \(\sigma(\cdot)\) is the sigmoid function. This is very similar to the update rule \eqref{eq:plasticity_rule}, but we have
replaced the indicator function with a smoother sigmoidal function, whose steepness we can control by
choosing the inverse temperature \(\beta\). One can even add a shift to the sigmoid argument, obtaining a strict generalization
of \eqref{eq:plasticity_rule}. In our experiments, we have found that this smoother plasticity rule tends to perform similarly
to the other one in the simple benchmarks we have considered.

In the next section, we will discuss specific implementations of the algorithm with different network architectures.

\begin{algorithm}
    \caption{Training for one epoch, agnostic to network architecture.}
    \begin{algorithmic}[1]
        \Require dataset $\mathcal D$, learning rate $\eta$, margin $k$, clamping strength $\lambda_y$, weight decay $\lambda_{wd}$.
        \Require initial values for $J \in \mathbb{R}^{M \times M}, \, W^{\text{in}} \in \mathbb{R}^{M \times N}, W^{\text{out}} \in \mathbb{R}^{C \times M}, W^{\text{back}} \in \mathbb{R}^{M \times C}$.
        \Require integer $T_{max}$, annealing schedule $\alpha(t)$
        \ForAll{$(x,y)\in\mathcal D$}  \Comment{batch size 1 for simplicity}
            \State $s\gets \mathbf{0}$, \ $z\gets \mathbf{0}$ \Comment{$s \in \mathbb{R}^M$, $z \in \mathbb{R}^C$}
            \State $t$ $\gets 0$
            \While{$t$ $<$ $T_{max}$}
                \State $s_{new} \gets\Sign\!\bigl(Js + W^{\text{in}}x + \alpha(t)W^{\text{back}}z\bigr)$
                \State $z_{new} \gets W^{\text{out}}s$ + $\lambda_y y$
                \If{$s = s_{new}$ \textbf{and} $z = z_{new}$}
                    \State \textbf{break}
                \EndIf
                \State $s \gets s_{new}, \quad z \gets z_{new}, \quad t \gets t + 1$
            \EndWhile
            \State $f\gets Js + W^{\text{in}}x + \alpha(T_{max})W^{\text{back}}z, \quad g\gets W^{\text{out}}s$
            \State $J \gets (1 - \lambda_{wd}) J + \eta \, s s^{\!\top} \odot \mathbf{1}(s \odot f \le k)$ \Comment{numpy broadcasting (see \ref{eq:plasticity_rule})}
            \State $W^{\text{in}} \gets (1 - \lambda_{wd}) W^{\text{in}} + \eta \, s x^{\!\top} \odot \mathbf{1}(s \odot f \le k)$
            \State $W^{\text{back}} \gets (1 - \lambda_{wd}) W^{\text{back}} + \eta \, s z^{\!\top} \odot \mathbf{1}(s \odot f \le k)$
            \State $W^{\text{out}} \gets (1 - \lambda_{wd}) W^{\text{out}} + \eta \, z s^{\!\top} \odot \mathbf{1}(z \odot g \le k)$
        \EndFor
    \end{algorithmic}
    \label{alg:training}
\end{algorithm} % Annealing variant, asynchronous dynamics, batch size 1

\begin{algorithm}
    \caption{Inference, agnostic to network architecture.}
    \begin{algorithmic}[1]
        \Require Trained weights $J,\;W^{\mathrm{in}},\;W^{\mathrm{out}}$
        \Require Input vector $x \in \mathbb{R}^N$, integer $T_{max}$
        \State $s \gets \mathbf{0}$, \ $t \gets 0$
        \While{$t < T_{max}$}
            \State $s_{new} \gets \Sign\!\bigl(J\,s + W^{\mathrm{in}}\,x\bigr)$
            \If{$s = s_{new}$}
                \State \textbf{break}
            \EndIf
            \State $s \gets s_{new}$, \ $t \gets t + 1$
        \EndWhile
        \State $z \gets W^{\mathrm{out}}\,s$
        \State \textbf{return} $\arg\max_{c} \;z_c$
    \end{algorithmic}
    \label{alg:inference}
\end{algorithm}

\section{Network Architectures}
\label{sec:network_architectures}

The algorithm as we described it in the previous section does not assume any particular structure for the architecture of
the recurrent neural network. Here, we discuss some choices that can be made in this regard, introducing the architectures
that we will consider in Section \ref{sec:experiment_results}.

In principle, the topology of the network can be arbitrary. However, to facilitate the interpretation of the network states,
and for easier comparison with other approaches, we have mostly focused on networks having a layered structure, and possibly
some skip connections to the input and output layers. In future work, we will consider more complex topologies.

In the layered case, we have a set of \(L \ge 1\) hidden layers of neurons, each containing \(H\) binary neurons \(s_i^\ell\).+
Each individual layer is a recurrent network like the one studied in Chapter \ref{ch:analytical}, initialized with asymmetric
random couplings \(J_{ij}^\ell \sim \mathcal{N}(0, \frac{1}{H})\), and with self-interaction terms \(J_{ii}^\ell = J_D\). From
the analysis of Chapter \ref{ch:analytical}, we know a lot about the existence and structure of the fixed points of the dynamics,
at least at the initialization: once the learning begins, the couplings become correlated, and things could change.

Between the hidden layers, we have longitudinal couplings \(J_{ij}^{\ell,\ell+1}\) from layer \(\ell+1\) to layer \(\ell\),
and \(J_{ij}^{\ell,\ell-1}\) from layer \(\ell-1\) to layer \(\ell\). We will describe how they are handled in each case.
Additionally, the input couplings \(W^{in}\) project from the input to the first hidden layer, the output couplings \(W^{out}\)
read from the last hidden layer into the readout layer, and the feedback couplings \(W^{back}\) project from the readout layer
onto the last hidden layer. In some cases, we will also consider skip connections from the input layer to the other hidden layers,
and from the readout layer to the other hidden layers.

When it comes to the feedback from the readout layer, the simplest option is to keep the matrix \(W^{back}\) frozen. In fact, 
in our algorithm, it is the dynamics itself that propagates "targets" for each layer, through the longitudinal couplings.
With a fixed \(W^{back}\), the network is able to use its plasticity to get in a situation where this initially random feedback
becomes useful, similar to what happens in feedback alignment.
Other options are viable, like for example learning \(W^{back}\) as well, or maintaining it equal to a rescaled version of the transpose of \(W^{out}\).
However, the simple option of keeping it frozen seems to be just as powerful based on our experiments.
Now, we describe some specific architectures that we will consider in the experiments.

\subsection{Single Hidden Layer}

We consider an architecture with a single hidden layer of \(H\) neurons. While this might seem similar to the framework of
reservoir computing, the key difference is that we learn the recurrent connections. This makes the architecture highly non-trivial.
In fact, we are learning three sets of couplings, whose effects on the output compound:
\begin{itemize}
    \item The input couplings \(W^{in} \in \mathbb{R}^{H \times N}\), projecting from the input layer to the hidden layer;
    \item The recurrent couplings \(J^{1} \in \mathbb{R}^{H \times H}\), projecting from the hidden layer to itself;
    \item The output couplings \(W^{out} \in \mathbb{R}^{C \times H}\), projecting from the hidden layer to the readout layer.
\end{itemize}
Without the recurrent couplings, this architecture would be similar to a multi-layer perceptron with one hidden layer, which
is why we refer to it as having "a single hidden layer". However, one should not take the analogy with feed-forward architectures
too far. In fact, in our framework, already a single hidden layer applies multiple steps of linear transformation plus non-linearity,
which can be understood if one imagines unrolling the dynamics in time as a computational graph. Of course, compared to a deep
feed-forward architecture, the sequence of transformations performed by our recurrent network over time shares the same
set of weights.

\subsection{Ferromagnetic Chain}
\label{sec:ferromagnetic_chain}

The simplest multi-layer architecture fitting the description of Section \ref{sec:network_architectures} is obtained
by having a small number of fixed, large longitudinal couplings between the hidden layers, and all the others
equal to zero. This is loosely inspired by the findings of \cite{songHighlyNonrandomFeatures2005a}, which show that networks
of pyramidal neurons in the cortex tend to exhibit few, strong synapses forming a skeleton immersed in a sea of weaker connections. 

As a simple realization of this idea, we consider a ferromagnetic chain, in which between each pair of adjacent hidden layers
there are exactly \(H\) fixed ferromagnetic couplings of order 1, each connecting a different pair of neurons in the two layers.
Note that the number of weak recurrent couplings is much larger, of order \(H^2\). Importantly, for this model, when we use the variant
of the plasticity rule that ignores the readout feedback, we also ignore the contributions to the local fields coming from the
right adjacent layers, to encourage forward propagation of information. This is unnecessary, instead, with the annealing variant.

We use this version of the model for the hetero-association task, described in the next section. To keep the model as simple as
possible, we choose the size of the hidden layers equal to the input size, and we replace the matrices \(W^{in}, W^{out}, W^{back}\)
with non-learnable ferromagnetic couplings as well. This allows to put all the emphasis on the dynamics, which must learn to deform
the input patterns into the output patterns progressively through the layers.

\section{Tasks and Datasets}
\label{sec:tasks_datasets}

In this section, we briefly describe the datasets and tasks that we will consider in the experiments of Section \ref{sec:experiment_results}.

The first task we consider is classification, which provides a good testbed since performance metrics are very
transparent, and network representations can easily be interpreted thanks to the chain topology. For classification, we
consider the MNIST dataset \cite{lecunMNISTDATABASEHandwritten}. To make it more challenging, we consider an entangled version
of MNIST, in which the digits are first projected into a lower-dimensional space using a random matrix of dimension \(N \times 784\),
and then the projected digits are binarized through a sign non-linearity. Notice that, while the initial digits are almost binary
to begin with, after the projection they have a much richer histogram, and therefore the binarization throws away a lot of information.

As a reference point, considering \(N=100\), a perceptron trained with Adam and a cross-entropy loss achieves an accuracy of about 80 \%
on the training and test sets: the dataset is not linearly separable, thus the task is not trivial to solve. At the same time, as
we will see, the data pre-processing still retains enough information to allow a network with hidden layers to significantly
outperform a perceptron. This makes the dataset a good benchmark task for our algorithm. Unfortunately, there is no significant
gain in performance from increasing the depth of the MLP beyond 1 hidden layer, even using backpropagation. For this reason,
we will consider a different task, which we now introduce, to evaluate gains from depth.

The second task we consider is learning hetero-associations: maps from an input of dimension \(N\) to an output of the same dimension.
This is significantly more challenging than classification. In this setting, we are interested in evaluating the capacity of the
network. We define the capacity as the maximum size of a dataset of hetero-associations that can be learned by the network. We say
that a dataset has been learned if, with probability at least \(1 - \delta\) over the choice of a random pair from the dataset, 
the network can reproduce the output given only the input, with an overlap (cosine similarity) of at least \(1 - \epsilon\) with the true output.
For this task, we consider random binary patterns of dimension \(N\), with each dimension sampled uniformly at random from \(\{-1, 1\}\).

\section{Experiment Results}
\label{sec:experiment_results}

\subsection{Learning a Non-Linearly-Separable Dataset}
\label{sec:experiment_results_1_layer}

We consider the Entangled MNIST dataset, described in Section \ref{sec:tasks_datasets}, with \(N=100\). Since the dataset is
not linearly separable, we can use it to verify that our algorithm can learn features useful for classification. We consider
the single-hidden-layer architecture described in Section \ref{sec:network_architectures}. We compare our algorithm against
two baselines:
\begin{itemize}
    \item \textbf{Random Features (+ input)}: we use a random projection matrix of dimension \((H-N) \times N\) to project the input to a high-dimensional space. For fairness, since our model uses binary states, we binarize the projected features using a sign non-linearity. Then, we concatenate the input with the projected features. We use this representation to train a perceptron using the perceptron learning rule, with a very high margin. We have verified that this achieves the same performance as training the perceptron with Adam and a cross-entropy loss.
    \item \textbf{Random Reservoir Features (+ input)}: we do a similar procedure as described above, concatenating the input with random features. However, instead of obtaining the features through random projection, we use a network identical to the one that we train as a reservoir. Given an input, we let the dynamics of the model run until convergence, and then we read the internal state of the network as the features. Throughout the dynamics, the state of the network is influenced by the input through the input couplings, and by the recurrent couplings, which are not learned.
\end{itemize}
The point of comparing against the first baseline is to show that our algorithm is capable of learning useful features. On a dataset
like this one, where there are essentially no gains from depth larger than 1, this is a challenging baseline, especially considering
that we concatenate the input with the features.
The point of the second baseline is to have a direct comparison that allows to isolate the impact of learning the recurrent couplings
in our algorithm. In fact, coupling learning is the only difference between our algorithm and the reservoir baseline.

\begin{figure}[htbp]
  \centering
  \includegraphics[width=0.8\linewidth]{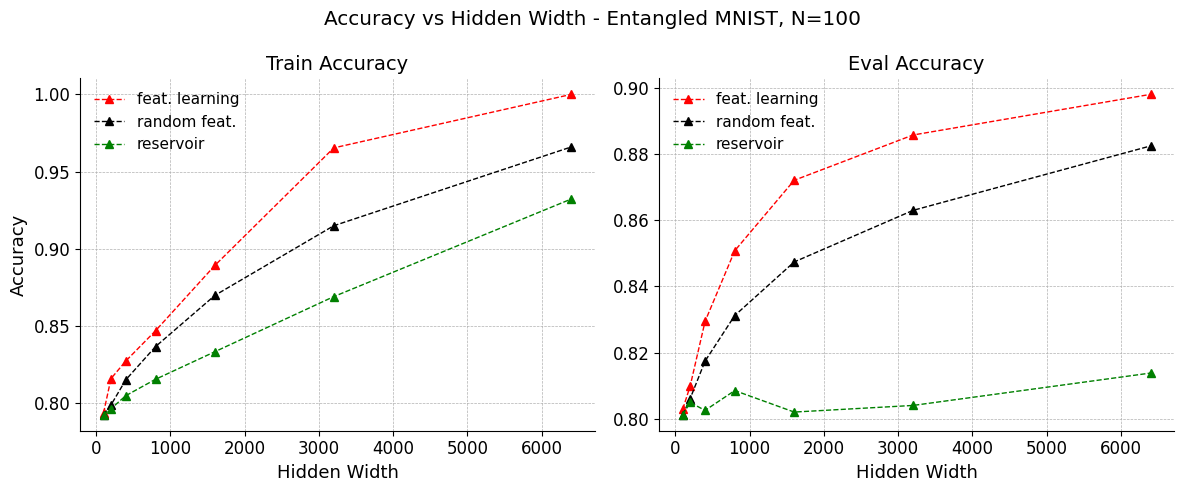}
  \caption{Comparison of the single-hidden-layer model trained with our algorithm against two baselines on the Entangled MNIST dataset. The train
  curve for feat. learning (ours) uses $J_D = 0.7$, while the test curve uses $J_D = 0.3$. The other hyperparameters are shared between the two.
  All models are trained for 100 epochs, and the maximum accuracy is reported.}
  \label{fig:ours_vs_baselines_1_layer}
\end{figure}

For each of the three approaches, we train a single-hidden-layer network multiple times, scaling the number of neurons \(H\). We
report the accuracy on the training set, as well as on a held-out test set, in Figure \ref{fig:ours_vs_baselines_1_layer}. We
notice that our algorithm significantly outperforms both baselines across all values of \(H\). With \(H=6400\), the largest
we tested, our algorithm is able to perfectly fit the training set, with a gap of about 4 \% with respect to the random features
baseline, and over 6 \% with respect to the reservoir baseline. In terms of generalization, we see that the biggest gap
with the random features baseline is obtained for intermediate values of \(H\), with a peak difference close to 3 \%, which
decreases to about 2 \% for \(H=6400\). The reservoir baseline, instead, is essentially unable to generalize, since it
barely outperforms a perceptron (82 \% vs 80 \% for \(H=6400\) - consider that the input is concatenated to the reservoir features). 

These results highlight two important points. First, our algorithm is able to perform feature learning. Second, learning
the recurrent couplings is \emph{very effective}, since not doing it leads to a drop in accuracy on the test set from 90 \% to 82 \%,
essentially losing any edge over what could be achieved without concatenating any features at all to the input (80 \%). 

\begin{figure}[htbp]
  \centering
  \includegraphics[width=0.8\linewidth]{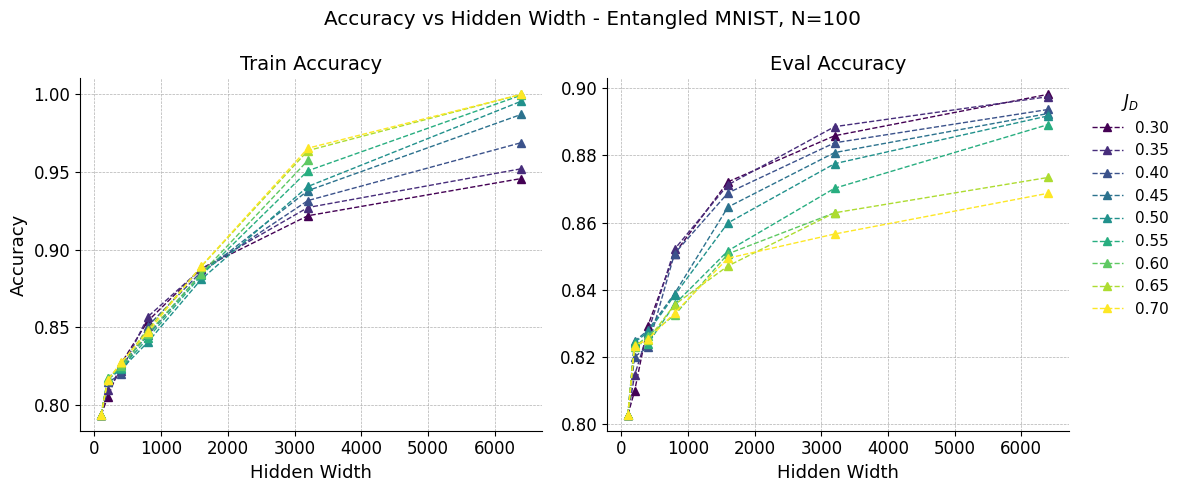}
  \caption{Impact of the choice of \(J_D\) on train and test accuracy for the single-hidden-layer model trained with our algorithm
  on the Entangled MNIST dataset. All models are trained for 100 epochs, and the maximum accuracy is reported.}
  \label{fig:acc_vs_jd}
\end{figure}

In the same setting, we have systematically investigated the impact of the choice of \(J_D\) on train and test performance. In fact, we know
from the theoretical analysis of Chapter \ref{ch:analytical} that the value of \(J_D\) has a strong impact on the structure of the fixed points
and their accessibility to the simple dynamics we use, at least at initialization. To this end, we perform a similar scaling experiment to the previous
one, but focusing on our approach only, and varying \(J_D\). The results are shown in Figure \ref{fig:acc_vs_jd}, and they show
that the choice of \(J_D\) has a very significant impact on the performance of the network. Both train and test
accuracy are maximized at an intermediate value of \(J_D\) but, interestingly, the optimal values do not coincide. 
On the contrary, varying \(J_D\) in the range \([0.3, 0.7]\) has opposite effects on train and test accuracy: train accuracy
increases with \(J_D\), while test accuracy decreases. This trend is very clearly visible and quite stable across
different values of \(H\).

Varying \(J_D\) also has a visible impact on the representations learned by the network. In order to visualize this, we build
a heatmap as follows. We consider a set of 300 randomly selected input patterns from the training set.
For each of them, we do inference with the network and we extract the internal state at the end of the dynamics. Then, for
each pair of inputs, we compute one minus the Hamming distance between the two internal states, and we normalize it by the number of neurons,
obtaining a similarity measure with values in \([0, 1]\): 0 means the two states are opposite, 1 means they are identical, and
0.5 means they are orthogonal. We then plot the resulting similarity matrix as a heatmap, with the inputs ordered by their
label.

\begin{figure}[htbp]
  \centering
  \includegraphics[width=0.8\linewidth]{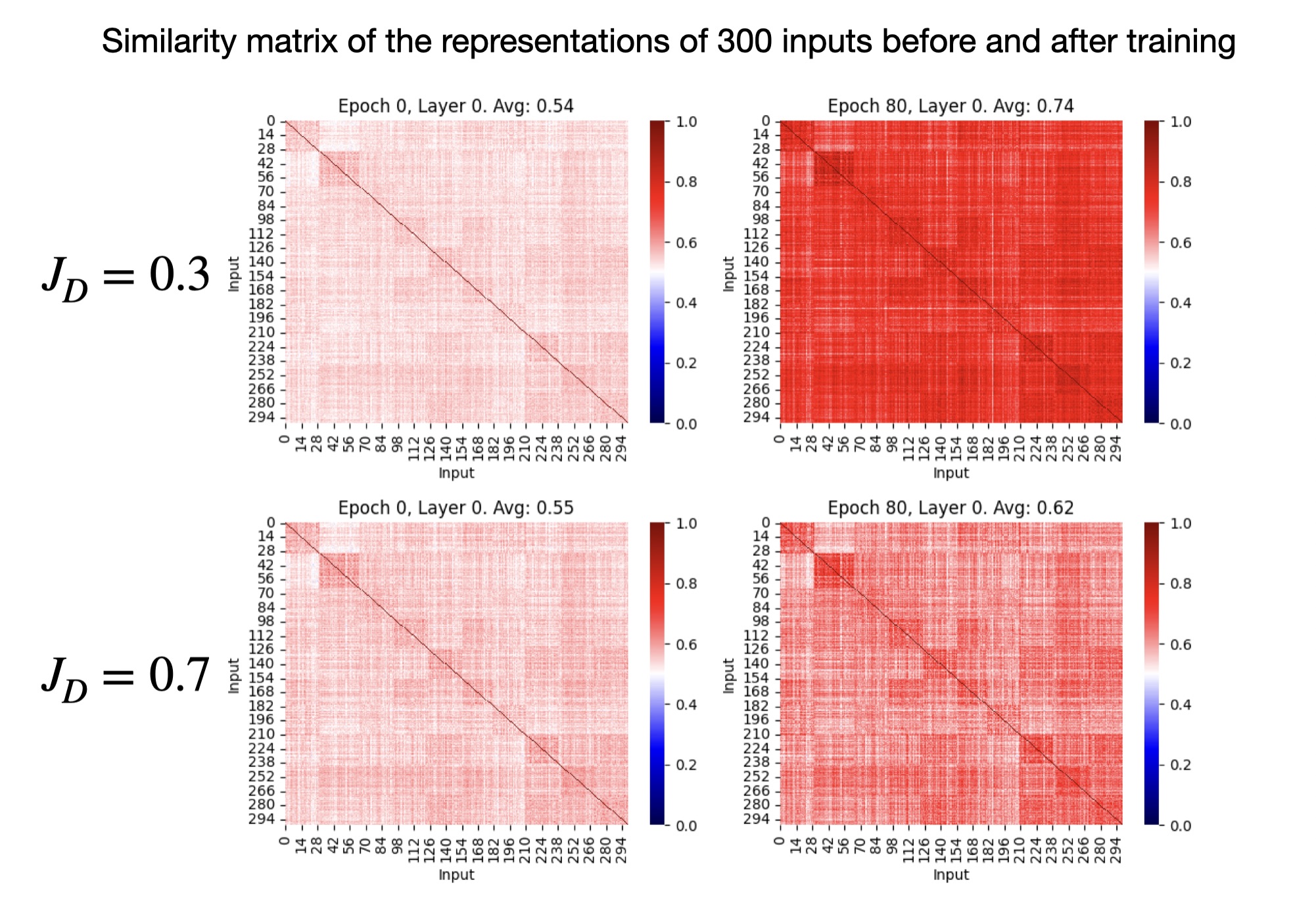}
  \caption{Similarity matrices of the internal representations of the same set of 300 inputs, ordered by their label,
  before and after training a single-hidden-layer model with our algorithm on the Entangled MNIST dataset. $H=6400$.
  Similarity is on a scale from 0 (opposite) to 1 (identical). The average similarity across all pairs of inputs is reported.}
  \label{fig:representations_1layer}
\end{figure}

In Figure \ref{fig:representations_1layer}, we show the heatmaps so computed for \(J_D = 0.3, 0.7\), both before and after training. We can see
that there is no significant difference in the representations before training. After training takes place, the representations
of all inputs tend to become more similar to each other. This is especially true for inputs belonging to the same class.
Interestingly, this effect is significantly more pronounced for \(J_D = 0.3\) than for \(J_D = 0.7\).

\subsection{Gains from Depth: Hetero-Association Capacity}

It is interesting to see whether there is a positive impact on performance from increasing the depth of the network. The hetero-association
task is particularly suitable to detect this, since it is very challenging. We considered networks of variable depths \(L\) having
the architecture described in Section \ref{sec:ferromagnetic_chain}. For each depth, we considered a range of values for \(N\),
which is both the pattern dimension and the network width. This means that, as \(N\) increases, the network becomes more expressive,
but the task also becomes more challenging, since the patterns are higher-dimensional.

For each pair \((L, N)\), we trained the network on datasets of hetero-associations generated as described in Section \ref{sec:tasks_datasets},
using progressively larger dataset sizes \(P\). Each time,
we evaluate whether the network can learn the dataset, using the criterion described in Section \ref{sec:tasks_datasets}, with
\(\delta = \epsilon = 0.05\). This way, we can determine the capacity of the network for that pair \((L, N)\).

\begin{figure}[htbp]
  \centering
  \includegraphics[width=0.8\linewidth]{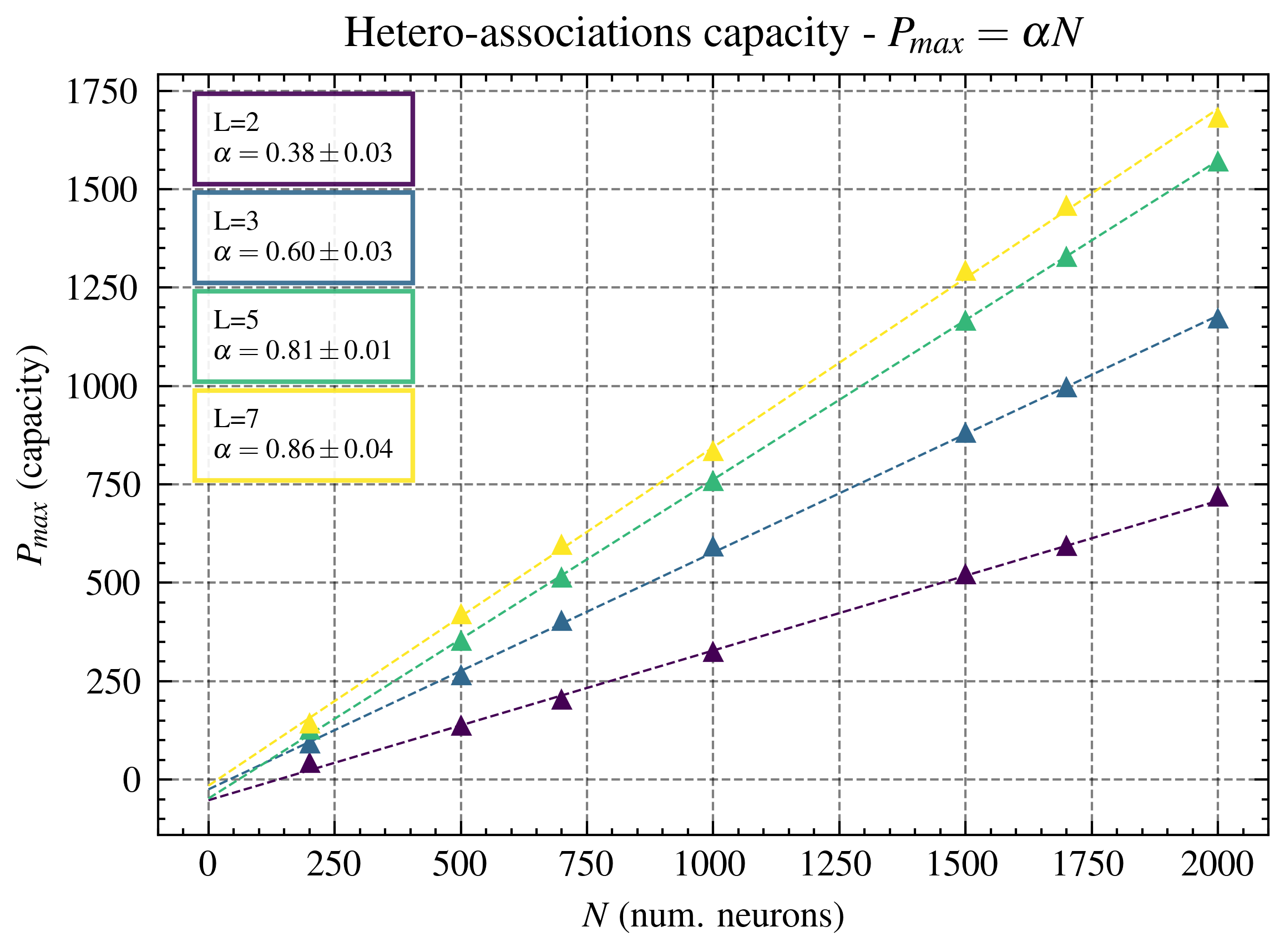}
  \caption{Scaling of the hetero-association capacity with the input dimension \(N\) for different values of the depth \(L\) of the ferromagnetic
  chain architecture. Inputs and outputs are random binary patterns, and the width of the network is also \(N\).}
  \label{fig:hetero_association_capacity}
\end{figure}

Once we have
determined the capacity for a range of values of \(N\), we can attempt to fit a parametric model to the data to estimate
the scaling of the capacity with \(N\), for each value of \(L\). By visual inspection, we found that a straight line approximates
the data quite well, so we fit a linear model of the form \(P_{max} = \alpha N\) using ordinary least squares (OLS) regression.
In Figure \ref{fig:hetero_association_capacity}, we show the results of this experiment, plotting the capacity as a function of \(N\) for different values of \(L\).
We also show the fitted linear models, with the estimated slope \(\alpha\) and its 95 \% confidence interval. The results show
that there are very significant gains in capacity as we increase the depth of the network, with the estimated slope of the linear model
increasing from 0.38 for \(L=2\) to 0.86 for \(L=7\). The gains seem to start saturating when \(L\) becomes larger than 5, which
already achieves a slope of 0.81.

Importantly, while a linear scaling with \(N\) might seem underwhelming, it should be kept in mind that, as already remarked, the task
is also becoming more challenging as \(N\) increases.

\subsection{Comparison between Plasticity Rules}

\begin{figure}[htbp]
  \centering
  \includegraphics[width=0.8\linewidth]{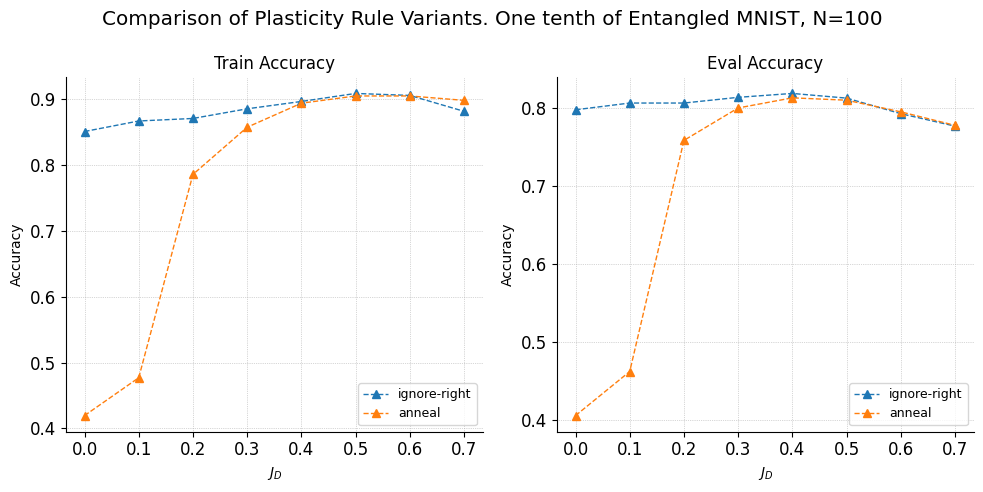}
  \caption{Comparison of the performance of the two variants of the plasticity rule on one tenth of the Entangled MNIST dataset, using
  the single-hidden-layer architecture with \(H=300\) neurons. The annealing variant uses a simple schedule, where the importance
  is kept constant for half the dynamics, then set to 0 for the rest.}
  \label{fig:double_dynamics}
\end{figure}

In this section, we compare the two variants of the plasticity rule described in Section \ref{sec:algorithm_description}, namely the one
with annealing of the readout feedback, and the one which ignores contributions from the right to the local fields. To do so,
we consider the single-hidden-layer architecture described in Section \ref{sec:network_architectures}, and we train it on
one tenth of the Entangled MNIST dataset, for \(J_D\) varying in the range \([0.0, 0.7]\). We use \(H=300\) neurons, 
and we train for 40 epochs. As annealing schedule, we make the simplest possible choice: we keep the importance constant
for half the time, then we set it to 0 for the rest of the dynamics. The results are shown in Figure \ref{fig:double_dynamics}.

We can see that, for \(J_D\) large enough, the two variants achieve similar performance. However, for small values of \(J_D\), the
variant with the annealing performs significantly worse. The exact threshold at which the two variants start to diverge
in performance depends on the strength of the external field coming from the input: with a stronger field, the threshold is lower (not shown).
This is likely because the annealing variant, especially with this extreme schedule, is very much reliant on the existing
structure of the fixed points from the point of view of the inference dynamics, since it needs to converge to a state that is 
already stable without the readout feedback. On the contrary, the other variant can converge to a point that is not stable
without the readout feedback, and then use plasticity to make it stable.

This difference in behavior suggests that, in addition to being more plausible from a biological perspective, the annealing variant
might also have closer links with the theoretical analysis of Chapter \ref{ch:analytical}, since it relies more directly on
the structure of the fixed points at initialization.

\section{More Preliminary Results}
\label{sec:more_preliminary_results}

\subsection{Ferromagnetic Couplings can Substitute the Self-Interaction}
\label{sec:ferromagnetic_couplings}

The models we have considered so far have all used the self-interaction term \(J_D\) in the recurrent network. This was motivated
by the theoretical analysis of Chapter \ref{ch:analytical}, and we have shown in Section \ref{sec:experiment_results_1_layer} that
the choice of \(J_D\) indeed has a significant impact on the performance of the network.

From the biological perspective, however, the self-interaction term only has an interpretation in rate-based models that
capture groups of neurons with a single scalar state variable. In that case, in fact, the presence of self-interactions
corresponds to the presence of strong recurrent connections within the group. If instead we wish to model individual neurons, 
the self-interaction term is not very plausible. As it turns out, we can do without the self-interaction term, and still achieve good performance, in networks with more
than one hidden layer. In fact, the ferromagnetic couplings between adjacent layers have a similar stabilizing effect on the
dynamics as $J_D$. This can be seen both analytically, and empirically.

Analytically, one can repeat the analysis of Chapter \ref{ch:analytical}
for the case of a ferromagnetic chain, which reveals that the ferromagnetic couplings play a similar role as the $J_D$ term
in the single-layer case, with the entropy of fixed points being maximum when the ferromagnetic interaction is symmetric.
This is currently a work in progress.

Empirically, we have considered a multi-layer architecture with \(L\) hidden layers, each containing \(H\) neurons, and ferromagnetic
couplings of constant strength \(\lambda\) between adjacent layers. We have additionally introduced skip connections from the input layer to all hidden layers, and from
the readout layer to all hidden layers. The skip connections are used to induce specialization among the layers, by decaying
their importance with distance: the first layers are mostly influenced by the input, while the last layers are mostly influenced by the readout.
Also the readout layer sees the states of all hidden layers, with decaying importance with distance.

With this architecture, we have performed a simple ablation study. We considered all combinations of two conditions:
\begin{itemize}
    \item the presence or absence of the self-interaction term \(J_D\) (when present, \(J_D = 0.5\));
    \item the presence or absence of the ferromagnetic couplings \(\lambda\) (when present, \(\lambda = 2.0\)).
\end{itemize}
We trained each combination on one tenth of Entangled MNIST, with \(H=300\) and \(L=5\). The results are shown in Table \ref{tab:jd_vs_nojd}.
We see that the performance of the network is roughly the same in all three cases in which at least one between the self-coupling
and the ferromagnetic couplings is present. When both are absent, instead, the performance drops abruptly. For the interpretation
of this result, the presence of the skip connections is crucial: even without the ferromagnetic interaction, the readout still
has access to the states of all hidden layers, and all layers are still influenced by the input.

In any case, while very interesting, this result is still preliminary and it should be more thoroughly investigated in future work.

\begin{table}[ht]
    \centering
    \caption{Ablation study on the impact of the self-interaction term \(J_D\) and the ferromagnetic couplings \(\lambda\) in a multi-layer architecture with skip connections whose importance decays with distance.
    All models are trained on one tenth of Entangled MNIST, with \(H=300\) and \(L=5\), for 40 epochs.}
    \label{tab:jd_vs_nojd}
    \begin{tabular}{@{}llrr@{}}
        \toprule
        $J_D$   & Ferro  & Train Acc & Eval Acc \\ 
        \midrule
        0.5  & 2.0     & 0.91     & 0.79     \\
        0.5  & 0.0     & 0.90     & 0.81     \\[0.5ex]
        0.0  & 2.0     & 0.90     & 0.81     \\
        0.0  & 0.0     & 0.50     & 0.48     \\
        \bottomrule
    \end{tabular}
\end{table}

\begin{figure}[htbp]
  \centering
  \includegraphics[width=0.8\linewidth]{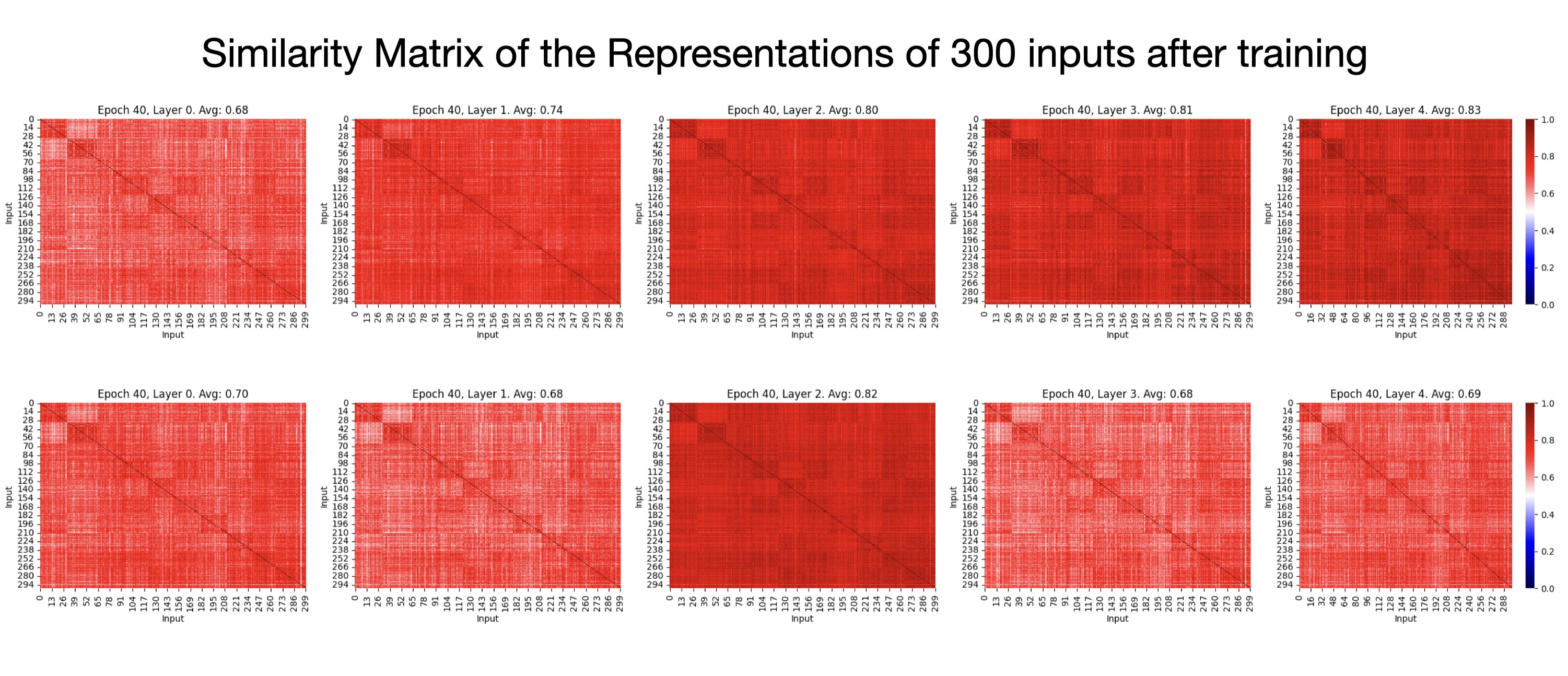}
  \caption{Similarity matrices of the internal representations of the same set of 300 inputs, ordered by their label,
  for the first two rows of Table \ref{tab:jd_vs_nojd}. Each column corresponds to a different layer. The first row
  has \(\lambda = 0.0\) and \(J_D = 0.5\), while the second row has \(\lambda = 2.0\) and \(J_D = 0.5\).}
  \label{fig:skip_connections_input_heatmap}
\end{figure}

\begin{figure}[htbp]
  \centering
  \includegraphics[width=0.8\linewidth]{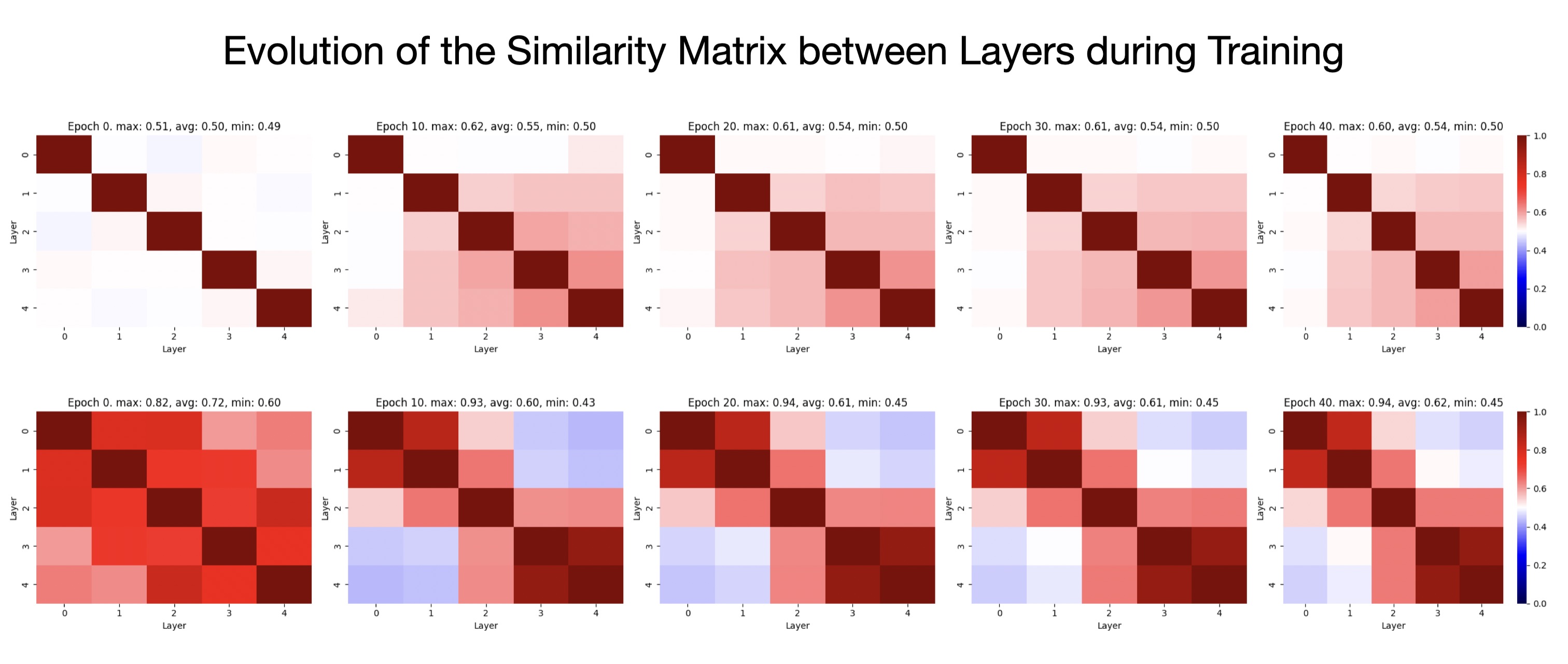}
  \caption{Similarity matrices of the internal representations of different layers, averaged over all inputs,
  for the first two rows of Table \ref{tab:jd_vs_nojd}. Each column corresponds to a different epoch of training.
  The first row has \(\lambda = 0.0\) and \(J_D = 0.5\), while the second row has \(\lambda = 2.0\) and \(J_D = 0.5\).}
  \label{fig:skip_connections_layer_heatmap}
\end{figure}

It is interesting to visualize the representations learned by the network with this deep architecture and skip connections. We show
the comparison between the first two rows of Table \ref{tab:jd_vs_nojd}, to isolate the effect of the ferromagnetic couplings on the 
representations. In Figure \ref{fig:skip_connections_input_heatmap}, we show the similarity matrix of the internal representations 
of the same set of 300 inputs, ordered by their label, after training both networks. In Figure \ref{fig:skip_connections_layer_heatmap},
instead, we show the evolution of another similarity matrix during training. This matrix is computed in the same way as before,
but considering pairs of layers instead of pairs of inputs, and averaging over all inputs.

These figures show, as expected, that
the ferromagnetic couplings have a strong impact on the representations. Without them, the heterogeneity of the input-input
similarity matrices across layers is only due to the decay of the skip connections, and the layer-layer similarity matrices
show that the internal states of different layers are essentially uncorrelated (the small correlation is due to the fact that,
in this model with skip connections, we use the same initialization for the readout matrices of different layers). With the
ferromagnetic couplings, instead, the layers seem to divide in three blocks, the first heavily influenced by the input, the last
heavily influenced by the readout, and the middle one being in between. There is also an effect on the degree of collapse
of the representations across different inputs, with the middle layers collapsing more than the extremal ones.

\subsection{Learning the Longitudinal Couplings}

It is definitely possible to learn the longitudinal couplings in a multi-layer architecture. We have tested several possibilities,
although not systematically. These include combinations of initializing them randomly vs initializing them as sparse ferromagnetic couplings
like in the ferromagnetic chain, and learning only the longitudinal couplings from left to right vs learning those from left to right
while maintaining the ones from right to left equal to the transpose of the ones from left to right (symmetric interaction).
Based on our preliminary experiments, it seems that most of these combinations are somewhat viable, with some effort.

Here, we want to emphasize some of these preliminary results, because they help showcase the potential of our framework.
We consider the case in which the longitudinal couplings are initialized ferromagnetically, like in the ferromagnetic chain model
described in Section \ref{sec:ferromagnetic_chain}, and we learn only the couplings from left to right, while keeping the ones
from right to left fixed (ferromagnetic). This asymmetry introduces directionality in the dynamics, and will be reflected in the
learned representations.

We consider this procedure for training:
\begin{itemize}
    \item do a few warm-up epochs with the recurrent couplings frozen, to learn a matrix \(W^{out}\) that adapts well to the spontaneous representations of the network;
    \item set \(W^{back} = (W^{out})^T\), and freeze both. Set a very aggressive importance for the readout feedback, and learn the recurrent couplings as well as \(W^{in}\);
    \item do a few final epochs of finetuning of \(W^{out}\), with the recurrent couplings frozen.
\end{itemize}
This procedure, by using a readout feedback that accommodates the spontaneous representations of the network, allows to use much
larger readout feedback strengths than in the case of normal training. This way, we are able to fully leverage the additional
expressivity provided by the learnable longitudinal couplings. In doing this, we need to be careful with the treatment of the
very last layer. The way the algorithm is formulated, since the readout feedback is very large, the states of the last layer
are strongly influenced by the class prototypes forced by \(W^{back}\). The problem is that these prototypes only depend on the
class of the input. This means that the updates to the couplings in the last layer tend to interfere constructively across
different batches of inputs, much more than in the previous layers. There are many ways around this, all attempting to control
the growth of the internal couplings of the last layer. For example, we can use a very small learning rate for them.

On the Entangled MNIST dataset, this procedure allows to train a multi-layer architecture with learnable longitudinal couplings
to an accuracy roughly on par with the one obtained in the single-hidden-layer architecture with the same width. That the accuracy is not higher is not
worrying, since not even with backpropagation and feed-forward MLPs are we able to squeeze additional performance from having more than one hidden layer
on this dataset.

\begin{figure}[htbp]
  \centering
  \includegraphics[width=0.8\linewidth]{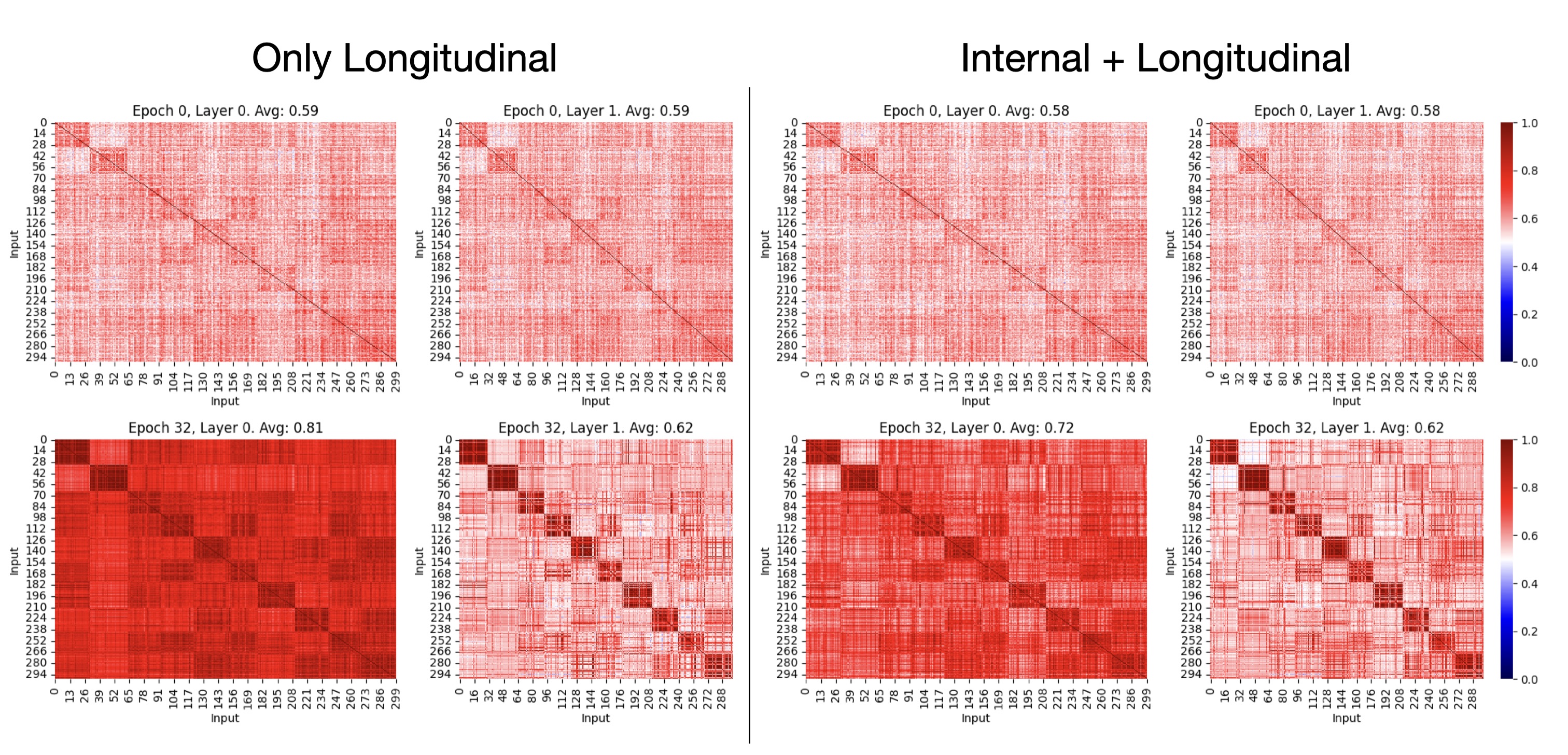}
  \caption{Similarity matrices of the internal representations of the same set of 300 inputs, ordered by their label. 
  First row show the representations before training, while the second row shows the representations after training.
  The right half is a multi-layer architecture with learnable longitudinal couplings from left to right, initialized as ferromagnetic,
  and internal couplings only in the first layer. The left half is a multi-layer architecture with same longitudinal couplings,
  but no internal couplings at all. Each column of each half corresponds to a different layer.}
  \label{fig:longi_representations}
\end{figure}

Furthermore, it is interesting to note that we can train a different version of this model and still achieve good performance and interesting representations:
the connections internal to each hidden layer can be partially or completely removed. In the extreme case in which all internal couplings are removed,
the only feedback remaining in the dynamics is through the backwards longitudinal couplings (ferromagnetic).
In figure \ref{fig:longi_representations}, we show the representations learned by a multi-layer architecture with \(L=2\) having
internal couplings only in the first layer, as well as the extreme case without internal couplings. It can be seen that the representations
in the first layer are less well separated for the less expressive model without internal couplings. However, in this case,
the longitudinal couplings are still expressive enough to obtain well-separated representations in the second layer, and performance
on par with the other model. 

The ability to learn progressively better-separated representations throughout the layers, as seen in Figure \ref{fig:longi_representations} is very encouraging: on more challenging
benchmarks, where depth is necessary to achieve good performance, this could allow to learn progressively more specialized representations
which allow to solve hard classification tasks. Furthermore, the fact that both a strongly horizontal dynamics with fixed longitudinal couplings,
like in the hetero-association task, and an exclusively vertical dynamics with no internal couplings, like here,
have proved to be viable, suggests a relative insensitivity to the details of the topology of the couplings. This is very promising,
looking forward to introducing inductive biases in complex architectures like, for example, convolutional networks.

\subsection{Symmetricity and Evolution of the Dynamics}
\label{sec:symmetricity_and_dynamics}

In this section, we report the results of some ongoing investigations concerning the symmetricity of the internal couplings,
and the evolution of the dynamics during training. We summarize here the main findings, which are still preliminary.

First, we are interested in understanding the effects of symmetry in the initialization of the internal couplings, as well as
the evolution of the level of symmetry during training. To quantify how symmetric the square matrix \(J^\ell\) of internal couplings is,
we do the following. We decompose \(J^\ell\) into its symmetric and antisymmetric parts:
\begin{equation}
    J^\ell = J^\ell_{sym} + J^\ell_{anti} = \frac{1}{2}(J^\ell + (J^\ell)^T) + \frac{1}{2}(J^\ell - (J^\ell)^T).
\end{equation}
The symmetric part is the best approximation of \(J^\ell\) with a symmetric matrix with respect to the Frobenius norm. Thus, 
denoting \(a = \|J^\ell_{sym}\|_F\) and \(b = \|J^\ell_{anti}\|_F\), we define the symmetricity of \(J^\ell\) as \((a - b)/(a + b)\).
This is a number in \([-1, 1]\), where 1 means that the matrix is perfectly symmetric, and -1 means it is perfectly antisymmetric.

We have two preliminary observations:
\begin{itemize}
    \item Initializing the matrix symmetrically does not improve the performance of the network compared to initializing it antisymmetrically. This is true even for values of \(J_D\) that are not large enough to guarantee convergence of the dynamics. In that case, indeed, one can truncate the dynamics after a fixed number of steps; after a number of plasticity steps, the dynamics becomes convergent.
    \item The symmetricity of the internal couplings tends to increase during training if the initial couplings are asymmetric, and to decrease if they are symmetric. In some cases, after a very long time, the symmetricity level of an initially symmetric matrix can even become of similar magnitude to the one of an initially antisymmetric matrix. 
\end{itemize}

The first observation is particularly intriguing, especially the fact that initializing the matrix symmetrically does not improve performance with \(J_D\) very small or even zero.
In fact, the dynamics is convergent in the symmetric case, even with \(J_D = 0\) (provided that one only updates a fraction \(p < 1\) of the neurons at each step, to break limit cycles).
This suggests that the benefits in performance from the presence of \(J_D\) might indeed have to do with the appearance of the large, dense
regions of fixed points that have been observed in the theoretical analysis of Chapter \ref{ch:analytical}. To better understand this, we are
planning to extend the theoretical analysis to the symmetric case, to see whether a similar phenomenology is present, but this is
still a work in progress.

When it comes to the dynamics, we have made two interesting observations concerning the difference between its properties
before and after training. We have considered in particular the inference dynamics, with three types of inputs providing
the external field: zeros (no external field), a random pattern, and a pattern from the training set.
For values in the typical range we use in practice, before training the dynamics is chaotic
without an external field, and less irregular, but still not quite convergent, with an external field (this last statement depends strongly on the value of \(J_D\) and on the strength of
the input external field). This observation is based on plots of the autocorrelation of the internal states, and on the evolution
of the number of unstable neurons.

After training, the dynamics tends to become much less irregular even without an external field, which might be linked
with the increased symmetricity level of the internal couplings, and very quickly convergent with an external field.
What is particularly interesting is that the speed of convergence is lightning fast (<1 \% unsat after 2 sweeps on average)
using inputs from the training set as external field, and considerably less so with random external fields. This suggests
that, maybe, the equilibration time itself could be considered as a criterion for inference in classification tasks (presenting
the input together with all possible classes, one by one, and comparing convergence times), or even as a criterion to perform
anomaly detection. Before exploring these ideas, however, a more thorough investigation is required, especially of the
effects of using correct vs incorrect targets as external fields in addition to the input fields.

\chapter{Conclusion}
\label{ch:conclusion}
\section{Summary and Limitations}

In the theoretical part of this thesis, we have studied the existence and structure of fixed points in random asymmetric
recurrent neural networks, using tools from the statistical mechanics of disordered systems. We have derived an analytical
expression, valid with high probability for large systems, for the number of fixed points as a function of the self-interaction
strength. We have also shown how to use a modification of the 1RSB formalism to study a large-deviation ensemble that gives
more statistical weight to fixed points surrounded by other fixed points. In this way, we found evidence of a phase transition,
as the self-interaction strength crosses a critical value, from a phase in which fixed points are organized in narrow clusters
that satisfy the overlap gap property (OGP), to a phase in which extended, dense regions of fixed points appear and connect
the narrow clusters. We have shown that, in this second phase, fixed points are accessible to fBP, a variant of
Belief Propagation that targets highly locally entropic solutions, while a simple dynamical rule that aligns neurons to their
local field behaves chaotically until a second threshold, scaling slowly with the system size, is crossed. These results
are very relevant for the understanding of the dynamical behavior of neural networks, and they have inspired the design
of the proposed algorithm.

% In future work, two extensions of this analysis could be considered. First, motivated by the
% observations of Section \ref{sec:symmetricity_and_dynamics}, it would be interesting to consider the special case
% of symmetric couplings, to see if a similar phenomenology emerges. If this were the case, it would be an indication
% that the strong impact on the performance of our algorithm due to the choice of the self-interaction strength is
% related to the dense regions of fixed points that we have already identified in the general, asymmetric case. Second, it would be interesting to extend the analysis to layered architectures, for example
% in the simple case in which different layers are connected by ferromagnetic couplings. In fact, evidence from Section
% \ref{sec:ferromagnetic_couplings} suggests that a few strong ferromagnetic couplings should play a similar dynamical
% role to that of the self-interaction. \\

In the second part of this thesis, we have proposed a novel, local and distributed learning algorithm for recurrent
neural networks. We have assessed its performance in a classification task, using a hard version of the MNIST dataset,
and in a hetero-association task, using a synthetic dataset. Our results show that the algorithm is flexible enough to handle a variety
of shallow and deep network architectures, and that it can learn a hierarchy of features to solve non-linearly-separable
classification tasks. We have also shown that depth benefits the performance of the algorithm, by estimating the
hetero-association capacity as function of the depth of a simple layered architecture. Among existing biologically
plausible learning algorithms, our approach is most similar to Predictive Coding (PC) and Equilibrium Propagation (EP).
However, while PC and EP attempt to estimate the backpropagation gradients, our method is non-perturbative, and it does not
even optimize an objective function, unlike both PC and EP. These features make our approach unique in the landscape
of biologically plausible learning algorithms.

In addition to solidifying some of the preliminary results presented in Section \ref{sec:more_preliminary_results},
two main limitations of our work should be addressed. First, our algorithm is currently limited to networks with binary
states, but it is reasonable to expect large performance gains by extending it to work with continuous states. This
will also allow a fair comparison with gradient-based methods and other biologically-plausible algorithms. Second,
since this is a known fragility of many biologically plausible learning algorithms, it will be important to assess the
scalability of our approach to more complex machine learning benchmarks.

\section{Future Directions}

Our work provides novel insights across computational neuroscience, machine learning, and statistical physics. Sparking
from these foundations, there is a wealth of interesting directions that could be pursued.

When it comes to the theoretical analysis, there are at least two natural extensions stemming from the observations made in Section \ref{sec:more_preliminary_results}.
First, motivated by the results of Section \ref{sec:symmetricity_and_dynamics}, it would be interesting to study the phase
diagram of the model assuming symmetric couplings: if a similar phenomenology emerges, this would confirm that the
strong impact of the self-interaction strength on the algorithm performance is related to the emergence of the rare, dense regions of
fixed points unveiled in Section \ref{ch:analytical}. Second, it would be interesting to extend the analysis to a layered
architecture, such as the one considered in Section \ref{sec:ferromagnetic_chain}, to obtain a sharper understanding of
the possibility of substituting the self-interaction with a few strong excitatory couplings, as suggested by the results
of Section \ref{sec:ferromagnetic_couplings}.

Sparking from the algorithm proposed in Section \ref{ch:numerical}, instead, there are two main directions that could
be pursued. First, from the biological perspective, a natural next step would be to further increase the degree of biological 
realism, for example by considering spiking neurons, excitatory and inhibitory couplings, sparse activity of the neurons,
and incorporating insights from connectomics. It would also be interesting to study 
the impact of time-dependent inputs in tasks like time-series forecasting, to consider always-on variants of the plasticity rule,
and to extend the algorithm to work with discrete couplings.
Second, from the machine learning perspective, it would be natural to explore how the algorithm can be adapted to work
with more sophisticated architectures which incorporate inductive biases, such as convolutions or attention mechanisms,
in order to handle complex data modalities more effectively. Also, it could be interesting to investigate the impact
of using more complex dynamical rules, such as fBP, to more effectively access the dense clusters of fixed points, and
to consider protocols for continuous learning.

% - in the theoretical part of this thesis, ... (used ... to study ...; study the phase diagram wrt JD; number of fixed points; existence of dense regions
% of fixed points; first region OGP; second region existence dense regions, fBP can access fixed points, simple dynamis
% chaotic; third region (N-dep thresh) simple dynamics converges)
% - introduced algo, bla bla bla. most similar approaches in literature are PC and EP, but they attempt to estimate backprop. out method is non-perturbative, unlike EQ, and does not even optimize an objective function, unline both PC and EP.
%   limitations: binary states, must consider scalability to more complex benchmarks.
%   future directions: further biological realism (spiking, excitatory/inhibitory, sparsity, connectomics), more sophisticated architectures with inductive biases (e.g. convolutions)

% Bibliography
\printbibliography

@article{addario-berryAlgorithmicHardnessThreshold2020,
  title = {The Algorithmic Hardness Threshold for Continuous Random Energy Models},
  author = {{Addario-Berry}, Louigi and Maillard, Pascal},
  year = {2020},
  month = feb,
  journal = {Mathematical Statistics and Learning},
  volume = {2},
  number = {1},
  pages = {77--101},
  issn = {2520-2316},
  doi = {10.4171/msl/12},
  urldate = {2025-06-06},
  langid = {english}
}

@misc{akroutDeepLearningWeight2020,
  title = {Deep {{Learning}} without {{Weight Transport}}},
  author = {Akrout, Mohamed and Wilson, Collin and Humphreys, Peter C. and Lillicrap, Timothy and Tweed, Douglas},
  year = {2020},
  month = jan,
  number = {arXiv:1904.05391},
  eprint = {1904.05391},
  primaryclass = {cs},
  publisher = {arXiv},
  doi = {10.48550/arXiv.1904.05391},
  urldate = {2025-05-25},
  archiveprefix = {arXiv}
}

@article{amitStoringInfiniteNumbers1985,
  title = {Storing {{Infinite Numbers}} of {{Patterns}} in a {{Spin-Glass Model}} of {{Neural Networks}}},
  author = {Amit, Daniel J. and Gutfreund, Hanoch and Sompolinsky, H.},
  year = {1985},
  month = sep,
  journal = {Phys. Rev. Lett.},
  volume = {55},
  number = {14},
  pages = {1530--1533},
  publisher = {American Physical Society},
  doi = {10.1103/PhysRevLett.55.1530},
  urldate = {2025-06-03}
}

@article{anumanchipalliSpeechSynthesisNeural2019,
  title = {Speech Synthesis from Neural Decoding of Spoken Sentences},
  author = {Anumanchipalli, Gopala K. and Chartier, Josh and Chang, Edward F.},
  year = {2019},
  month = apr,
  journal = {Nature},
  volume = {568},
  number = {7753},
  pages = {493--498},
  publisher = {Nature Publishing Group},
  issn = {1476-4687},
  doi = {10.1038/s41586-019-1119-1},
  urldate = {2025-06-12},
  copyright = {2019 The Author(s), under exclusive licence to Springer Nature Limited},
  langid = {english}
}

@article{asabukiInteractiveReservoirComputing2018,
  title = {Interactive Reservoir Computing for Chunking Information Streams},
  author = {Asabuki, Toshitake and Hiratani, Naoki and Fukai, Tomoki},
  year = {2018},
  month = oct,
  journal = {PLOS Computational Biology},
  volume = {14},
  number = {10},
  pages = {e1006400},
  publisher = {Public Library of Science},
  issn = {1553-7358},
  doi = {10.1371/journal.pcbi.1006400},
  urldate = {2025-06-12},
  langid = {english}
}

@misc{bahdanauNeuralMachineTranslation2014,
  title = {Neural {{Machine Translation}} by {{Jointly Learning}} to {{Align}} and {{Translate}}},
  author = {Bahdanau, Dzmitry and Cho, Kyunghyun and Bengio, Yoshua},
  year = {2014},
  month = sep,
  journal = {arXiv.org},
  urldate = {2025-06-12},
  howpublished = {https://arxiv.org/abs/1409.0473v7},
  langid = {english}
}

@article{baldassiEfficientSupervisedLearning2007,
  title = {Efficient Supervised Learning in Networks with Binary Synapses},
  author = {Baldassi, Carlo and Braunstein, Alfredo and Brunel, Nicolas and Zecchina, Riccardo},
  year = {2007},
  month = jun,
  journal = {Proceedings of the National Academy of Sciences},
  volume = {104},
  number = {26},
  pages = {11079--11084},
  publisher = {Proceedings of the National Academy of Sciences},
  doi = {10.1073/pnas.0700324104},
  urldate = {2025-06-12}
}

@article{baldassiGeneralizationLearningPerceptron2009,
  title = {Generalization {{Learning}} in a {{Perceptron}} with {{Binary Synapses}}},
  author = {Baldassi, Carlo},
  year = {2009},
  month = sep,
  journal = {J Stat Phys},
  volume = {136},
  number = {5},
  pages = {902--916},
  issn = {1572-9613},
  doi = {10.1007/s10955-009-9822-1},
  urldate = {2025-06-12},
  langid = {english}
}

@article{baldassiMaxSumAlgorithmTraining2015,
  title = {A {{Max-Sum}} Algorithm for Training Discrete Neural Networks},
  author = {Baldassi, Carlo and Braunstein, Alfredo},
  year = {2015},
  month = aug,
  journal = {J. Stat. Mech.},
  volume = {2015},
  number = {8},
  pages = {P08008},
  publisher = {{IOP Publishing and SISSA}},
  issn = {1742-5468},
  doi = {10.1088/1742-5468/2015/08/P08008},
  urldate = {2025-06-12},
  langid = {english}
}

@article{baldassiShapingLearningLandscape2020,
  title = {Shaping the Learning Landscape in Neural Networks around Wide Flat Minima},
  author = {Baldassi, Carlo and Pittorino, Fabrizio and Zecchina, Riccardo},
  year = {2020},
  month = jan,
  journal = {Proc. Natl. Acad. Sci. U.S.A.},
  volume = {117},
  number = {1},
  eprint = {1905.07833},
  primaryclass = {cs},
  pages = {161--170},
  issn = {0027-8424, 1091-6490},
  doi = {10.1073/pnas.1908636117},
  urldate = {2025-05-25},
  archiveprefix = {arXiv}
}

@article{baldassiSubdominantDenseClusters2015,
  title = {Subdominant {{Dense Clusters Allow}} for {{Simple Learning}} and {{High Computational Performance}} in {{Neural Networks}} with {{Discrete Synapses}}},
  author = {Baldassi, Carlo and Ingrosso, Alessandro and Lucibello, Carlo and Saglietti, Luca and Zecchina, Riccardo},
  year = {2015},
  month = sep,
  journal = {Phys. Rev. Lett.},
  volume = {115},
  number = {12},
  eprint = {1509.05753},
  primaryclass = {cond-mat},
  pages = {128101},
  issn = {0031-9007, 1079-7114},
  doi = {10.1103/PhysRevLett.115.128101},
  urldate = {2025-05-25},
  archiveprefix = {arXiv}
}

@article{baldassiUnreasonableEffectivenessLearning2016,
  title = {Unreasonable {{Effectiveness}} of {{Learning Neural Networks}}: {{From Accessible States}} and {{Robust Ensembles}} to {{Basic Algorithmic Schemes}}},
  shorttitle = {Unreasonable {{Effectiveness}} of {{Learning Neural Networks}}},
  author = {Baldassi, Carlo and Borgs, Christian and Chayes, Jennifer and Ingrosso, Alessandro and Lucibello, Carlo and Saglietti, Luca and Zecchina, Riccardo},
  year = {2016},
  month = nov,
  journal = {Proc. Natl. Acad. Sci. U.S.A.},
  volume = {113},
  number = {48},
  eprint = {1605.06444},
  primaryclass = {stat},
  issn = {0027-8424, 1091-6490},
  doi = {10.1073/pnas.1608103113},
  urldate = {2025-05-25},
  archiveprefix = {arXiv}
}

@article{baldassiUnveilingStructureWide2021,
  title = {Unveiling the Structure of Wide Flat Minima in Neural Networks},
  author = {Baldassi, Carlo and Lauditi, Clarissa and Malatesta, Enrico M. and Perugini, Gabriele and Zecchina, Riccardo},
  year = {2021},
  month = dec,
  journal = {Phys. Rev. Lett.},
  volume = {127},
  number = {27},
  eprint = {2107.01163},
  primaryclass = {cond-mat},
  pages = {278301},
  issn = {0031-9007, 1079-7114},
  doi = {10.1103/PhysRevLett.127.278301},
  urldate = {2025-05-25},
  archiveprefix = {arXiv}
}

@misc{bengioEarlyInferenceEnergyBased2016,
  title = {Early {{Inference}} in {{Energy-Based Models Approximates Back-Propagation}}},
  author = {Bengio, Yoshua and Fischer, Asja},
  year = {2016},
  month = feb,
  number = {arXiv:1510.02777},
  eprint = {1510.02777},
  primaryclass = {cs},
  publisher = {arXiv},
  doi = {10.48550/arXiv.1510.02777},
  urldate = {2025-06-03},
  archiveprefix = {arXiv}
}

@misc{bengioHowAutoEncodersCould2014,
  title = {How {{Auto-Encoders Could Provide Credit Assignment}} in {{Deep Networks}} via {{Target Propagation}}},
  author = {Bengio, Yoshua},
  year = {2014},
  month = sep,
  number = {arXiv:1407.7906},
  eprint = {1407.7906},
  primaryclass = {cs},
  publisher = {arXiv},
  doi = {10.48550/arXiv.1407.7906},
  urldate = {2025-05-31},
  archiveprefix = {arXiv}
}

@article{biSynapticModificationsCultured1998,
  title = {Synaptic Modifications in Cultured Hippocampal Neurons: Dependence on Spike Timing, Synaptic Strength, and Postsynaptic Cell Type},
  shorttitle = {Synaptic Modifications in Cultured Hippocampal Neurons},
  author = {Bi, G. Q. and Poo, M. M.},
  year = {1998},
  month = dec,
  journal = {J Neurosci},
  volume = {18},
  number = {24},
  pages = {10464--10472},
  issn = {0270-6474},
  doi = {10.1523/JNEUROSCI.18-24-10464.1998},
  langid = {english},
  pmcid = {PMC6793365},
  pmid = {9852584}
}

@misc{bommasaniOpportunitiesRisksFoundation2021,
  title = {On the {{Opportunities}} and {{Risks}} of {{Foundation Models}}},
  author = {Bommasani, Rishi and Hudson, Drew A. and Adeli, Ehsan and Altman, Russ and Arora, Simran and {von Arx}, Sydney and Bernstein, Michael S. and Bohg, Jeannette and Bosselut, Antoine and Brunskill, Emma and Brynjolfsson, Erik and Buch, Shyamal and Card, Dallas and Castellon, Rodrigo and Chatterji, Niladri and Chen, Annie and Creel, Kathleen and Davis, Jared Quincy and Demszky, Dora and Donahue, Chris and Doumbouya, Moussa and Durmus, Esin and Ermon, Stefano and Etchemendy, John and Ethayarajh, Kawin and {Fei-Fei}, Li and Finn, Chelsea and Gale, Trevor and Gillespie, Lauren and Goel, Karan and Goodman, Noah and Grossman, Shelby and Guha, Neel and Hashimoto, Tatsunori and Henderson, Peter and Hewitt, John and Ho, Daniel E. and Hong, Jenny and Hsu, Kyle and Huang, Jing and Icard, Thomas and Jain, Saahil and Jurafsky, Dan and Kalluri, Pratyusha and Karamcheti, Siddharth and Keeling, Geoff and Khani, Fereshte and Khattab, Omar and Koh, Pang Wei and Krass, Mark and Krishna, Ranjay and Kuditipudi, Rohith and Kumar, Ananya and Ladhak, Faisal and Lee, Mina and Lee, Tony and Leskovec, Jure and Levent, Isabelle and Li, Xiang Lisa and Li, Xuechen and Ma, Tengyu and Malik, Ali and Manning, Christopher D. and Mirchandani, Suvir and Mitchell, Eric and Munyikwa, Zanele and Nair, Suraj and Narayan, Avanika and Narayanan, Deepak and Newman, Ben and Nie, Allen and Niebles, Juan Carlos and Nilforoshan, Hamed and Nyarko, Julian and Ogut, Giray and Orr, Laurel and Papadimitriou, Isabel and Park, Joon Sung and Piech, Chris and Portelance, Eva and Potts, Christopher and Raghunathan, Aditi and Reich, Rob and Ren, Hongyu and Rong, Frieda and Roohani, Yusuf and Ruiz, Camilo and Ryan, Jack and R{\'e}, Christopher and Sadigh, Dorsa and Sagawa, Shiori and Santhanam, Keshav and Shih, Andy and Srinivasan, Krishnan and Tamkin, Alex and Taori, Rohan and Thomas, Armin W. and Tram{\`e}r, Florian and Wang, Rose E. and Wang, William and Wu, Bohan and Wu, Jiajun and Wu, Yuhuai and Xie, Sang Michael and Yasunaga, Michihiro and You, Jiaxuan and Zaharia, Matei and Zhang, Michael and Zhang, Tianyi and Zhang, Xikun and Zhang, Yuhui and Zheng, Lucia and Zhou, Kaitlyn and Liang, Percy},
  year = {2021},
  month = aug,
  journal = {arXiv.org},
  urldate = {2025-06-12},
  howpublished = {https://arxiv.org/abs/2108.07258v3},
  langid = {english}
}

@article{bordelonReplicaMethodMachine,
  title = {Replica {{Method}} for the {{Machine Learning Theorist}}},
  author = {Bordelon, Blake and Shan, Haozhe and Canatar, Abdul and Barak, Boaz and Pehlevan, Cengiz},
  langid = {english}
}

@article{braunsteinLearningMessagePassing2006,
  title = {Learning by {{Message Passing}} in {{Networks}} of {{Discrete Synapses}}},
  author = {Braunstein, Alfredo and Zecchina, Riccardo},
  year = {2006},
  month = jan,
  journal = {Phys. Rev. Lett.},
  volume = {96},
  number = {3},
  pages = {030201},
  publisher = {American Physical Society},
  doi = {10.1103/PhysRevLett.96.030201},
  urldate = {2025-06-12}
}

@article{buonomanoStatedependentComputationsSpatiotemporal2009,
  title = {State-Dependent Computations: Spatiotemporal Processing in Cortical Networks},
  shorttitle = {State-Dependent Computations},
  author = {Buonomano, Dean V. and Maass, Wolfgang},
  year = {2009},
  month = feb,
  journal = {Nat Rev Neurosci},
  volume = {10},
  number = {2},
  pages = {113--125},
  publisher = {Nature Publishing Group},
  issn = {1471-0048},
  doi = {10.1038/nrn2558},
  urldate = {2025-06-12},
  copyright = {2009 Springer Nature Limited},
  langid = {english}
}

@article{buonomanoTemporalInformationTransformed1995,
  title = {Temporal {{Information Transformed}} into a {{Spatial Code}} by a {{Neural Network}} with {{Realistic Properties}}},
  author = {Buonomano, Dean V. and Merzenich, Michael M.},
  year = {1995},
  month = feb,
  journal = {Science},
  volume = {267},
  number = {5200},
  pages = {1028--1030},
  publisher = {American Association for the Advancement of Science},
  doi = {10.1126/science.7863330},
  urldate = {2025-06-04}
}

@misc{chaudhariEntropySGDBiasingGradient2017a,
  title = {Entropy-{{SGD}}: {{Biasing Gradient Descent Into Wide Valleys}}},
  shorttitle = {Entropy-{{SGD}}},
  author = {Chaudhari, Pratik and Choromanska, Anna and Soatto, Stefano and LeCun, Yann and Baldassi, Carlo and Borgs, Christian and Chayes, Jennifer and Sagun, Levent and Zecchina, Riccardo},
  year = {2017},
  month = apr,
  number = {arXiv:1611.01838},
  eprint = {1611.01838},
  primaryclass = {cs},
  publisher = {arXiv},
  doi = {10.48550/arXiv.1611.01838},
  urldate = {2025-06-05},
  archiveprefix = {arXiv}
}

@misc{clarkTransientDynamicsAssociative2025,
  title = {Transient Dynamics of Associative Memory Models},
  author = {Clark, David G.},
  year = {2025},
  month = jun,
  number = {arXiv:2506.05303},
  eprint = {2506.05303},
  primaryclass = {cond-mat},
  publisher = {arXiv},
  doi = {10.48550/arXiv.2506.05303},
  urldate = {2025-06-07},
  archiveprefix = {arXiv}
}

@article{crickRecentExcitementNeural1989,
  title = {The Recent Excitement about Neural Networks},
  author = {Crick, F.},
  year = {1989},
  month = jan,
  journal = {Nature},
  volume = {337},
  number = {6203},
  pages = {129--132},
  issn = {0028-0836},
  doi = {10.1038/337129a0},
  langid = {english},
  pmid = {2911347}
}

@article{crisantiDynamicsSpinSystems1987,
  title = {Dynamics of Spin Systems with Randomly Asymmetric Bonds: {{Langevin}} Dynamics and a Spherical Model},
  shorttitle = {Dynamics of Spin Systems with Randomly Asymmetric Bonds},
  author = {Crisanti, A. and Sompolinsky, H.},
  year = {1987},
  month = nov,
  journal = {Phys. Rev. A},
  volume = {36},
  number = {10},
  pages = {4922--4939},
  publisher = {American Physical Society},
  doi = {10.1103/PhysRevA.36.4922},
  urldate = {2025-06-04}
}

@article{daviesLoihiNeuromorphicManycore2018,
  title = {Loihi: {{A Neuromorphic Manycore Processor}} with {{On-Chip Learning}}},
  shorttitle = {Loihi},
  author = {Davies, Mike and Srinivasa, Narayan and Lin, Tsung-Han and Chinya, Gautham and Cao, Yongqiang and Choday, Sri Harsha and Dimou, Georgios and Joshi, Prasad and Imam, Nabil and Jain, Shweta and Liao, Yuyun and Lin, Chit-Kwan and Lines, Andrew and Liu, Ruokun and Mathaikutty, Deepak and McCoy, Steven and Paul, Arnab and Tse, Jonathan and Venkataramanan, Guruguhanathan and Weng, Yi-Hsin and Wild, Andreas and Yang, Yoonseok and Wang, Hong},
  year = {2018},
  month = jan,
  journal = {IEEE Micro},
  volume = {38},
  number = {1},
  pages = {82--99},
  issn = {1937-4143},
  doi = {10.1109/MM.2018.112130359},
  urldate = {2025-06-12}
}

@article{demircigilModelAssociativeMemory2017,
  title = {On a {{Model}} of {{Associative Memory}} with {{Huge Storage Capacity}}},
  author = {Demircigil, Mete and Heusel, Judith and L{\"o}we, Matthias and Upgang, Sven and Vermet, Franck},
  year = {2017},
  month = jul,
  journal = {J Stat Phys},
  volume = {168},
  number = {2},
  pages = {288--299},
  issn = {1572-9613},
  doi = {10.1007/s10955-017-1806-y},
  urldate = {2025-06-03},
  langid = {english}
}

@article{depasqualeFullFORCETargetbasedMethod2018,
  title = {Full-{{FORCE}}: {{A}} Target-Based Method for Training Recurrent Networks},
  shorttitle = {Full-{{FORCE}}},
  author = {DePasquale, Brian and Cueva, Christopher J. and Rajan, Kanaka and Escola, G. Sean and Abbott, L. F.},
  editor = {Chacron, Maurice J.},
  year = {2018},
  month = feb,
  journal = {PLoS ONE},
  volume = {13},
  number = {2},
  pages = {e0191527},
  issn = {1932-6203},
  doi = {10.1371/journal.pone.0191527},
  urldate = {2025-05-29},
  langid = {english}
}

@article{desimoneNeuralMechanismsSelective1995,
  title = {Neural Mechanisms of Selective Visual Attention},
  author = {Desimone, R. and Duncan, J.},
  year = {1995},
  journal = {Annu Rev Neurosci},
  volume = {18},
  pages = {193--222},
  issn = {0147-006X},
  doi = {10.1146/annurev.ne.18.030195.001205},
  langid = {english},
  pmid = {7605061}
}

@article{donohoMessagepassingAlgorithmsCompressed2009,
  title = {Message-Passing Algorithms for Compressed Sensing},
  author = {Donoho, David L. and Maleki, Arian and Montanari, Andrea},
  year = {2009},
  month = nov,
  journal = {Proceedings of the National Academy of Sciences},
  volume = {106},
  number = {45},
  pages = {18914--18919},
  publisher = {Proceedings of the National Academy of Sciences},
  doi = {10.1073/pnas.0909892106},
  urldate = {2025-06-06}
}

@book{engelStatisticalMechanicsLearning2001,
  title = {Statistical {{Mechanics}} of {{Learning}}},
  author = {Engel, A. and {Van den Broeck}, C.},
  year = {2001},
  publisher = {Cambridge University Press},
  address = {Cambridge},
  doi = {10.1017/CBO9781139164542},
  urldate = {2025-06-07},
  isbn = {978-0-521-77307-2}
}

@misc{ernoultEquilibriumPropagationContinual2020,
  title = {Equilibrium {{Propagation}} with {{Continual Weight Updates}}},
  author = {Ernoult, Maxence and Grollier, Julie and Querlioz, Damien and Bengio, Yoshua and Scellier, Benjamin},
  year = {2020},
  month = apr,
  number = {arXiv:2005.04168},
  eprint = {2005.04168},
  primaryclass = {cs},
  publisher = {arXiv},
  doi = {10.48550/arXiv.2005.04168},
  urldate = {2025-06-03},
  archiveprefix = {arXiv}
}

@misc{ernoultScalingDifferenceTarget2022,
  title = {Towards {{Scaling Difference Target Propagation}} by {{Learning Backprop Targets}}},
  author = {Ernoult, Maxence and Normandin, Fabrice and Moudgil, Abhinav and Spinney, Sean and Belilovsky, Eugene and Rish, Irina and Richards, Blake and Bengio, Yoshua},
  year = {2022},
  month = jan,
  journal = {arXiv.org},
  urldate = {2025-06-08},
  howpublished = {https://arxiv.org/abs/2201.13415v1},
  langid = {english}
}

@article{feigelmanAugmentedModelsAssociative1987,
  title = {The Augmented Models of Associative Memory Asymmetric Interaction and Hierarchy of Patterns},
  author = {Feigelman, M.v. and Ioffe, L.b.},
  year = {1987},
  month = apr,
  journal = {Int. J. Mod. Phys. B},
  volume = {01},
  number = {01},
  pages = {51--68},
  publisher = {World Scientific Publishing Co.},
  issn = {0217-9792},
  doi = {10.1142/S0217979287000050},
  urldate = {2025-06-04}
}

@article{feigelmanStatisticalPropertiesHopfield1986,
  title = {The {{Statistical Properties}} of the {{Hopfield Model}} of {{Memory}}},
  author = {Feigelman, M. V. and Ioffe, L. B.},
  year = {1986},
  month = feb,
  journal = {EPL},
  volume = {1},
  number = {4},
  pages = {197},
  issn = {0295-5075},
  doi = {10.1209/0295-5075/1/4/007},
  urldate = {2025-06-04},
  langid = {english}
}

@article{fremauxNeuromodulatedSpikeTimingDependentPlasticity2016,
  title = {Neuromodulated {{Spike-Timing-Dependent Plasticity}}, and {{Theory}} of {{Three-Factor Learning Rules}}},
  author = {Fr{\'e}maux, Nicolas and Gerstner, Wulfram},
  year = {2016},
  month = jan,
  journal = {Front. Neural Circuits},
  volume = {9},
  publisher = {Frontiers},
  issn = {1662-5110},
  doi = {10.3389/fncir.2015.00085},
  urldate = {2025-06-07},
  langid = {english}
}

@article{fukushimaNeocognitronSelforganizingNeural1980,
  title = {Neocognitron: {{A}} Self-Organizing Neural Network Model for a Mechanism of Pattern Recognition Unaffected by Shift in Position},
  shorttitle = {Neocognitron},
  author = {Fukushima, Kunihiko},
  year = {1980},
  month = apr,
  journal = {Biol. Cybernetics},
  volume = {36},
  number = {4},
  pages = {193--202},
  issn = {1432-0770},
  doi = {10.1007/BF00344251},
  urldate = {2025-06-12},
  langid = {english}
}

@article{gamarnikDisorderedSystemsInsights2022,
  title = {Disordered Systems Insights on Computational Hardness},
  author = {Gamarnik, David and Moore, Cristopher and Zdeborov{\'a}, Lenka},
  year = {2022},
  month = nov,
  journal = {J. Stat. Mech.},
  volume = {2022},
  number = {11},
  pages = {114015},
  publisher = {{IOP Publishing and SISSA}},
  issn = {1742-5468},
  doi = {10.1088/1742-5468/ac9cc8},
  urldate = {2025-06-06},
  langid = {english}
}

@article{gamarnikOverlapGapProperty2021,
  title = {The {{Overlap Gap Property}}: A {{Geometric Barrier}} to {{Optimizing}} over {{Random Structures}}},
  shorttitle = {The {{Overlap Gap Property}}},
  author = {Gamarnik, David},
  year = {2021},
  month = oct,
  journal = {Proc. Natl. Acad. Sci. U.S.A.},
  volume = {118},
  number = {41},
  eprint = {2109.14409},
  primaryclass = {cs},
  pages = {e2108492118},
  issn = {0027-8424, 1091-6490},
  doi = {10.1073/pnas.2108492118},
  urldate = {2025-06-05},
  archiveprefix = {arXiv}
}

@misc{gandhiExtendingForwardForward2023,
  title = {Extending the {{Forward Forward Algorithm}}},
  author = {Gandhi, Saumya and Gala, Ritu and Kornberg, Jonah and Sridhar, Advaith},
  year = {2023},
  month = jul,
  number = {arXiv:2307.04205},
  eprint = {2307.04205},
  primaryclass = {cs},
  publisher = {arXiv},
  doi = {10.48550/arXiv.2307.04205},
  urldate = {2025-06-03},
  archiveprefix = {arXiv}
}

@misc{geirhosGeneralisationHumansDeep2018,
  title = {Generalisation in Humans and Deep Neural Networks},
  author = {Geirhos, Robert and Temme, Carlos R. Medina and Rauber, Jonas and Sch{\"u}tt, Heiko H. and Bethge, Matthias and Wichmann, Felix A.},
  year = {2018},
  month = aug,
  journal = {arXiv.org},
  urldate = {2025-06-12},
  howpublished = {https://arxiv.org/abs/1808.08750v3},
  langid = {english}
}

@misc{ghaderBackpropagationfreeSpikingNeural2025,
  title = {Backpropagation-Free {{Spiking Neural Networks}} with the {{Forward-Forward Algorithm}}},
  author = {Ghader, Mohammadnavid and Kheradpisheh, Saeed Reza and Farahani, Bahar and Fazlali, Mahmood},
  year = {2025},
  month = feb,
  journal = {arXiv.org},
  urldate = {2025-06-08},
  howpublished = {https://arxiv.org/abs/2502.20411v2},
  langid = {english}
}

@article{greenburyStructureGenotypephenotypeMaps2022,
  title = {The Structure of Genotype-Phenotype Maps Makes Fitness Landscapes Navigable},
  author = {Greenbury, Sam F. and Louis, Ard A. and Ahnert, Sebastian E.},
  year = {2022},
  month = nov,
  journal = {Nat Ecol Evol},
  volume = {6},
  number = {11},
  pages = {1742--1752},
  publisher = {Nature Publishing Group},
  issn = {2397-334X},
  doi = {10.1038/s41559-022-01867-z},
  urldate = {2025-06-06},
  copyright = {2022 Crown},
  langid = {english}
}

@article{guerraBrokenReplicaSymmetry2003,
  title = {Broken {{Replica Symmetry Bounds}} in the {{Mean Field Spin Glass Model}}},
  author = {Guerra, Francesco},
  year = {2003},
  month = feb,
  journal = {Communications in Mathematical Physics},
  volume = {233},
  number = {1},
  eprint = {cond-mat/0205123},
  pages = {1--12},
  issn = {0010-3616, 1432-0916},
  doi = {10.1007/s00220-002-0773-5},
  urldate = {2025-06-12},
  archiveprefix = {arXiv}
}

@article{hassabisNeuroscienceInspiredArtificialIntelligence2017a,
  title = {Neuroscience-{{Inspired Artificial Intelligence}}},
  author = {Hassabis, Demis and Kumaran, Dharshan and Summerfield, Christopher and Botvinick, Matthew},
  year = {2017},
  month = jul,
  journal = {Neuron},
  volume = {95},
  number = {2},
  pages = {245--258},
  issn = {1097-4199},
  doi = {10.1016/j.neuron.2017.06.011},
  langid = {english},
  pmid = {28728020}
}

@book{hertzIntroductionTheoryNeural2018,
  title = {Introduction {{To The Theory Of Neural Computation}}},
  author = {Hertz, John A.},
  year = {2018},
  month = mar,
  publisher = {CRC Press},
  address = {Boca Raton},
  doi = {10.1201/9780429499661},
  isbn = {978-0-429-49966-1}
}

@misc{HighlyNonrandomFeatures,
  title = {Highly {{Nonrandom Features}} of {{Synaptic Connectivity}} in {{Local Cortical Circuits}} {\textbar} {{PLOS Biology}}},
  urldate = {2025-05-31},
  howpublished = {https://journals.plos.org/plosbiology/article?id=10.1371/journal.pbio.0030068}
}

@misc{hintonForwardForwardAlgorithmPreliminary2022,
  title = {The {{Forward-Forward Algorithm}}: {{Some Preliminary Investigations}}},
  shorttitle = {The {{Forward-Forward Algorithm}}},
  author = {Hinton, Geoffrey},
  year = {2022},
  month = dec,
  number = {arXiv:2212.13345},
  eprint = {2212.13345},
  primaryclass = {cs},
  publisher = {arXiv},
  doi = {10.48550/arXiv.2212.13345},
  urldate = {2025-05-31},
  archiveprefix = {arXiv}
}

@article{hoerzerEmergenceComplexComputational2014,
  title = {Emergence of {{Complex Computational Structures From Chaotic Neural Networks Through Reward-Modulated Hebbian Learning}}},
  author = {Hoerzer, Gregor M. and Legenstein, Robert and Maass, Wolfgang},
  year = {2014},
  month = mar,
  journal = {Cerebral Cortex},
  volume = {24},
  number = {3},
  pages = {677--690},
  issn = {1047-3211},
  doi = {10.1093/cercor/bhs348},
  urldate = {2025-06-07}
}

@article{hopfieldNeuralNetworksPhysical1982,
  title = {Neural Networks and Physical Systems with Emergent Collective Computational Abilities.},
  author = {Hopfield, J J},
  year = {1982},
  month = apr,
  journal = {Proceedings of the National Academy of Sciences},
  volume = {79},
  number = {8},
  pages = {2554--2558},
  publisher = {Proceedings of the National Academy of Sciences},
  doi = {10.1073/pnas.79.8.2554},
  urldate = {2025-06-03}
}

@article{hubelReceptiveFieldsBinocular1962,
  title = {Receptive Fields, Binocular Interaction and Functional Architecture in the Cat's Visual Cortex},
  author = {Hubel, D. H. and Wiesel, T. N.},
  year = {1962},
  month = jan,
  journal = {J Physiol},
  volume = {160},
  number = {1},
  pages = {106-154.2},
  issn = {0022-3751},
  doi = {10.1113/jphysiol.1962.sp006837},
  urldate = {2025-06-12},
  pmcid = {PMC1359523},
  pmid = {14449617}
}

@article{indiveriMemoryInformationProcessing2015,
  title = {Memory and {{Information Processing}} in {{Neuromorphic Systems}}},
  author = {Indiveri, Giacomo and Liu, Shih-Chii},
  year = {2015},
  month = aug,
  journal = {Proceedings of the IEEE},
  volume = {103},
  number = {8},
  pages = {1379--1397},
  issn = {1558-2256},
  doi = {10.1109/JPROC.2015.2444094},
  urldate = {2025-06-12}
}

@inproceedings{jaegerAdaptiveNonlinearSystem2002,
  title = {Adaptive {{Nonlinear System Identification}} with {{Echo State Networks}}},
  booktitle = {Advances in {{Neural Information Processing Systems}}},
  author = {Jaeger, Herbert},
  year = {2002},
  volume = {15},
  publisher = {MIT Press},
  urldate = {2025-06-04}
}

@article{jaegerHarnessingNonlinearityPredicting2004a,
  title = {Harnessing {{Nonlinearity}}: {{Predicting Chaotic Systems}} and {{Saving Energy}} in {{Wireless Communication}}},
  shorttitle = {Harnessing {{Nonlinearity}}},
  author = {Jaeger, Herbert and Haas, Harald},
  year = {2004},
  month = apr,
  journal = {Science},
  volume = {304},
  number = {5667},
  pages = {78--80},
  publisher = {American Association for the Advancement of Science},
  doi = {10.1126/science.1091277},
  urldate = {2025-06-03}
}

@misc{jiangFantasticGeneralizationMeasures2019,
  title = {Fantastic {{Generalization Measures}} and {{Where}} to {{Find Them}}},
  author = {Jiang, Yiding and Neyshabur, Behnam and Mobahi, Hossein and Krishnan, Dilip and Bengio, Samy},
  year = {2019},
  month = dec,
  number = {arXiv:1912.02178},
  eprint = {1912.02178},
  primaryclass = {cs},
  publisher = {arXiv},
  doi = {10.48550/arXiv.1912.02178},
  urldate = {2025-06-05},
  archiveprefix = {arXiv}
}

@article{jumperHighlyAccurateProtein2021,
  title = {Highly Accurate Protein Structure Prediction with {{AlphaFold}}},
  author = {Jumper, John and Evans, Richard and Pritzel, Alexander and Green, Tim and Figurnov, Michael and Ronneberger, Olaf and Tunyasuvunakool, Kathryn and Bates, Russ and {\v Z}{\'i}dek, Augustin and Potapenko, Anna and Bridgland, Alex and Meyer, Clemens and Kohl, Simon A. A. and Ballard, Andrew J. and Cowie, Andrew and {Romera-Paredes}, Bernardino and Nikolov, Stanislav and Jain, Rishub and Adler, Jonas and Back, Trevor and Petersen, Stig and Reiman, David and Clancy, Ellen and Zielinski, Michal and Steinegger, Martin and Pacholska, Michalina and Berghammer, Tamas and Bodenstein, Sebastian and Silver, David and Vinyals, Oriol and Senior, Andrew W. and Kavukcuoglu, Koray and Kohli, Pushmeet and Hassabis, Demis},
  year = {2021},
  month = aug,
  journal = {Nature},
  volume = {596},
  number = {7873},
  pages = {583--589},
  publisher = {Nature Publishing Group},
  issn = {1476-4687},
  doi = {10.1038/s41586-021-03819-2},
  urldate = {2025-06-12},
  copyright = {2021 The Author(s)},
  langid = {english}
}

@article{kadmonTransitionChaosRandom2015,
  title = {Transition to {{Chaos}} in {{Random Neuronal Networks}}},
  author = {Kadmon, Jonathan and Sompolinsky, Haim},
  year = {2015},
  month = nov,
  journal = {Phys. Rev. X},
  volume = {5},
  number = {4},
  pages = {041030},
  publisher = {American Physical Society},
  doi = {10.1103/PhysRevX.5.041030},
  urldate = {2025-05-25}
}

@misc{khajehnejadBiologicalNeuronsCompete2024,
  title = {Biological {{Neurons Compete}} with {{Deep Reinforcement Learning}} in {{Sample Efficiency}} in a {{Simulated Gameworld}}},
  author = {Khajehnejad, Moein and Habibollahi, Forough and Paul, Aswin and Razi, Adeel and Kagan, Brett J.},
  year = {2024},
  month = may,
  number = {arXiv:2405.16946},
  eprint = {2405.16946},
  primaryclass = {q-bio},
  publisher = {arXiv},
  doi = {10.48550/arXiv.2405.16946},
  urldate = {2025-06-12},
  archiveprefix = {arXiv}
}

@book{kosslynFrontiersCognitiveNeuroscience1995,
  title = {Frontiers in {{Cognitive Neuroscience}}},
  author = {Kosslyn, Stephen Michael and Andersen, Richard A.},
  year = {1995},
  publisher = {MIT Press},
  googlebooks = {YwSKMQIymxsC},
  isbn = {978-0-262-61110-7},
  langid = {english}
}

@misc{kozachkovNeuronAstrocyteAssociativeMemory2024,
  title = {Neuron-{{Astrocyte Associative Memory}}},
  author = {Kozachkov, Leo and Slotine, Jean-Jacques and Krotov, Dmitry},
  year = {2024},
  month = jul,
  number = {arXiv:2311.08135},
  eprint = {2311.08135},
  primaryclass = {q-bio},
  publisher = {arXiv},
  doi = {10.48550/arXiv.2311.08135},
  urldate = {2025-06-07},
  archiveprefix = {arXiv}
}

@article{kriegeskorteDeepNeuralNetworks2015,
  title = {Deep {{Neural Networks}}: {{A New Framework}} for {{Modeling Biological Vision}} and {{Brain Information Processing}}},
  shorttitle = {Deep {{Neural Networks}}},
  author = {Kriegeskorte, Nikolaus},
  year = {2015},
  month = nov,
  journal = {Annual Review of Vision Science},
  volume = {1},
  number = {Volume 1, 2015},
  pages = {417--446},
  publisher = {Annual Reviews},
  issn = {2374-4642, 2374-4650},
  doi = {10.1146/annurev-vision-082114-035447},
  urldate = {2025-06-12},
  langid = {english}
}

@inproceedings{krizhevskyImageNetClassificationDeep2012,
  title = {{{ImageNet Classification}} with {{Deep Convolutional Neural Networks}}},
  booktitle = {Advances in {{Neural Information Processing Systems}}},
  author = {Krizhevsky, Alex and Sutskever, Ilya and Hinton, Geoffrey E},
  year = {2012},
  volume = {25},
  publisher = {Curran Associates, Inc.},
  urldate = {2025-06-12}
}

@inproceedings{krotovDenseAssociativeMemory2016,
  title = {Dense {{Associative Memory}} for {{Pattern Recognition}}},
  booktitle = {Advances in {{Neural Information Processing Systems}}},
  author = {Krotov, Dmitry and Hopfield, John J.},
  year = {2016},
  volume = {29},
  publisher = {Curran Associates, Inc.},
  urldate = {2025-06-03}
}

@misc{laborieuxScalingEquilibriumPropagation2020,
  title = {Scaling {{Equilibrium Propagation}} to {{Deep ConvNets}} by {{Drastically Reducing}} Its {{Gradient Estimator Bias}}},
  author = {Laborieux, Axel and Ernoult, Maxence and Scellier, Benjamin and Bengio, Yoshua and Grollier, Julie and Querlioz, Damien},
  year = {2020},
  month = jun,
  number = {arXiv:2006.03824},
  eprint = {2006.03824},
  primaryclass = {cs},
  publisher = {arXiv},
  doi = {10.48550/arXiv.2006.03824},
  urldate = {2025-06-08},
  archiveprefix = {arXiv}
}

@misc{launayDirectFeedbackAlignment2020,
  title = {Direct {{Feedback Alignment Scales}} to {{Modern Deep Learning Tasks}} and {{Architectures}}},
  author = {Launay, Julien and Poli, Iacopo and Boniface, Fran{\c c}ois and Krzakala, Florent},
  year = {2020},
  month = dec,
  number = {arXiv:2006.12878},
  eprint = {2006.12878},
  primaryclass = {stat},
  publisher = {arXiv},
  doi = {10.48550/arXiv.2006.12878},
  urldate = {2025-05-25},
  archiveprefix = {arXiv}
}

@article{lecunDeepLearning2015a,
  title = {Deep Learning},
  author = {LeCun, Yann and Bengio, Yoshua and Hinton, Geoffrey},
  year = {2015},
  month = may,
  journal = {Nature},
  volume = {521},
  number = {7553},
  pages = {436--444},
  publisher = {Nature Publishing Group},
  issn = {1476-4687},
  doi = {10.1038/nature14539},
  urldate = {2025-06-12},
  copyright = {2015 Springer Nature Limited},
  langid = {english}
}

@article{lecunGradientbasedLearningApplied1998,
  title = {Gradient-Based Learning Applied to Document Recognition},
  author = {Lecun, Y. and Bottou, L. and Bengio, Y. and Haffner, P.},
  year = {1998},
  month = nov,
  journal = {Proceedings of the IEEE},
  volume = {86},
  number = {11},
  pages = {2278--2324},
  issn = {1558-2256},
  doi = {10.1109/5.726791},
  urldate = {2025-06-12}
}

@article{lecunMNISTDATABASEHandwritten,
  title = {{{THE MNIST DATABASE}} of Handwritten Digits},
  author = {LECUN, Y.},
  journal = {http://yann.lecun.com/exdb/mnist/},
  urldate = {2025-06-10}
}

@misc{leeDifferenceTargetPropagation2015,
  title = {Difference {{Target Propagation}}},
  author = {Lee, Dong-Hyun and Zhang, Saizheng and Fischer, Asja and Bengio, Yoshua},
  year = {2015},
  month = nov,
  number = {arXiv:1412.7525},
  eprint = {1412.7525},
  primaryclass = {cs},
  publisher = {arXiv},
  doi = {10.48550/arXiv.1412.7525},
  urldate = {2025-06-02},
  archiveprefix = {arXiv}
}

@misc{levyBraintoTextDecodingNoninvasive2025,
  title = {Brain-to-{{Text Decoding}}: {{A Non-invasive Approach}} via {{Typing}}},
  shorttitle = {Brain-to-{{Text Decoding}}},
  author = {L{\'e}vy, Jarod and Zhang, Mingfang and Pinet, Svetlana and Rapin, J{\'e}r{\'e}my and Banville, Hubert and {d'Ascoli}, St{\'e}phane and King, Jean-R{\'e}mi},
  year = {2025},
  month = feb,
  number = {arXiv:2502.17480},
  eprint = {2502.17480},
  primaryclass = {eess},
  publisher = {arXiv},
  doi = {10.48550/arXiv.2502.17480},
  urldate = {2025-06-12},
  archiveprefix = {arXiv}
}

@misc{liaoHowImportantWeight2016,
  title = {How {{Important}} Is {{Weight Symmetry}} in {{Backpropagation}}?},
  author = {Liao, Qianli and Leibo, Joel Z. and Poggio, Tomaso},
  year = {2016},
  month = feb,
  number = {arXiv:1510.05067},
  eprint = {1510.05067},
  primaryclass = {cs},
  publisher = {arXiv},
  doi = {10.48550/arXiv.1510.05067},
  urldate = {2025-05-25},
  archiveprefix = {arXiv}
}

@inproceedings{liCompositeFORCELearning2022,
  title = {Composite {{FORCE Learning}} of {{Chaotic Echo State Networks}} for {{Time-Series Prediction}}},
  booktitle = {2022 41st {{Chinese Control Conference}} ({{CCC}})},
  author = {Li, Yansong and Hu, Kai and Nakajima, Kohei and Pan, Yongping},
  year = {2022},
  month = jul,
  pages = {7355--7360},
  issn = {1934-1768},
  doi = {10.23919/CCC55666.2022.9901897},
  urldate = {2025-06-12}
}

@article{lillicrapBackpropagationBrain2020,
  title = {Backpropagation and the Brain},
  author = {Lillicrap, Timothy P. and Santoro, Adam and Marris, Luke and Akerman, Colin J. and Hinton, Geoffrey},
  year = {2020},
  month = jun,
  journal = {Nat Rev Neurosci},
  volume = {21},
  number = {6},
  pages = {335--346},
  publisher = {Nature Publishing Group},
  issn = {1471-0048},
  doi = {10.1038/s41583-020-0277-3},
  urldate = {2025-05-31},
  copyright = {2020 Springer Nature Limited},
  langid = {english}
}

@misc{liNoPropTrainingNeural2025,
  title = {{{NoProp}}: {{Training Neural Networks}} without {{Back-propagation}} or {{Forward-propagation}}},
  shorttitle = {{{NoProp}}},
  author = {Li, Qinyu and Teh, Yee Whye and Pascanu, Razvan},
  year = {2025},
  month = mar,
  number = {arXiv:2503.24322},
  eprint = {2503.24322},
  primaryclass = {cs},
  publisher = {arXiv},
  doi = {10.48550/arXiv.2503.24322},
  urldate = {2025-05-22},
  archiveprefix = {arXiv}
}

@article{macadangdangAcceleratedEvolutionDiversityGenerating2022,
  title = {Accelerated {{Evolution}} by {{Diversity-Generating Retroelements}}},
  author = {Macadangdang, Benjamin R. and Makanani, Sara K. and Miller, Jeff F.},
  year = {2022},
  month = sep,
  journal = {Annual Review of Microbiology},
  volume = {76},
  number = {Volume 76, 2022},
  pages = {389--411},
  publisher = {Annual Reviews},
  issn = {0066-4227, 1545-3251},
  doi = {10.1146/annurev-micro-030322-040423},
  urldate = {2025-06-06},
  langid = {english}
}

@article{manneschiSpaRCeImprovedLearning2023,
  title = {{{SpaRCe}}: {{Improved Learning}} of {{Reservoir Computing Systems Through Sparse Representations}}},
  shorttitle = {{{SpaRCe}}},
  author = {Manneschi, Luca and Lin, Andrew C. and Vasilaki, Eleni},
  year = {2023},
  month = feb,
  journal = {IEEE Transactions on Neural Networks and Learning Systems},
  volume = {34},
  number = {2},
  pages = {824--838},
  issn = {2162-2388},
  doi = {10.1109/TNNLS.2021.3102378},
  urldate = {2025-06-12}
}

@misc{martinStatisticalMechanicsMethods2001,
  title = {Statistical Mechanics Methods and Phase Transitions in Optimization Problems},
  author = {Martin, O. C. and Monasson, R. and Zecchina, R.},
  year = {2001},
  month = apr,
  number = {arXiv:cond-mat/0104428},
  eprint = {cond-mat/0104428},
  publisher = {arXiv},
  doi = {10.48550/arXiv.cond-mat/0104428},
  urldate = {2025-06-06},
  archiveprefix = {arXiv}
}

@article{matteraChaoticRecurrentNeural2025,
  title = {Chaotic Recurrent Neural Networks for Brain Modelling: {{A}} Review},
  shorttitle = {Chaotic Recurrent Neural Networks for Brain Modelling},
  author = {Mattera, Andrea and Alfieri, Valerio and Granato, Giovanni and Baldassarre, Gianluca},
  year = {2025},
  doi = {10.1016/j.neunet.2024.107079},
  urldate = {2025-06-07},
  langid = {english}
}

@article{mauriTransitionPathsPottslike2023,
  title = {Transition Paths in {{Potts-like}} Energy Landscapes: {{General}} Properties and Application to Protein Sequence Models},
  shorttitle = {Transition Paths in {{Potts-like}} Energy Landscapes},
  author = {Mauri, Eugenio and Cocco, Simona and Monasson, R{\'e}mi},
  year = {2023},
  month = aug,
  journal = {Phys. Rev. E},
  volume = {108},
  number = {2},
  pages = {024141},
  publisher = {American Physical Society},
  doi = {10.1103/PhysRevE.108.024141},
  urldate = {2025-06-06}
}

@article{mazzoniMoreBiologicallyPlausible1991,
  title = {A More Biologically Plausible Learning Rule for Neural Networks.},
  author = {Mazzoni, P and Andersen, R A and Jordan, M I},
  year = {1991},
  month = may,
  journal = {Proceedings of the National Academy of Sciences},
  volume = {88},
  number = {10},
  pages = {4433--4437},
  publisher = {Proceedings of the National Academy of Sciences},
  doi = {10.1073/pnas.88.10.4433},
  urldate = {2025-05-31}
}

@article{merollaMillionSpikingneuronIntegrated2014,
  title = {A Million Spiking-Neuron Integrated Circuit with a Scalable Communication Network and Interface},
  author = {Merolla, Paul A. and Arthur, John V. and {Alvarez-Icaza}, Rodrigo and Cassidy, Andrew S. and Sawada, Jun and Akopyan, Filipp and Jackson, Bryan L. and Imam, Nabil and Guo, Chen and Nakamura, Yutaka and Brezzo, Bernard and Vo, Ivan and Esser, Steven K. and Appuswamy, Rathinakumar and Taba, Brian and Amir, Arnon and Flickner, Myron D. and Risk, William P. and Manohar, Rajit and Modha, Dharmendra S.},
  year = {2014},
  month = aug,
  journal = {Science},
  volume = {345},
  number = {6197},
  pages = {668--673},
  publisher = {American Association for the Advancement of Science},
  doi = {10.1126/science.1254642},
  urldate = {2025-06-12}
}

@misc{meulemansTheoreticalFrameworkTarget2020,
  title = {A {{Theoretical Framework}} for {{Target Propagation}}},
  author = {Meulemans, Alexander and Carzaniga, Francesco S. and Suykens, Johan A. K. and Sacramento, Jo{\~a}o and Grewe, Benjamin F.},
  year = {2020},
  month = dec,
  number = {arXiv:2006.14331},
  eprint = {2006.14331},
  primaryclass = {cs},
  publisher = {arXiv},
  doi = {10.48550/arXiv.2006.14331},
  urldate = {2025-06-02},
  archiveprefix = {arXiv}
}

@article{mezardAnalyticAlgorithmicSolution2002,
  title = {Analytic and {{Algorithmic Solution}} of {{Random Satisfiability Problems}}},
  author = {M{\'e}zard, M. and Parisi, G. and Zecchina, R.},
  year = {2002},
  month = aug,
  journal = {Science},
  volume = {297},
  number = {5582},
  pages = {812--815},
  publisher = {American Association for the Advancement of Science},
  doi = {10.1126/science.1073287},
  urldate = {2025-06-06}
}

@book{mezardInformationPhysicsComputation2009,
  title = {Information, {{Physics}}, and {{Computation}}},
  author = {M{\'e}zard, Marc and Montanari, Andrea},
  year = {2009},
  month = jan,
  edition = {1},
  publisher = {Oxford University PressOxford},
  doi = {10.1093/acprof:oso/9780198570837.001.0001},
  urldate = {2025-06-06},
  isbn = {978-0-19-857083-7 978-0-19-171875-5},
  langid = {english}
}

@book{mezardSpinGlassTheory1986,
  title = {Spin {{Glass Theory}} and {{Beyond}}},
  author = {Mezard, M and Parisi, G and Virasoro, M},
  year = {1986},
  month = nov,
  series = {World {{Scientific Lecture Notes}} in {{Physics}}},
  volume = {Volume 9},
  publisher = {WORLD SCIENTIFIC},
  doi = {10.1142/0271},
  urldate = {2025-06-06},
  isbn = {978-9971-5-0116-7}
}

@article{mikhaeilDifficultyLearningChaotic2022,
  title = {On the Difficulty of Learning Chaotic Dynamics with {{RNNs}}},
  author = {Mikhaeil, Jonas and Monfared, Zahra and Durstewitz, Daniel},
  year = {2022},
  month = dec,
  journal = {Advances in Neural Information Processing Systems},
  volume = {35},
  pages = {11297--11312},
  urldate = {2025-06-07},
  langid = {english}
}

@misc{millidgePredictiveCodingApproximates2020,
  title = {Predictive {{Coding Approximates Backprop}} along {{Arbitrary Computation Graphs}}},
  author = {Millidge, Beren and Tschantz, Alexander and Buckley, Christopher L.},
  year = {2020},
  month = oct,
  number = {arXiv:2006.04182},
  eprint = {2006.04182},
  primaryclass = {cs},
  publisher = {arXiv},
  doi = {10.48550/arXiv.2006.04182},
  urldate = {2025-05-25},
  archiveprefix = {arXiv}
}

@misc{millidgePredictiveCodingTheoretical2022,
  title = {Predictive {{Coding}}: A {{Theoretical}} and {{Experimental Review}}},
  shorttitle = {Predictive {{Coding}}},
  author = {Millidge, Beren and Seth, Anil and Buckley, Christopher L.},
  year = {2022},
  month = jul,
  number = {arXiv:2107.12979},
  eprint = {2107.12979},
  primaryclass = {cs},
  publisher = {arXiv},
  doi = {10.48550/arXiv.2107.12979},
  urldate = {2025-06-02},
  archiveprefix = {arXiv}
}

@misc{millidgePredictiveCodingTheoretical2022a,
  title = {Predictive {{Coding}}: A {{Theoretical}} and {{Experimental Review}}},
  shorttitle = {Predictive {{Coding}}},
  author = {Millidge, Beren and Seth, Anil and Buckley, Christopher L.},
  year = {2022},
  month = jul,
  number = {arXiv:2107.12979},
  eprint = {2107.12979},
  primaryclass = {cs},
  publisher = {arXiv},
  doi = {10.48550/arXiv.2107.12979},
  urldate = {2025-06-03},
  archiveprefix = {arXiv}
}

@article{monassonDeterminingComputationalComplexity1999,
  title = {Determining Computational Complexity from Characteristic `Phase Transitions'},
  author = {Monasson, R{\'e}mi and Zecchina, Riccardo and Kirkpatrick, Scott and Selman, Bart and Troyansky, Lidror},
  year = {1999},
  month = jul,
  journal = {Nature},
  volume = {400},
  number = {6740},
  pages = {133--137},
  publisher = {Nature Publishing Group},
  issn = {1476-4687},
  doi = {10.1038/22055},
  urldate = {2025-06-12},
  copyright = {1999 Macmillan Magazines Ltd.},
  langid = {english}
}

@article{morrisHebbOrganizationBehavior1999,
  title = {D.{{O}}. {{Hebb}}: {{The Organization}} of {{Behavior}}, {{Wiley}}: {{New York}}; 1949},
  shorttitle = {D.{{O}}. {{Hebb}}},
  author = {Morris, R. G.},
  year = {1999},
  journal = {Brain Res Bull},
  volume = {50},
  number = {5-6},
  pages = {437},
  issn = {0361-9230},
  doi = {10.1016/s0361-9230(99)00182-3},
  langid = {english},
  pmid = {10643472}
}

@misc{noklandDirectFeedbackAlignment2016,
  title = {Direct {{Feedback Alignment Provides Learning}} in {{Deep Neural Networks}}},
  author = {N{\o}kland, Arild},
  year = {2016},
  month = dec,
  number = {arXiv:1609.01596},
  eprint = {1609.01596},
  primaryclass = {stat},
  publisher = {arXiv},
  doi = {10.48550/arXiv.1609.01596},
  urldate = {2025-05-25},
  archiveprefix = {arXiv}
}

@article{oconnorINITIALIZEDEQUILIBRIUMPROPAGATION2019,
  title = {{{INITIALIZED EQUILIBRIUM PROPAGATION FOR BACKPROP-FREE TRAINING}}},
  author = {O'Connor, Peter and Gavves, Efstratios and Welling, Max},
  year = {2019},
  langid = {english}
}

@misc{openaiGPT4TechnicalReport2024a,
  title = {{{GPT-4 Technical Report}}},
  author = {OpenAI and Achiam, Josh and Adler, Steven and Agarwal, Sandhini and Ahmad, Lama and Akkaya, Ilge and Aleman, Florencia Leoni and Almeida, Diogo and Altenschmidt, Janko and Altman, Sam and Anadkat, Shyamal and Avila, Red and Babuschkin, Igor and Balaji, Suchir and Balcom, Valerie and Baltescu, Paul and Bao, Haiming and Bavarian, Mohammad and Belgum, Jeff and Bello, Irwan and Berdine, Jake and {Bernadett-Shapiro}, Gabriel and Berner, Christopher and Bogdonoff, Lenny and Boiko, Oleg and Boyd, Madelaine and Brakman, Anna-Luisa and Brockman, Greg and Brooks, Tim and Brundage, Miles and Button, Kevin and Cai, Trevor and Campbell, Rosie and Cann, Andrew and Carey, Brittany and Carlson, Chelsea and Carmichael, Rory and Chan, Brooke and Chang, Che and Chantzis, Fotis and Chen, Derek and Chen, Sully and Chen, Ruby and Chen, Jason and Chen, Mark and Chess, Ben and Cho, Chester and Chu, Casey and Chung, Hyung Won and Cummings, Dave and Currier, Jeremiah and Dai, Yunxing and Decareaux, Cory and Degry, Thomas and Deutsch, Noah and Deville, Damien and Dhar, Arka and Dohan, David and Dowling, Steve and Dunning, Sheila and Ecoffet, Adrien and Eleti, Atty and Eloundou, Tyna and Farhi, David and Fedus, Liam and Felix, Niko and Fishman, Sim{\'o}n Posada and Forte, Juston and Fulford, Isabella and Gao, Leo and Georges, Elie and Gibson, Christian and Goel, Vik and Gogineni, Tarun and Goh, Gabriel and {Gontijo-Lopes}, Rapha and Gordon, Jonathan and Grafstein, Morgan and Gray, Scott and Greene, Ryan and Gross, Joshua and Gu, Shixiang Shane and Guo, Yufei and Hallacy, Chris and Han, Jesse and Harris, Jeff and He, Yuchen and Heaton, Mike and Heidecke, Johannes and Hesse, Chris and Hickey, Alan and Hickey, Wade and Hoeschele, Peter and Houghton, Brandon and Hsu, Kenny and Hu, Shengli and Hu, Xin and Huizinga, Joost and Jain, Shantanu and Jain, Shawn and Jang, Joanne and Jiang, Angela and Jiang, Roger and Jin, Haozhun and Jin, Denny and Jomoto, Shino and Jonn, Billie and Jun, Heewoo and Kaftan, Tomer and Kaiser, {\L}ukasz and Kamali, Ali and Kanitscheider, Ingmar and Keskar, Nitish Shirish and Khan, Tabarak and Kilpatrick, Logan and Kim, Jong Wook and Kim, Christina and Kim, Yongjik and Kirchner, Jan Hendrik and Kiros, Jamie and Knight, Matt and Kokotajlo, Daniel and Kondraciuk, {\L}ukasz and Kondrich, Andrew and Konstantinidis, Aris and Kosic, Kyle and Krueger, Gretchen and Kuo, Vishal and Lampe, Michael and Lan, Ikai and Lee, Teddy and Leike, Jan and Leung, Jade and Levy, Daniel and Li, Chak Ming and Lim, Rachel and Lin, Molly and Lin, Stephanie and Litwin, Mateusz and Lopez, Theresa and Lowe, Ryan and Lue, Patricia and Makanju, Anna and Malfacini, Kim and Manning, Sam and Markov, Todor and Markovski, Yaniv and Martin, Bianca and Mayer, Katie and Mayne, Andrew and McGrew, Bob and McKinney, Scott Mayer and McLeavey, Christine and McMillan, Paul and McNeil, Jake and Medina, David and Mehta, Aalok and Menick, Jacob and Metz, Luke and Mishchenko, Andrey and Mishkin, Pamela and Monaco, Vinnie and Morikawa, Evan and Mossing, Daniel and Mu, Tong and Murati, Mira and Murk, Oleg and M{\'e}ly, David and Nair, Ashvin and Nakano, Reiichiro and Nayak, Rajeev and Neelakantan, Arvind and Ngo, Richard and Noh, Hyeonwoo and Ouyang, Long and O'Keefe, Cullen and Pachocki, Jakub and Paino, Alex and Palermo, Joe and Pantuliano, Ashley and Parascandolo, Giambattista and Parish, Joel and Parparita, Emy and Passos, Alex and Pavlov, Mikhail and Peng, Andrew and Perelman, Adam and Peres, Filipe de Avila Belbute and Petrov, Michael and Pinto, Henrique Ponde de Oliveira and Michael and Pokorny and Pokrass, Michelle and Pong, Vitchyr H. and Powell, Tolly and Power, Alethea and Power, Boris and Proehl, Elizabeth and Puri, Raul and Radford, Alec and Rae, Jack and Ramesh, Aditya and Raymond, Cameron and Real, Francis and Rimbach, Kendra and Ross, Carl and Rotsted, Bob and Roussez, Henri and Ryder, Nick and Saltarelli, Mario and Sanders, Ted and Santurkar, Shibani and Sastry, Girish and Schmidt, Heather and Schnurr, David and Schulman, John and Selsam, Daniel and Sheppard, Kyla and Sherbakov, Toki and Shieh, Jessica and Shoker, Sarah and Shyam, Pranav and Sidor, Szymon and Sigler, Eric and Simens, Maddie and Sitkin, Jordan and Slama, Katarina and Sohl, Ian and Sokolowsky, Benjamin and Song, Yang and Staudacher, Natalie and Such, Felipe Petroski and Summers, Natalie and Sutskever, Ilya and Tang, Jie and Tezak, Nikolas and Thompson, Madeleine B. and Tillet, Phil and Tootoonchian, Amin and Tseng, Elizabeth and Tuggle, Preston and Turley, Nick and Tworek, Jerry and Uribe, Juan Felipe Cer{\'o}n and Vallone, Andrea and Vijayvergiya, Arun and Voss, Chelsea and Wainwright, Carroll and Wang, Justin Jay and Wang, Alvin and Wang, Ben and Ward, Jonathan and Wei, Jason and Weinmann, C. J. and Welihinda, Akila and Welinder, Peter and Weng, Jiayi and Weng, Lilian and Wiethoff, Matt and Willner, Dave and Winter, Clemens and Wolrich, Samuel and Wong, Hannah and Workman, Lauren and Wu, Sherwin and Wu, Jeff and Wu, Michael and Xiao, Kai and Xu, Tao and Yoo, Sarah and Yu, Kevin and Yuan, Qiming and Zaremba, Wojciech and Zellers, Rowan and Zhang, Chong and Zhang, Marvin and Zhao, Shengjia and Zheng, Tianhao and Zhuang, Juntang and Zhuk, William and Zoph, Barret},
  year = {2024},
  month = mar,
  number = {arXiv:2303.08774},
  eprint = {2303.08774},
  primaryclass = {cs},
  publisher = {arXiv},
  doi = {10.48550/arXiv.2303.08774},
  urldate = {2025-06-12},
  archiveprefix = {arXiv}
}

@misc{ororbiaPredictiveForwardForwardAlgorithm2023,
  title = {The {{Predictive Forward-Forward Algorithm}}},
  author = {Ororbia, Alexander and Mali, Ankur},
  year = {2023},
  month = apr,
  number = {arXiv:2301.01452},
  eprint = {2301.01452},
  primaryclass = {cs},
  publisher = {arXiv},
  doi = {10.48550/arXiv.2301.01452},
  urldate = {2025-06-03},
  archiveprefix = {arXiv}
}

@article{papkouRuggedEasilyNavigable2023,
  title = {A Rugged yet Easily Navigable Fitness Landscape},
  author = {Papkou, Andrei and {Garcia-Pastor}, Lucia and Escudero, Jos{\'e} Antonio and Wagner, Andreas},
  year = {2023},
  month = nov,
  journal = {Science},
  volume = {382},
  number = {6673},
  pages = {eadh3860},
  publisher = {American Association for the Advancement of Science},
  doi = {10.1126/science.adh3860},
  urldate = {2025-06-06}
}

@article{parisiAsymmetricNeuralNetworks1986,
  title = {Asymmetric Neural Networks and the Process of Learning},
  author = {Parisi, G.},
  year = {1986},
  month = aug,
  journal = {J. Phys. A: Math. Gen.},
  volume = {19},
  number = {11},
  pages = {L675},
  issn = {0305-4470},
  doi = {10.1088/0305-4470/19/11/005},
  urldate = {2025-06-04},
  langid = {english}
}

@article{parisiSequenceApproximatedSolutions1980,
  title = {A Sequence of Approximated Solutions to the {{S-K}} Model for Spin Glasses},
  author = {Parisi, G.},
  year = {1980},
  month = apr,
  journal = {J. Phys. A: Math. Gen.},
  volume = {13},
  number = {4},
  pages = {L115},
  issn = {0305-4470},
  doi = {10.1088/0305-4470/13/4/009},
  urldate = {2025-06-12},
  langid = {english}
}

@article{pawlakTimingNotEverything2010,
  title = {Timing Is Not {{Everything}}: {{Neuromodulation Opens}} the {{STDP Gate}}},
  shorttitle = {Timing Is Not {{Everything}}},
  author = {Pawlak, Verena and Wickens, Jeffery R. and Kirkwood, Alfredo and Kerr, Jason N. D.},
  year = {2010},
  month = oct,
  journal = {Front Synaptic Neurosci},
  volume = {2},
  pages = {146},
  issn = {1663-3563},
  doi = {10.3389/fnsyn.2010.00146},
  urldate = {2025-06-12},
  pmcid = {PMC3059689},
  pmid = {21423532}
}

@article{pittorinoEntropicGradientDescent2021,
  title = {Entropic Gradient Descent Algorithms and Wide Flat Minima},
  author = {Pittorino, Fabrizio and Lucibello, Carlo and Feinauer, Christoph and Perugini, Gabriele and Baldassi, Carlo and Demyanenko, Elizaveta and Zecchina, Riccardo},
  year = {2021},
  month = dec,
  journal = {J. Stat. Mech.},
  volume = {2021},
  number = {12},
  eprint = {2006.07897},
  primaryclass = {cs},
  pages = {124015},
  issn = {1742-5468},
  doi = {10.1088/1742-5468/ac3ae8},
  urldate = {2025-06-05},
  archiveprefix = {arXiv}
}

@article{priceProbabilisticWeatherForecasting2025,
  title = {Probabilistic Weather Forecasting with Machine Learning},
  author = {Price, Ilan and {Sanchez-Gonzalez}, Alvaro and Alet, Ferran and Andersson, Tom R. and {El-Kadi}, Andrew and Masters, Dominic and Ewalds, Timo and Stott, Jacklynn and Mohamed, Shakir and Battaglia, Peter and Lam, Remi and Willson, Matthew},
  year = {2025},
  month = jan,
  journal = {Nature},
  volume = {637},
  number = {8044},
  pages = {84--90},
  publisher = {Nature Publishing Group},
  issn = {1476-4687},
  doi = {10.1038/s41586-024-08252-9},
  urldate = {2025-06-12},
  copyright = {2024 The Author(s)},
  langid = {english}
}

@article{rabinovichNeuroscienceTransientDynamics2008,
  title = {Neuroscience. {{Transient}} Dynamics for Neural Processing},
  author = {Rabinovich, Misha and Huerta, Ramon and Laurent, Gilles},
  year = {2008},
  month = jul,
  journal = {Science},
  volume = {321},
  number = {5885},
  pages = {48--50},
  issn = {1095-9203},
  doi = {10.1126/science.1155564},
  langid = {english},
  pmid = {18599763}
}

@misc{ramsauerHopfieldNetworksAll2021,
  title = {Hopfield {{Networks}} Is {{All You Need}}},
  author = {Ramsauer, Hubert and Sch{\"a}fl, Bernhard and Lehner, Johannes and Seidl, Philipp and Widrich, Michael and Adler, Thomas and Gruber, Lukas and Holzleitner, Markus and Pavlovi{\'c}, Milena and Sandve, Geir Kjetil and Greiff, Victor and Kreil, David and Kopp, Michael and Klambauer, G{\"u}nter and Brandstetter, Johannes and Hochreiter, Sepp},
  year = {2021},
  month = apr,
  number = {arXiv:2008.02217},
  eprint = {2008.02217},
  primaryclass = {cs},
  publisher = {arXiv},
  doi = {10.48550/arXiv.2008.02217},
  urldate = {2025-06-03},
  archiveprefix = {arXiv}
}

@article{raoPredictiveCodingVisual1999,
  title = {Predictive Coding in the Visual Cortex: A Functional Interpretation of Some Extra-Classical Receptive-Field Effects},
  shorttitle = {Predictive Coding in the Visual Cortex},
  author = {Rao, Rajesh P. N. and Ballard, Dana H.},
  year = {1999},
  month = jan,
  journal = {Nat Neurosci},
  volume = {2},
  number = {1},
  pages = {79--87},
  issn = {1097-6256, 1546-1726},
  doi = {10.1038/4580},
  urldate = {2025-05-25},
  copyright = {http://www.springer.com/tdm},
  langid = {english}
}

@article{rosenblattPerceptronProbabilisticModel1958,
  title = {The Perceptron: {{A}} Probabilistic Model for Information Storage and Organization in the Brain},
  shorttitle = {The Perceptron},
  author = {Rosenblatt, F.},
  year = {1958},
  journal = {Psychological Review},
  volume = {65},
  number = {6},
  pages = {386--408},
  publisher = {American Psychological Association},
  address = {US},
  issn = {1939-1471},
  doi = {10.1037/h0042519}
}

@article{rumelhartLearningRepresentationsBackpropagating1986,
  title = {Learning Representations by Back-Propagating Errors},
  author = {Rumelhart, David E. and Hinton, Geoffrey E. and Williams, Ronald J.},
  year = {1986},
  month = oct,
  journal = {Nature},
  volume = {323},
  number = {6088},
  pages = {533--536},
  publisher = {Nature Publishing Group},
  issn = {1476-4687},
  doi = {10.1038/323533a0},
  urldate = {2025-06-12},
  copyright = {1986 Springer Nature Limited},
  langid = {english}
}

@misc{scellierEquilibriumPropagationBridging2017,
  title = {Equilibrium {{Propagation}}: {{Bridging}} the {{Gap Between Energy-Based Models}} and {{Backpropagation}}},
  shorttitle = {Equilibrium {{Propagation}}},
  author = {Scellier, Benjamin and Bengio, Yoshua},
  year = {2017},
  month = mar,
  number = {arXiv:1602.05179},
  eprint = {1602.05179},
  primaryclass = {cs},
  publisher = {arXiv},
  doi = {10.48550/arXiv.1602.05179},
  urldate = {2025-05-25},
  archiveprefix = {arXiv}
}

@article{schrauwenOverviewReservoirComputing2007,
  title = {An Overview of Reservoir Computing: Theory, Applications and Implementations},
  shorttitle = {An Overview of Reservoir Computing},
  author = {Schrauwen, Benjamin and Verstraeten, David and Van Campenhout, Jan},
  year = {2007},
  journal = {Proceedings of the 15th European Symposium on Artificial Neural Networks. p. 471-482 2007},
  eprint = {1854/LU-416607},
  eprinttype = {hdl},
  pages = {471--482},
  urldate = {2025-06-07},
  copyright = {No license (in copyright)},
  langid = {english}
}

@article{schultzNeuralSubstratePrediction1997,
  title = {A Neural Substrate of Prediction and Reward},
  author = {Schultz, W. and Dayan, P. and Montague, P. R.},
  year = {1997},
  month = mar,
  journal = {Science},
  volume = {275},
  number = {5306},
  pages = {1593--1599},
  issn = {0036-8075},
  doi = {10.1126/science.275.5306.1593},
  langid = {english},
  pmid = {9054347}
}

@article{silverMasteringGameGo2017,
  title = {Mastering the Game of {{Go}} without Human Knowledge},
  author = {Silver, David and Schrittwieser, Julian and Simonyan, Karen and Antonoglou, Ioannis and Huang, Aja and Guez, Arthur and Hubert, Thomas and Baker, Lucas and Lai, Matthew and Bolton, Adrian and Chen, Yutian and Lillicrap, Timothy and Hui, Fan and Sifre, Laurent and {van den Driessche}, George and Graepel, Thore and Hassabis, Demis},
  year = {2017},
  month = oct,
  journal = {Nature},
  volume = {550},
  number = {7676},
  pages = {354--359},
  publisher = {Nature Publishing Group},
  issn = {1476-4687},
  doi = {10.1038/nature24270},
  urldate = {2025-06-12},
  copyright = {2017 Macmillan Publishers Limited, part of Springer Nature. All rights reserved.},
  langid = {english}
}

@article{singhFixedPointsHopfield1995,
  title = {Fixed Points in a {{Hopfield}} Model with Random Asymmetric Interactions},
  author = {Singh, Manoranjan P. and Chengxiang, Zhang and Dasgupta, Chandan},
  year = {1995},
  month = nov,
  journal = {Phys. Rev. E},
  volume = {52},
  number = {5},
  pages = {5261--5272},
  publisher = {American Physical Society},
  doi = {10.1103/PhysRevE.52.5261},
  urldate = {2025-06-04}
}

@article{sompolinskyChaosRandomNeural1988,
  title = {Chaos in {{Random Neural Networks}}},
  author = {Sompolinsky, H. and Crisanti, A. and Sommers, H. J.},
  year = {1988},
  month = jul,
  journal = {Phys. Rev. Lett.},
  volume = {61},
  number = {3},
  pages = {259--262},
  publisher = {American Physical Society},
  doi = {10.1103/PhysRevLett.61.259},
  urldate = {2025-05-25}
}

@article{songHighlyNonrandomFeatures2005a,
  title = {Highly {{Nonrandom Features}} of {{Synaptic Connectivity}} in {{Local Cortical Circuits}}},
  author = {Song, Sen and Sj{\"o}str{\"o}m, Per Jesper and Reigl, Markus and Nelson, Sacha and Chklovskii, Dmitri B.},
  year = {2005},
  month = mar,
  journal = {PLOS Biology},
  volume = {3},
  number = {3},
  pages = {e68},
  publisher = {Public Library of Science},
  issn = {1545-7885},
  doi = {10.1371/journal.pbio.0030068},
  urldate = {2025-06-10},
  langid = {english}
}

@article{stadlerGeneralizedTopologicalSpaces2002,
  title = {Generalized {{Topological Spaces}} in {{Evolutionary Theory}} and {{Combinatorial Chemistry}}},
  author = {Stadler, B{\"a}rbel M. R. and Stadler, Peter F.},
  year = {2002},
  month = may,
  journal = {J. Chem. Inf. Comput. Sci.},
  volume = {42},
  number = {3},
  pages = {577--585},
  publisher = {American Chemical Society},
  issn = {0095-2338},
  doi = {10.1021/ci0100898},
  urldate = {2025-06-06}
}

@article{sternDynamicsRandomNeural2014,
  title = {Dynamics of {{Random Neural Networks}} with {{Bistable Units}}},
  author = {Stern, Merav and Sompolinsky, Haim and Abbott, L. F.},
  year = {2014},
  month = dec,
  journal = {Phys. Rev. E},
  volume = {90},
  number = {6},
  eprint = {1409.2535},
  primaryclass = {cond-mat},
  pages = {062710},
  issn = {1539-3755, 1550-2376},
  doi = {10.1103/PhysRevE.90.062710},
  urldate = {2025-05-29},
  archiveprefix = {arXiv},
  langid = {english}
}

@article{sussilloGeneratingCoherentPatterns2009,
  title = {Generating {{Coherent Patterns}} of {{Activity}} from {{Chaotic Neural Networks}}},
  author = {Sussillo, David and Abbott, L. F.},
  year = {2009},
  month = aug,
  journal = {Neuron},
  volume = {63},
  number = {4},
  pages = {544--557},
  issn = {0896-6273},
  doi = {10.1016/j.neuron.2009.07.018},
  urldate = {2025-05-25},
  pmcid = {PMC2756108},
  pmid = {19709635}
}

@article{suttonLearningPredictMethods1988,
  title = {Learning to Predict by the Methods of Temporal Differences},
  author = {Sutton, Richard S.},
  year = {1988},
  month = aug,
  journal = {Mach Learn},
  volume = {3},
  number = {1},
  pages = {9--44},
  issn = {1573-0565},
  doi = {10.1007/BF00115009},
  urldate = {2025-06-12},
  langid = {english}
}

@article{talagrandParisiFormula2006,
  title = {The {{Parisi Formula}}},
  author = {Talagrand, Michel},
  year = {2006},
  journal = {Annals of Mathematics},
  volume = {163},
  number = {1},
  eprint = {20159953},
  eprinttype = {jstor},
  pages = {221--263},
  publisher = {Annals of Mathematics},
  issn = {0003-486X},
  urldate = {2025-06-12}
}

@inproceedings{tamuraPartialFORCEFastRobust2021,
  title = {Partial-{{FORCE}}: {{A}} Fast and Robust Online Training Method for Recurrent Neural Networks},
  shorttitle = {Partial-{{FORCE}}},
  booktitle = {2021 {{International Joint Conference}} on {{Neural Networks}} ({{IJCNN}})},
  author = {Tamura, Hiroto and Tanaka, Gouhei},
  year = {2021},
  month = jul,
  pages = {1--8},
  issn = {2161-4407},
  doi = {10.1109/IJCNN52387.2021.9533964},
  urldate = {2025-06-12}
}

@article{tamuraTransferRLSMethodTransferFORCE2021,
  title = {Transfer-{{RLS}} Method and Transfer-{{FORCE}} Learning for Simple and Fast Training of Reservoir Computing Models},
  author = {Tamura, Hiroto and Tanaka, Gouhei},
  year = {2021},
  month = nov,
  journal = {Neural Networks},
  volume = {143},
  pages = {550--563},
  issn = {0893-6080},
  doi = {10.1016/j.neunet.2021.06.031},
  urldate = {2025-06-12}
}

@article{tanakaRecentAdvancesPhysical2019,
  title = {Recent Advances in Physical Reservoir Computing: {{A}} Review},
  shorttitle = {Recent Advances in Physical Reservoir Computing},
  author = {Tanaka, Gouhei and Yamane, Toshiyuki and H{\'e}roux, Jean Benoit and Nakane, Ryosho and Kanazawa, Naoki and Takeda, Seiji and Numata, Hidetoshi and Nakano, Daiju and Hirose, Akira},
  year = {2019},
  month = jul,
  journal = {Neural Networks},
  volume = {115},
  pages = {100--123},
  issn = {0893-6080},
  doi = {10.1016/j.neunet.2019.03.005},
  urldate = {2025-06-07}
}

@misc{ThreeFactorLearningSpiking,
  title = {Three-{{Factor Learning}} in {{Spiking Neural Networks}}: {{An Overview}} of {{Methods}} and {{Trends}} from a {{Machine Learning Perspective}}},
  urldate = {2025-06-03},
  howpublished = {https://arxiv.org/html/2504.05341v1}
}

@article{toulouseSpinGlassModel1986,
  title = {Spin Glass Model of Learning by Selection.},
  author = {Toulouse, G and Dehaene, S and Changeux, J P},
  year = {1986},
  month = mar,
  journal = {Proceedings of the National Academy of Sciences},
  volume = {83},
  number = {6},
  pages = {1695--1698},
  publisher = {Proceedings of the National Academy of Sciences},
  doi = {10.1073/pnas.83.6.1695},
  urldate = {2025-06-04}
}

@article{unnikrishnanAlopexCorrelationBasedLearning1994,
  title = {Alopex: {{A Correlation-Based Learning Algorithm}} for {{Feedforward}} and {{Recurrent Neural Networks}}},
  shorttitle = {Alopex},
  author = {Unnikrishnan, K. P. and Venugopal, K. P.},
  year = {1994},
  month = may,
  journal = {Neural Computation},
  volume = {6},
  number = {3},
  pages = {469--490},
  issn = {0899-7667},
  doi = {10.1162/neco.1994.6.3.469},
  urldate = {2025-05-31}
}

@misc{vaswaniAttentionAllYou2023,
  title = {Attention {{Is All You Need}}},
  author = {Vaswani, Ashish and Shazeer, Noam and Parmar, Niki and Uszkoreit, Jakob and Jones, Llion and Gomez, Aidan N. and Kaiser, Lukasz and Polosukhin, Illia},
  year = {2023},
  month = aug,
  number = {arXiv:1706.03762},
  eprint = {1706.03762},
  primaryclass = {cs},
  publisher = {arXiv},
  doi = {10.48550/arXiv.1706.03762},
  urldate = {2025-06-07},
  archiveprefix = {arXiv}
}

@inproceedings{werfelLearningCurvesStochastic2003,
  title = {Learning {{Curves}} for {{Stochastic Gradient Descent}} in {{Linear Feedforward Networks}}},
  booktitle = {Advances in {{Neural Information Processing Systems}}},
  author = {Werfel, Justin and Xie, Xiaohui and Seung, H.},
  year = {2003},
  volume = {16},
  publisher = {MIT Press},
  urldate = {2025-05-31}
}

@article{whittingtonApproximationErrorBackpropagation2017,
  title = {An {{Approximation}} of the {{Error Backpropagation Algorithm}} in a {{Predictive}}          {{Coding Network}} with {{Local Hebbian Synaptic Plasticity}}},
  author = {Whittington, James C. R. and Bogacz, Rafal},
  year = {2017},
  month = may,
  journal = {Neural Computation},
  volume = {29},
  number = {5},
  pages = {1229--1262},
  publisher = {MIT Press - Journals},
  issn = {0899-7667, 1530-888X},
  doi = {10.1162/neco_a_00949},
  urldate = {2025-05-22},
  copyright = {https://creativecommons.org/licenses/by-nc/4.0/},
  langid = {english}
}

@article{whittingtonTheoriesErrorBackPropagation2019,
  title = {Theories of {{Error Back-Propagation}} in the {{Brain}}},
  author = {Whittington, James C. R. and Bogacz, Rafal},
  year = {2019},
  month = mar,
  journal = {Trends in Cognitive Sciences},
  volume = {23},
  number = {3},
  pages = {235--250},
  publisher = {Elsevier},
  issn = {1364-6613, 1879-307X},
  doi = {10.1016/j.tics.2018.12.005},
  urldate = {2025-05-31},
  langid = {english},
  pmid = {30704969}
}

@article{williamsSimpleStatisticalGradientfollowing1992,
  title = {Simple Statistical Gradient-Following Algorithms for Connectionist Reinforcement Learning},
  author = {Williams, Ronald J.},
  year = {1992},
  month = may,
  journal = {Mach Learn},
  volume = {8},
  number = {3},
  pages = {229--256},
  issn = {1573-0565},
  doi = {10.1007/BF00992696},
  urldate = {2025-05-31},
  langid = {english}
}

@article{wuAdaptationProteinFitness2016,
  title = {Adaptation in Protein Fitness Landscapes Is Facilitated by Indirect Paths},
  author = {Wu, Nicholas C and Dai, Lei and Olson, C Anders and {Lloyd-Smith}, James O and Sun, Ren},
  editor = {Neher, Richard A},
  year = {2016},
  month = jul,
  journal = {eLife},
  volume = {5},
  pages = {e16965},
  publisher = {eLife Sciences Publications, Ltd},
  issn = {2050-084X},
  doi = {10.7554/eLife.16965},
  urldate = {2025-06-06}
}

@misc{xiaoBiologicallyplausibleLearningAlgorithms2018,
  title = {Biologically-Plausible Learning Algorithms Can Scale to Large Datasets},
  author = {Xiao, Will and Chen, Honglin and Liao, Qianli and Poggio, Tomaso},
  year = {2018},
  month = dec,
  number = {arXiv:1811.03567},
  eprint = {1811.03567},
  primaryclass = {cs},
  publisher = {arXiv},
  doi = {10.48550/arXiv.1811.03567},
  urldate = {2025-06-02},
  archiveprefix = {arXiv}
}

@article{yaminsUsingGoaldrivenDeep2016,
  title = {Using Goal-Driven Deep Learning Models to Understand Sensory Cortex},
  author = {Yamins, Daniel L. K. and DiCarlo, James J.},
  year = {2016},
  month = mar,
  journal = {Nat Neurosci},
  volume = {19},
  number = {3},
  pages = {356--365},
  publisher = {Nature Publishing Group},
  issn = {1546-1726},
  doi = {10.1038/nn.4244},
  urldate = {2025-06-12},
  copyright = {2016 Springer Nature America, Inc.},
  langid = {english}
}

@inproceedings{yinImprovingFullFORCEDynamical2021,
  title = {Improving Full-{{FORCE}} with Dynamical Data Coupling and Multilayer Architecture},
  booktitle = {2021 {{IEEE Biomedical Circuits}} and {{Systems Conference}} ({{BioCAS}})},
  author = {Yin, Yue and Neftci, Emre},
  year = {2021},
  month = oct,
  pages = {1--6},
  issn = {2163-4025},
  doi = {10.1109/BioCAS49922.2021.9645012},
  urldate = {2025-06-12}
}

@article{yuEvaluatingGrayWhite2018,
  title = {Evaluating the Gray and White Matter Energy Budgets of Human Brain Function},
  author = {Yu, Yuguo and Herman, Peter and Rothman, Douglas L and Agarwal, Divyansh and Hyder, Fahmeed},
  year = {2018},
  month = aug,
  journal = {J Cereb Blood Flow Metab},
  volume = {38},
  number = {8},
  pages = {1339--1353},
  issn = {0271-678X},
  doi = {10.1177/0271678X17708691},
  urldate = {2025-06-12},
  pmcid = {PMC6092772},
  pmid = {28589753}
}

@article{zhangTargetPropagation2015,
  title = {Target {{Propagation}}},
  author = {Zhang, Saizheng and Bengio, Y},
  year = {2015},
  langid = {english}
}

@misc{zhengRFORCERobustLearning2025,
  title = {R-{{FORCE}}: {{Robust Learning}} for {{Random Recurrent Neural Networks}}},
  shorttitle = {R-{{FORCE}}},
  author = {Zheng, Yang and Shlizerman, Eli},
  year = {2025},
  month = jun,
  number = {arXiv:2003.11660},
  eprint = {2003.11660},
  primaryclass = {cs},
  publisher = {arXiv},
  doi = {10.48550/arXiv.2003.11660},
  urldate = {2025-06-12},
  archiveprefix = {arXiv}
}

% Appendices
\appendix
\chapter{Additional Materials}
\label{app:additional}
\section{Code}

Code to reproduce our results is available on GitHub at the following url:

\noindent
https://github.com/mattia-scardecchia/Biological-Learning.

\end{document}